\documentclass{article}

\usepackage{PRIMEarxiv}

\usepackage[utf8]{inputenc} 
\usepackage[T1]{fontenc}    
\usepackage{hyperref}       
\usepackage{url}            
\usepackage{booktabs}       
\usepackage{amsfonts}       
\usepackage{nicefrac}       
\usepackage{microtype}      
\usepackage{lipsum}
\usepackage{fancyhdr}       
\usepackage{graphicx}       
\graphicspath{{media/}}     

\usepackage{amsmath}
\usepackage{amssymb}

\usepackage{ragged2e} %
\usepackage{multirow} %
\usepackage{makecell}
\usepackage{array} 

\usepackage[most]{tcolorbox}
\usepackage{tabularx} 

\newcolumntype{L}{>{\raggedright\arraybackslash}X} %
\newcolumntype{C}[1]{>{\centering\arraybackslash}X{#1}} %

\usepackage[backend=biber, style=numeric, sorting=none]{biblatex}

\defbibenvironment{appendixbib}
  {\list
     {\printtext[labelnumberwidth]{%
        \printfield{prefixnumber}%
        \printfield{labelnumber}}}
     {\setlength{\labelwidth}{\labelnumberwidth}%
      \setlength{\leftmargin}{\labelwidth}%
      \setlength{\labelsep}{\biblabelsep}%
      \addtolength{\leftmargin}{\labelsep}%
      \setlength{\itemsep}{\bibitemsep}%
      \setlength{\parsep}{\bibparsep}}}
  {\endlist}
  {\item}

\newcommand*{\setappendixprefix}[1]{%
  \DeclareFieldFormat{labelnumber}{#1##1}
}

\title{Dynamical Alignment: A Principle for Adaptive Neural Computation}

\author{
  Xia Chen \\
  Georg Nemetschek Institute \\
  Munich Data Science Institute \\
  Technische Universität München \\
  \texttt{x.c.chen@tum.de} \\
}

\begin{document}
\maketitle

\begin{abstract}

The computational capabilities of a neural network are widely assumed to be determined by its static architecture. Here we challenge this view by establishing that a fixed neural structure can operate in fundamentally different computational modes, driven not by its structure but by the temporal dynamics of its input signals. We term this principle 'Dynamical Alignment'.

Applying this principle offers a novel resolution to the long-standing paradox of why brain-inspired spiking neural networks (SNNs) underperform. By encoding static input into controllable dynamical trajectories, we uncover a bimodal optimization landscape with a critical phase transition governed by phase space volume dynamics. A 'dissipative' mode, driven by contracting dynamics, achieves superior energy efficiency through sparse temporal codes. In contrast, an 'expansive' mode, driven by expanding dynamics, unlocks the representational power required for SNNs to match or even exceed their artificial neural network counterparts on diverse tasks, including classification, reinforcement learning, and cognitive integration.

We find this computational advantage emerges from a timescale alignment between input dynamics and neuronal integration. This principle, in turn, offers a unified, computable perspective on long-observed dualities in neuroscience, from stability-plasticity dilemma to segregation-integration dynamic. It demonstrates that computation in both biological and artificial systems can be dynamically sculpted by 'software' on fixed 'hardware', pointing toward a potential paradigm shift for AI research: away from designing complex static architectures and toward mastering adaptive, dynamic computation principles.
\begin{figure}[h!]
    \centering
    \includegraphics[width=\textwidth]{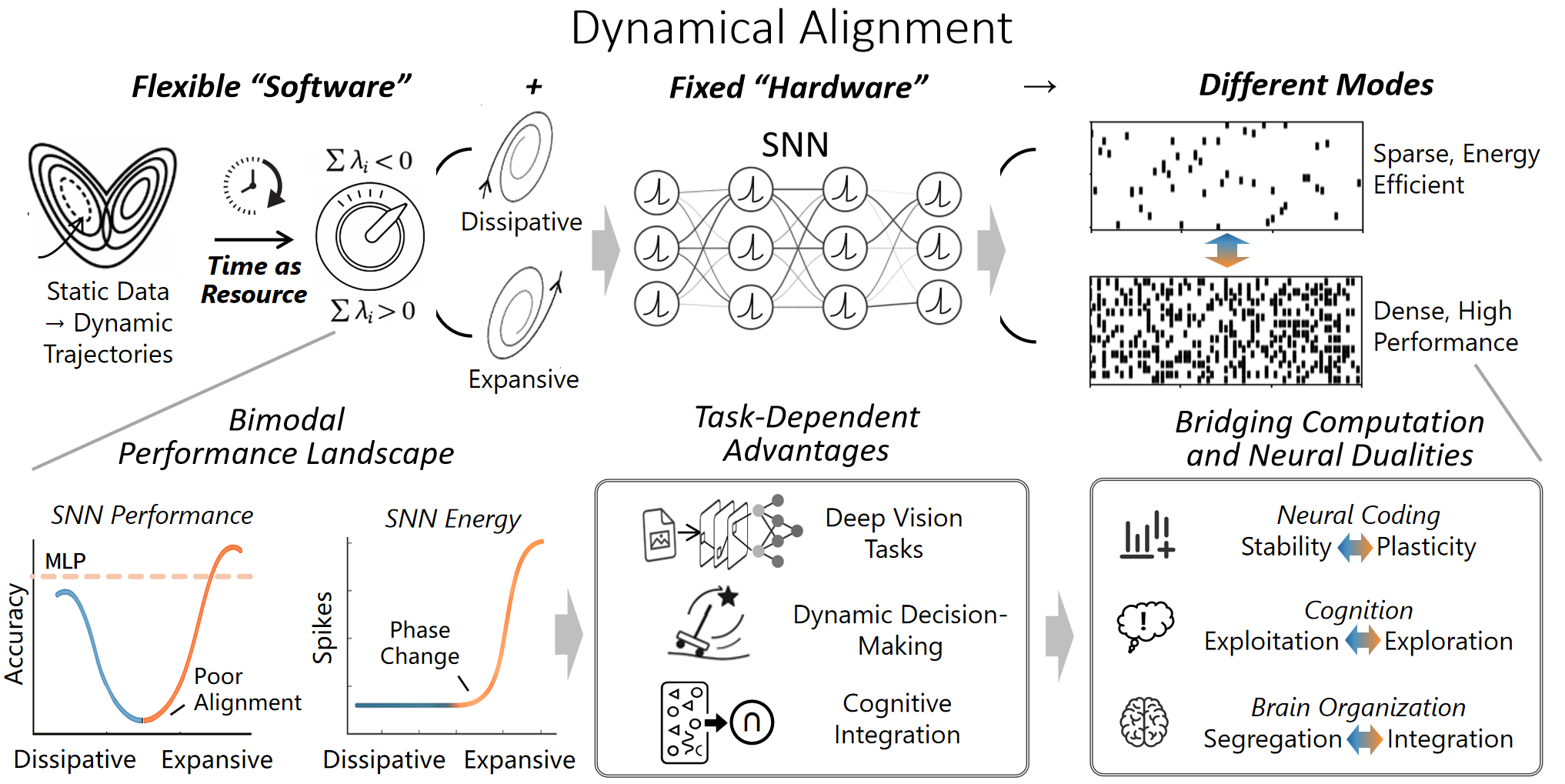}
    \label{fig:graphical_abstract}
\end{figure}
\end{abstract}


\section{The Computational Challenge of Neural Efficiency}
The information processing capacity and efficiency of biological neural systems have long fascinated and perplexed scientists. The human brain consumes only about 20 watts of power, whereas AI systems performing similar functions require several kilowatts\cite{merolla2014million}. However, despite being inspired by biological neural systems with significant theoretical energy advantages, Spiking Neural Networks (SNNs) have long struggled to match the performance of traditional deep learning systems\cite{tavanaei2019deep, pfeiffer2018deep}. Equally intriguing is the observation that biological neural activity exhibits properties consistent with a system operating at a critical state between order and disorder\cite{rabinovich1998role}, a regime often described as 'the edge of chaos'\cite{langton1990computation}. Its functional networks can flexibly switch between different dynamical modes\cite{deco2015rethinking,tononi1994measure}, yet the underlying computational principles remain largely unclear. This paradox of theoretical efficiency yet practical underperformance, combined with the brain's distinct dynamical state, raises a fundamental question: \textit{Does efficient neural computation require alignment between input and network dynamics?}

\begin{tcolorbox}[
    colback=black!3!white,  
    colframe=black!30!black, 
    fonttitle=\bfseries,     
    title= SNNs vs. ANNs: A Primer
    ]
Traditional ANNs compute using continuous-valued neuron activations and adjustments to static connection weights, with information primarily represented in the spatial domain. In contrast, SNNs mimic the spiking mechanism of biological neurons, where neurons generate discrete pulses, or 'spikes', only when their membrane potential exceeds a threshold. This allows information to be encoded in both space and time. While this event-driven computation theoretically offers significant energy savings, it also increases the complexity of information encoding and processing.
\end{tcolorbox}

Revisiting the performance paradox of neuromorphic computing, we propose the 'Dynamical Alignment' hypothesis: \textit{Neural efficiency emerges when input dynamics are properly aligned with network processing characteristics.} Although encoding information in the temporal dimension is known to be crucial for SNNs, a systematic framework for aligning these temporal codes with a network's inherent processing characteristics has been missing\cite{tavanaei2019deep, rabinovich2008transient, panzeri2010sensory, eshraghian2023training}. We argue that to fully leverage the spatiotemporal processing capabilities of SNNs, information must be appropriately distributed in the phase space of a dynamical system. This perspective is supported by studies of biological neural systems, which show that these systems exploit nonlinear dynamics, particularly the sensitive dependence on initial conditions characteristic of chaos, to efficiently encode and process information\cite{skarda1987brains,tsuda2001toward,freeman1987simulation}. To ensure clarity throughout our analysis, we explicitly distinguish between two key concepts:

\begin{itemize}
    \item \textbf{Neural Encoding:} The process of converting static information into spatiotemporal patterns with specific dynamical properties.
    \item \textbf{Neural Computation:} The process by which a neural network processes these spatiotemporal patterns to perform a specific task.
\end{itemize}

In Shaw's pioneering work on chaotic systems\cite{Shaw1984}, he noted that what appears to be random background behavior may not be noise, but rather a manifestation of deterministic chaos, an inherent property of the system's essential dynamics. This observation has led to the conventional view that the 'Edge of Chaos' is the ideal state for information processing\cite{6790172,buzsaki2006rhythms}. This study takes a different path by focusing primarily on the neural encoding stage, thereby proposing a new role for the time dimension: not merely as a channel for data, but as a dynamic resource for controlling and optimizing computation.


\section{Experimental Validation of Dynamical Alignment}
\label{sec:validation}

\subsection{Core Encoding Principle}
\label{subsec:encoding_principle}

Our approach, termed dynamical encoding, redefines neural information representation by transforming static data points into rich spatiotemporal trajectories. This is achieved by leveraging the time-evolution of a dynamical system in three key steps:
\begin{enumerate}
\item Using the input data as the initial conditions for the dynamical system.
\item Generating the system's trajectory in phase space through numerical integration.
\item Converting these trajectories, which contain time-evolved information, into spike pattern inputs for the SNN.
\end{enumerate}

The core of this approach lies in exploiting the unique structures formed by the behavior of the dynamical system in its phase space (e.g., strange attractors). This process ensures that even subtly different inputs develop into distinct patterns over time, providing a far richer means of representation than simple time-series conversion\cite{auge2021survey}, forming a flexible yet orderly code for the SNN.

\subsection{Validation Strategy and Experiments}
\label{subsec:strategy_results}

To systematically test our core 'Dynamical Alignment' hypothesis, we designed two experiments to validate the effectiveness: (1) \textbf{network-out preprocessing}, where dynamical encoding is performed as feature engineering before data input; and (2) \textbf{network-in bridging}, which integrates dynamical encoding as a bridging layer within the network. These two strategies are designed to respectively probe the hypothesis's validity and its engineering feasibility.

In the first experiment, we evaluated our method on the MNIST handwritten digit classification dataset\cite{lecun1998mnist}. We compared it against an identically structured SNN using several state-of-the-art (SOTA) encoding methods, as well as a Multi-Layer Perceptron (MLP), a representative ANN architecture for this task. The SNN encoding schemes included Rate\cite{eshraghian2023training, adrian1926impulses}, Phase\cite{montemurro2008phase}, Latency\cite{eshraghian2023training}, Time-to-First-Spike\cite{rueckauer2018conversion}, Delta\cite{eshraghian2023training}, and Burst\cite{park2019fast}. To isolate the effects of encoding quality from dimensionality changes, we first preprocessed the data using UMAP\cite{mcinnes2018umap}. This nonlinear dimensionality reduction process also serves to simulate the primary feature extraction found in biological sensory systems\cite{dicarlo2012does}.

In our second experiment, we evaluated the 'network-in bridging' approach on the more challenging CIFAR-10 dataset\cite{krizhevsky2009learning}. The setup was based on a hybrid architecture: a convolutional neural network (CNN) front-end for feature extraction, followed by one of two alternative classifier heads: (1) a baseline MLP, or (2) a dynamical encoding layer that transforms the features into spatiotemporal patterns for a subsequent SNN classifier. A key distinction from the first experiment is that the entire architecture was trained end-to-end. This joint optimization functionally simulates the progressive re-encoding in sensory systems\cite{dicarlo2012does}, allowed the CNN's feature representations to be specifically tuned for the subsequent dynamical encoding stage. To assess the method's robustness to varying information compression, we tested nine different CNN configurations. These configurations spanned three channel depths (C8-32, C16-64, C32-128) and three kernel sizes ($3 \times 3$, $5 \times 5$, $7 \times 7$), allowing us to compare performance across a spectrum of feature representation complexities. 

\begin{figure}[h!]
    \centering
    \includegraphics[width=\textwidth]{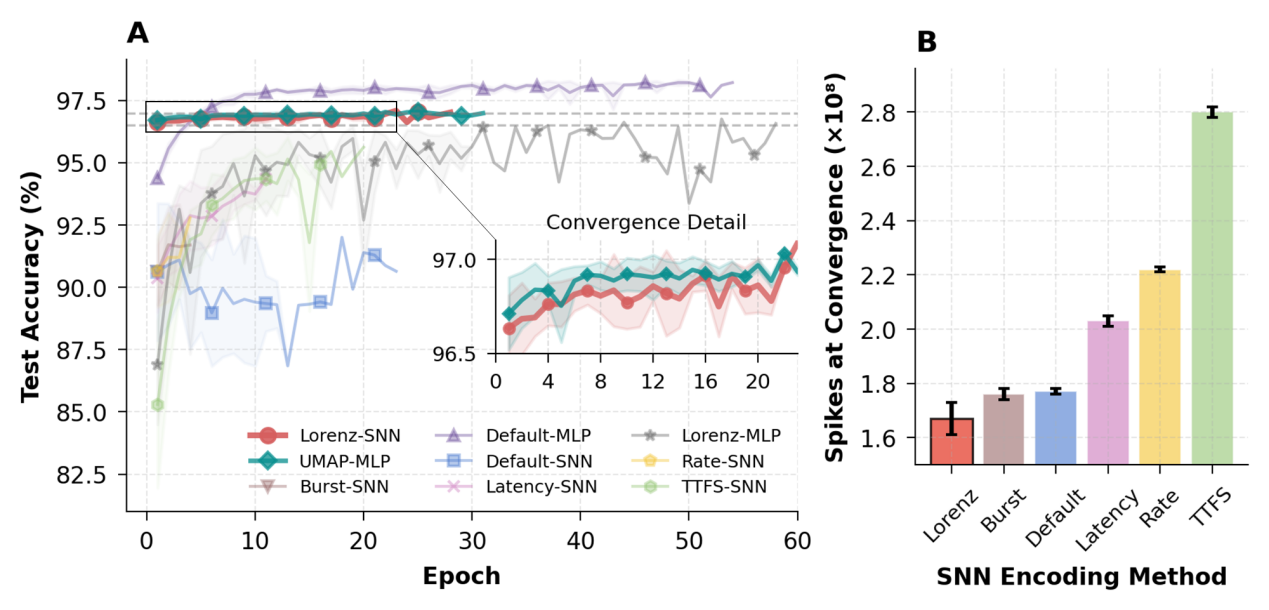}
    \caption{
        \textbf{Experiment 1: Dynamical encoding via network-out preprocessing enhances performance and stability on MNIST.} 
        \textbf{a,} The Lorenz-SNN encoding achieves accuracy and stability that rival an equivalent MLP and surpass other SNN encoding methods.  \textbf{Inset}: The learning curve for Lorenz-SNN (standard deviation $\pm 0.06\%$) demonstrates better stability compared to the significant fluctuations observed in traditional methods ($\pm 1.23\%$). 
        \textbf{b,} Lorenz encoding exhibits superior energy efficiency, requiring the fewest spikes to converge among all tested SNN methods.
    }
    \label{fig:exp1_mnist}
\end{figure}

\begin{figure}[h!]
    \centering
    \includegraphics[width=\textwidth]{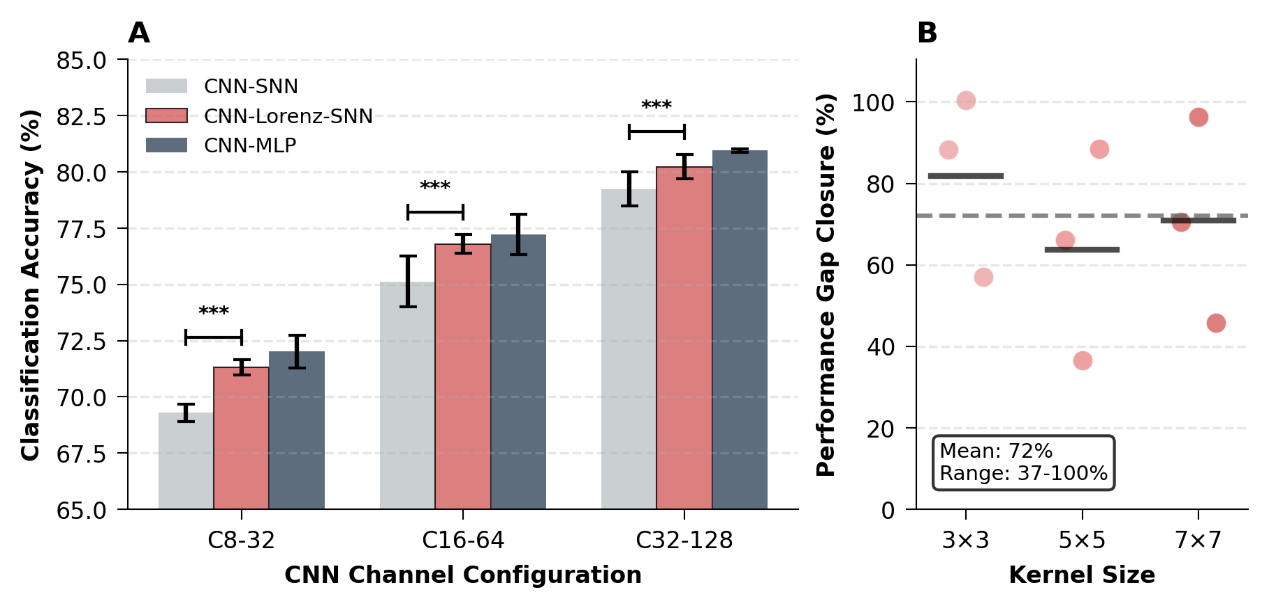}
    \caption{
        \textbf{Experiment 2: A network-in dynamical bridge consistently closes the SNN-ANN performance gap.} 
        \textbf{a,} The CNN-Lorenz-SNN model consistently outperforms the baseline CNN-SNN and approaches the performance of the CNN-MLP across all tested channel configurations. Error bars represent standard deviation; *** denotes $p < 0.001$. 
        \textbf{b,} Dynamical encoding substantially closes the performance gap between SNNs and ANNs. The gray dashed line indicates the average gap closure of 72\% across nine different CNN configurations ($3$ channel configurations $\times$ $3$ kernel sizes). A value of 100\% represents full gap closure.
    }
    \label{fig:exp2_cifar}
\end{figure}

As the dynamical encoder for both experiments, we employed the Lorenz system\cite{lorenz2017deterministic}, known for its classic 'butterfly' attractor. For clarity, we hereafter refer to the SNN using this encoding as the 'Lorenz-SNN', and the hybrid model in our second experiment as the 'CNN-Lorenz-SNN'. Detailed experimental setups are provided in Appendices~\ref{app:A} and \ref{app:B}, with results shown in Figures~\ref{fig:exp1_mnist} and \ref{fig:exp2_cifar}, respectively.

Dynamical encoding yielded significant gains across our experiments:

\begin{enumerate}
    \item \textbf{Bridging the Performance Gap:} In Experiment 1, the Lorenz-SNN achieved 96.98\% accuracy, virtually eliminating the performance gap to its MLP counterpart (96.99\%) and outperforming all other SOTA SNN encoding methods\footnote{An additional control experiment applied UMAP preprocessing to traditional SNN encoding methods. The results showed no significant performance gains, detailed in Appendix~\ref{app:a6}}. In Experiment 2, the CNN-Lorenz-SNN significantly outperformed the CNN-SNN across all channel configurations ($p < 0.001$). The gap closure analysis in Figure~\ref{fig:exp2_cifar}B demonstrates its cross-architectural consistency. Dynamical encoding reduced the performance gap between SNNs and ANNs by an average of 72\% across all tested levels of information compression.

    \item \textbf{Significant Energy Efficiency Gains:} The Lorenz-SNN demonstrated superior energy efficiency, with its total spike count ($1.67 \pm 0.06 \times 10^8$) being the lowest among all SNN encoding methods. It consumed 40.4\% fewer spikes than the TTFS-SNN ($2.80 \pm 0.02 \times 10^8$), while achieving higher performance. This highlights the positive impact of dynamical encoding on energy efficiency (Figure~\ref{fig:exp1_mnist}).

    \item \textbf{Optimization of Learning Dynamics:} The inset in Figure~\ref{fig:exp1_mnist} reveals that the Lorenz-SNN exhibited rapid convergence (comparable to the MLP) and notably high stability (standard deviation $\pm 0.06\%$) during training. In contrast, traditional SNN encoding methods (e.g., Rate) showed significant fluctuations in accuracy, with a standard deviation as high as $\pm 1.23\%$. Although some methods (e.g., Burst-SNN) had an advantage in convergence speed, they stagnated at a lower accuracy level. This indicates that dynamical encoding creates an information structure more conducive to gradient propagation, enabling the network to find optimal solutions more robustly.

    \item \textbf{Advantage of Temporal Dimension Processing:} Crucially, the encoding that yielded significant performance improvements for the SNN provided almost no benefit to the ANN (in Experiment 1, the Lorenz-MLP performed worse than both the Lorenz-SNN and the standard MLP). This disparity reveals that the spatiotemporal patterns generated by dynamical encoding are better aligned with the sequential processing mechanism of SNNs. Traditional ANNs, which primarily process static spatial representations, lack the inherent mechanism to utilize such temporal patterns and therefore cannot benefit from dynamical encoding.
\end{enumerate}

It is worth noting that the 'network-in bridging' approach also validates a modular strategy: A dynamical encoding layer can be inserted into convolutional pipelines to connect with SNNs and boost performance without costly redesigns. This offers a vital proof of concept for creating hybrid systems toward synergizing the energy efficiency of SNNs with established deep learning architectures.

Taken together, the performance gains in both experimental setups strongly indicate that dynamical alignment is a consistent, architecture-independent phenomenon: that achieving \textbf{a match between the input signal's dynamics and the network's properties is a key for efficient neural computation.}

\section{Dissecting the Dynamical Correlates of Encoding Efficiency}
\label{sec:dynamics_study}

To gain a deeper understanding of how dynamical properties affect neural encoding efficiency, we conducted a comparative study of six classic chaotic attractor systems: Lorenz, Rössler\cite{rossler1976equation}, Aizawa\cite{aizawa1982global, langford1984numerical}, Nosé-Hoover\cite{hoover1985canonical}, Sprott\cite{sprott1994some}, and Chua\cite{chua2003double}. These systems were chosen for their diversity, spanning a spectrum from strongly dissipative to nearly conservative and featuring varied geometric structures. We held the computational framework constant, employing the same network-out preprocessing setup, SNN architecture, and MNIST classification task as in the previous experiment. We quantitatively characterized the dynamics of each system using Lyapunov exponents: the maximum exponent ($\lambda_{\max}$) to measure sensitivity to initial conditions ($\lambda_{\max} > 0$ indicates chaos), and the sum of exponents ($\Sigma \lambda_i$) to measure the rate of change of phase space volume (dissipation or expansion)\cite{wolf1985determining}. Through a systematic parameter search, we aimed to identify the key factors influencing encoding efficiency. Detailed descriptions of these attractors and our methodology are provided in Appendices~\ref{app:c1} through \ref{app:c3}.

\subsection{Parameter Optimization and Performance Analysis}

\begin{figure}[h!]
    \centering
    \includegraphics[width=0.8\textwidth]{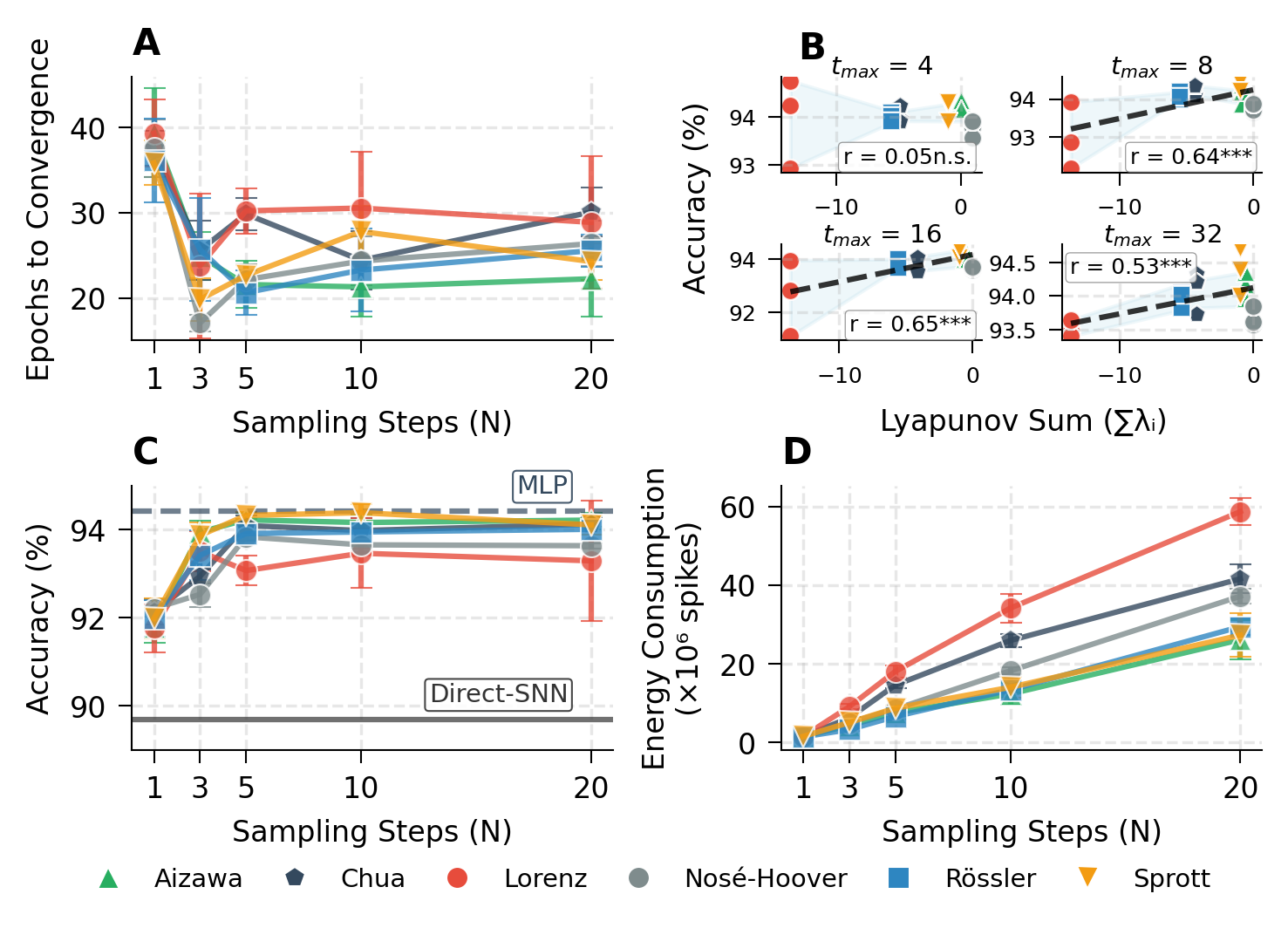}
    \caption{
        \textbf{Effect of time-embedding parameters on the efficiency of dynamical encoding in SNNs.}
        \textbf{a}, Convergence speed (epochs) as a function of sampling steps ($N$). Most systems reach optimal speed at $N=5-10$.
        \textbf{b}, Correlation between the Lyapunov sum and accuracy at different evolution timescales ($t_{\max}$). The strongest correlation is observed at $t_{\max}=8-16$.
        \textbf{c}, Accuracy as a function of sampling steps, showing the cumulative contributions of the spatial ($N=1$) and temporal ($N>1$) dimensions.
        \textbf{d}, Energy consumption (spike count) grows approximately linearly with sampling steps, but the growth rate differs significantly across systems.
    }
    \label{fig:param_search}
\end{figure}

Our parameter exploration first revealed a consistent optimal configuration for the encoding process: a maximum evolution timescale of $t_{\max}=8$ with $N=5$ sampling steps (Figures~\ref{fig:param_search}A and \ref{fig:param_search}B). The consistent optimality of $N=5$ is not a coincidence of dataset-specific hyperparameter tuning, but reflects a system fundamental requirement. This finding is strongly supported by Takens' Embedding Theorem\cite{takens2006detecting}, which establishes the conditions for accurately reconstructing a system's dynamics from a time series (see Appendix~\ref{app:c4} for theoretical discussion). For three-dimensional systems like those studied here, $N=5$ provides sufficient sampling to fully unfold the dynamics in phase space. This consistency property of dynamical encoding offers a major advantage by providing a unified foundation for our subsequent analyses and cross-task applications.

Analyzing the results of our parameter exploration revealed another key insight into the encoding's mechanism: the spatial and temporal components of the spatiotemporal patterns make distinct and additive contributions to performance. As illustrated in Figure~\ref{fig:param_search}C, a purely spatial transformation (i.e., at $N=1$) first establishes a substantial performance baseline, boosting accuracy by approximately 2\%. Extending the encoding into the temporal dimension (for $N > 1$) yields further significant gains, with an additional 2\% improvement as the number of steps reaches 5-10, ultimately approaching the performance of a conventional MLP. This two-stage improvement was consistently observed across all six chaotic systems, demonstrating the generality of this additive effect.

However, the results in Figure~\ref{fig:param_search}D reveal a crucial divergence: While all chaotic encoders yield comparable performance gains, they lead to vastly different energy consumption growth rates. This demonstrates that the dynamical properties of the encoded trajectories directly govern the energy efficiency of the subsequent neural computation.

\subsection{Dynamical Predictors of Encoding Efficiency}  

To further investigate this discrepancy and pinpoint the specific dynamical predictors of efficiency, we correlated the properties of each attractor with its corresponding performance under the optimal parameter configuration identified above ($t_{\max}=8, N=5$). We also incorporated the Active Information Storage (AIS), a measure of the mutual information between a system's present and past states, as a complementary metric for evaluating the encoded input's quality (Figure~\ref{fig:dynamics_correlation}, see Appendix~\ref{app:c5} for metric details).

\begin{figure}[h!]
    \centering
    \includegraphics[width=0.8\textwidth]{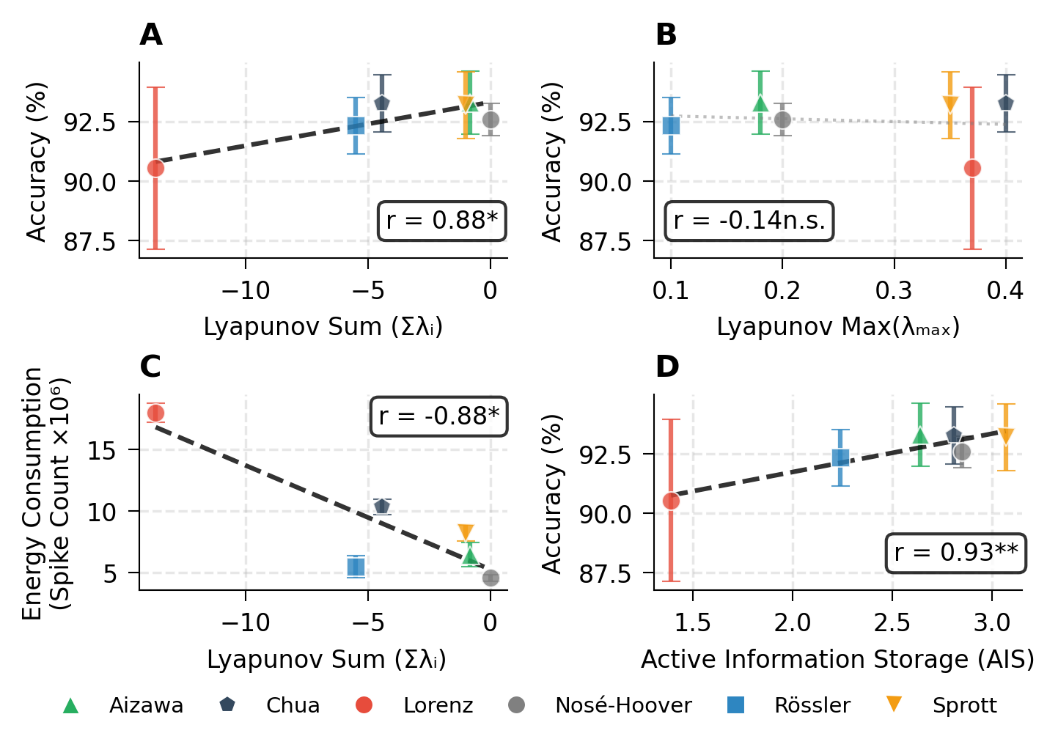}
    \caption{
        \textbf{Dynamical properties of encoded input trajectory predict neural computation performance and efficiency.}
        \textbf{a}, The Lyapunov sum ($\Sigma \lambda_i$) shows a strong positive correlation with classification accuracy. Systems with values closer to zero perform better.
        \textbf{b}, The maximum Lyapunov exponent ($\lambda_{\max}$) shows no significant correlation with accuracy across the different attractor systems.
        \textbf{c}, $\Sigma \lambda_i$ is strongly negatively correlated with energy consumption, with systems near zero achieving lower energy use.
        \textbf{d}, AIS exhibits a strong positive correlation with accuracy. 
        Error bars represent standard deviation. Colors and symbols distinguish the six chaotic attractor systems.
    }
    \label{fig:dynamics_correlation}
\end{figure}

Our analysis revealed two key predictors for encoding efficiency: The Lyapunov sum ($\Sigma \lambda_i$) proved to be a robust indicator, correlating positively with accuracy ($r=0.88, p<0.05$, Figure~\ref{fig:dynamics_correlation}A) and negatively with energy consumption ($r=-0.88, p<0.05$, Figure~\ref{fig:dynamics_correlation}C). Concurrently, AIS also exhibited a strong positive correlation with accuracy ($r=0.93, p<0.01$, Figure~\ref{fig:dynamics_correlation}D).

In contrast, the maximum Lyapunov exponent ($\lambda_{\max}$), which traditionally is the core indicator of a chaotic system, showed no statistically significant correlation with any measure of encoding efficiency. Taken together, these findings suggest that the global dynamics of phase space volume (captured by $\Sigma \lambda_i$) and the system's memory capacity (measured by AIS) are better determinants of dynamical encoding efficiency than the local rate of trajectory separation (indicated by $\lambda_{\max}$). This insight challenges the traditional focus on the 'edge of chaos' that tied primarily to $\lambda_{\max}$ and suggests that the optimal conditions for neural encoding are not confined to a single critical state, but are instead more multifaceted, involving a balance of global dynamics and information storage.

These observed monotonic relationships naturally lead to two fundamental questions: \textit{Is chaos ($\lambda_{\max}>0$) a necessary prerequisite for efficient neural encoding? Is phase space contraction ($\Sigma \lambda_i<0$) essential?} Answering these questions is impossible with classic attractor systems, where dynamical metrics and geometric structures are inherently coupled. To disentangle these factors, we need to design a unified encoder model with continuously tunable dynamics, enabling a systematic exploration of the full dynamical spectrum and providing a more rigorous test of our hypotheses.

\section{Shaping Neural Computation via Input Dynamics: Evidence for Phase Transitions and Bimodal Optimization}
\label{sec:phase_transition}

To this end, we designed a parameterized, three-dimensional dynamical system based on the classic Duffing oscillator\cite{kovacic2011duffing}. This system preserves the core nonlinear features of the Duffing oscillator while allowing for continuous control over its dynamics---from strongly expansive to strongly dissipative---by tuning a core parameter, $\delta$. The approach is designed to eliminate the confounding influence of specific attractor geometries and allow a exploration of the full spectrum of dynamical properties (see Appendix~\ref{app:d} for system and experimental design details).

\subsection{Critical Phase Transitions and Bimodal Optimization}
\label{subsec:bimodal_discovery}

\begin{figure}[h!]
    \centering
    \includegraphics[width=\textwidth]{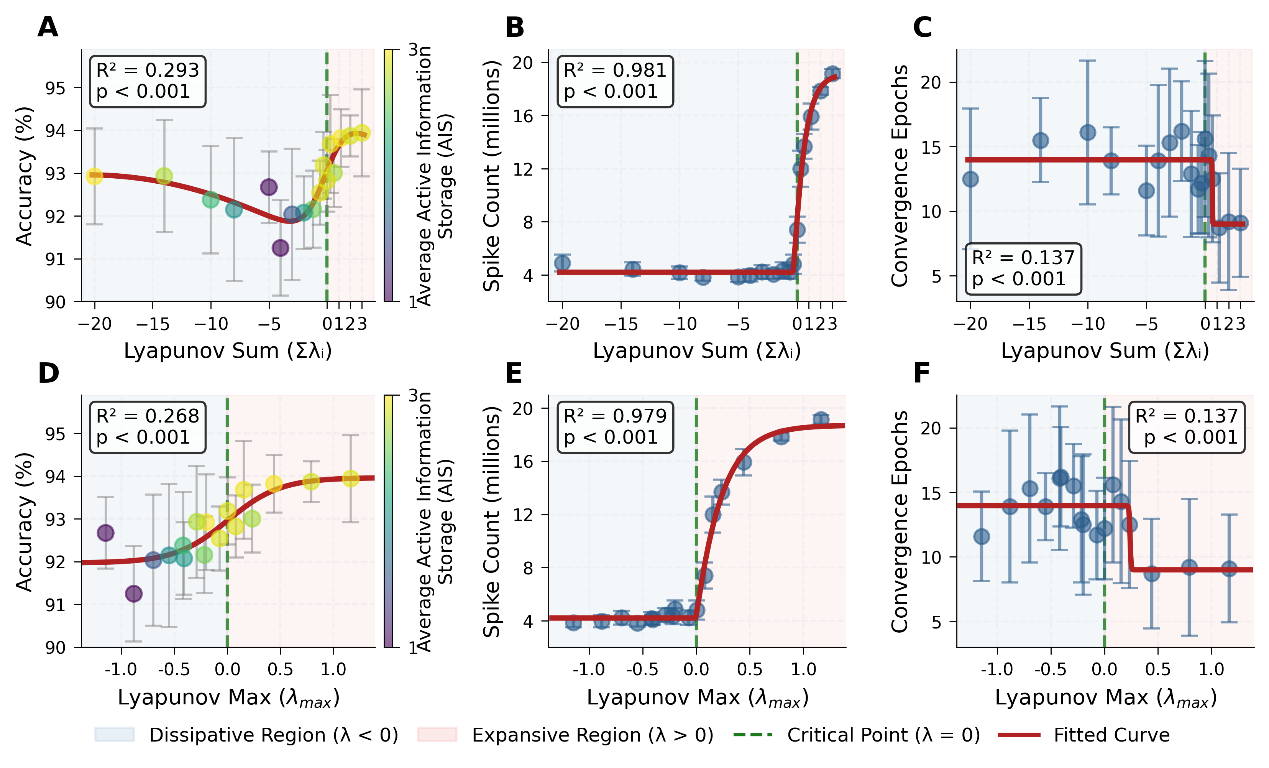}
    \caption{
        \textbf{Shaping neural computation with input dynamics: bimodal performance and phase transitions}
        \textbf{a}, Classification accuracy as a function of the Lyapunov sum ($\Sigma \lambda_i$) exhibits a bimodal distribution, with two performance peaks in the strongly dissipative and expansive regions.
        \textbf{b}, Energy consumption (spike count) undergoes a sharp phase transition near the critical point ($\Sigma \lambda_i = 0$).
        \textbf{c}, Convergence speed shows a similar step-like change across the critical point.
        \textbf{d-f}, Accuracy (\textbf{d}), energy consumption (\textbf{e}), and convergence speed (\textbf{f}) show highly consistent patterns when plotted against the maximum Lyapunov exponent ($\lambda_{\max}$). 
        Data point colors in (\textbf{a, d}) represent the intensity of AIS. Blue and pink backgrounds denote the dissipative ($\lambda < 0$) and expansive ($\lambda > 0$) dynamical regions, respectively. The green dashed line marks the critical point ($\lambda=0$), and the red solid lines are statistically significant nonlinear fits ($p < 0.001$).
    }
    \label{fig:phase_transition}
\end{figure}

Our exploration uncovered two pivotal phenomena in the subsequent neural computation: a \textbf{bimodal distribution of computational performance} and a \textbf{phase transition in energy consumption} (Figure~\ref{fig:phase_transition}).

First, Figure~\ref{fig:phase_transition}A reveals that the classification accuracy of the SNN does not vary monotonically with the encoding's dynamical parameters. Instead, two distinct high-performance regions form at opposite ends of the dynamical spectrum, separated by a 'performance valley' in the transition region. Notably, this performance valley (transition region, $\Sigma \lambda_i \approx -7 \text{ to } -3$) corresponds precisely with a significant drop in AIS (purple/blue data points, Figures~\ref{fig:phase_transition}A and D). This observation also corroborates the conclusion from Figure~\ref{fig:dynamics_correlation}D and provides an information-theoretic measure of degraded input quality in this dynamical state. The mechanistic explanation for why this region fails to support effective computation requires deeper theoretical analysis, which we provide in the next sub-section.

Concurrently, Figures~\ref{fig:phase_transition}B and \ref{fig:phase_transition}E reveal another key phenomenon: the SNN's energy consumption undergoes an abrupt transition as the encoder's dynamics cross the critical point ($\lambda = 0$). It maintains a low level of spike activity ($\sim$4M) in the dissipative region ($\lambda < 0$) but jumps fivefold ($\sim$20M) upon entering the expansive region. This phenomenon closely resembles a critical phase transition in physical systems, characterized by a discontinuous change in a system property across a critical point. This sharp transition is well-modeled by sigmoid functions, which fits the data with high fidelity ($R^2 > 0.98$). Detailed statistical analysis and further quantitative characterization of the phase transition are provided in Appendix~\ref{app:d3}.

The coexistence of these two phenomena suggests that neural computation operates in two distinct strategic modes:
\begin{itemize}
    \item \textbf{The Dissipative Mode:} In the strongly dissipative region ($\Sigma \lambda_i \ll 0$), the encoding process allows trajectories to \textbf{converge and contract}, progressively compressing input information into highly structured and stable spatiotemporal patterns. This results in an excellent trade-off, yielding high accuracy with minimal energy expense.

    \item \textbf{The Expansive Mode:} In the expansive dynamics region ($\Sigma \lambda_i > 0$), the process amplifies subtle differences in the input data, causing their corresponding trajectories to rapidly \textbf{diverge and unfold} in phase space. This generates highly discriminative and flexible representations, achieving maximal accuracy, but at a substantial energetic cost.
\end{itemize}

\subsection{Mechanisms of Bimodal Neural Computation}
\label{subsec:bimodal_mechanisms}

To understand the computational mechanisms behind this bimodality, we performed an in-depth analysis of the internal neural activity of an SNN when receiving signals encoded with different dynamics (Figure~\ref{fig:neural_dynamics}, see Appendix~\ref{app:d4} for detailed setup).

\begin{figure}[h!]
    \centering
    \includegraphics[width=\textwidth]{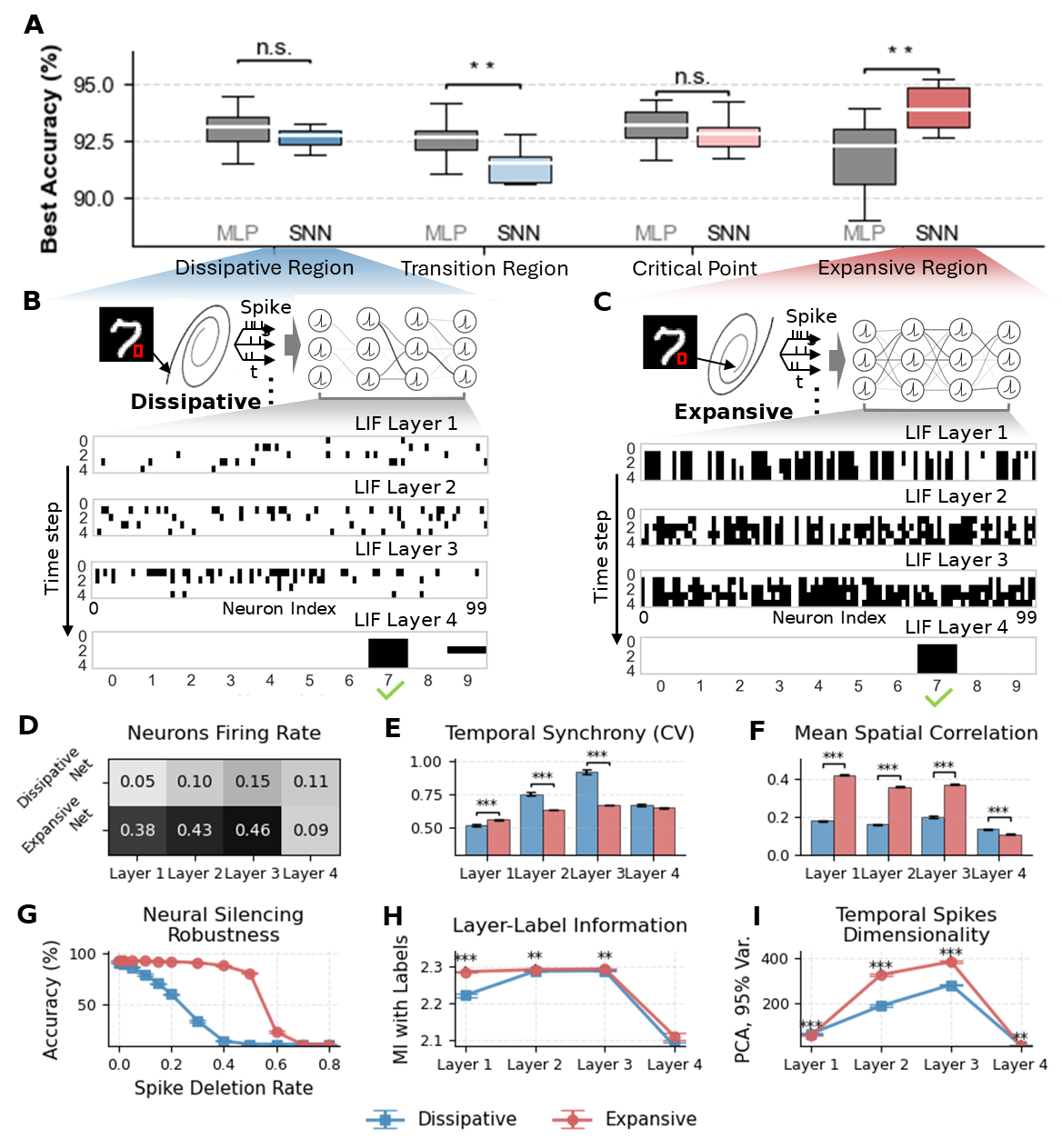}
    \caption{
        \textbf{Input dynamics shape internal network activity.}
        \textbf{a}, Accuracy comparison between MLP and SNN across four dynamical regions. Consistent with the bimodal landscape in Figure~\ref{fig:phase_transition}, the SNN significantly outperforms the MLP in the expansive region ($p < 0.01$).
        \textbf{b, c}, Raster plots of a SNN's activity in response to the same input sample (the digit '7') after being processed by dissipative (\textbf{b}) and expansive (\textbf{c}) encoding.
        \textbf{d}, Mean firing rates of neurons in each layer for both modes.
        \textbf{e}, Coefficient of variation (CV) of spike firing, where higher values indicate more precise temporal coding.
        \textbf{f}, Mean spatial correlation, indicating the degree of neuronal co-activation.
        \textbf{g}, Robustness test showing accuracy after deleting various proportions of spikes.
        \textbf{h}, Mutual information between layer representations and class labels.
        \textbf{i}, Effective dimensionality of neural representations (number of principal components needed to explain 95\% of variance).
        Statistical significance: *$p < 0.05$, **$p < 0.01$, ***$p < 0.001$, n.s. (not significant).
    }
    \label{fig:neural_dynamics}
\end{figure}

Figure~\ref{fig:neural_dynamics}A highlights a key finding: in the expansive dynamics region, the SNN significantly outperforms the identically structured ANN under the same conditions ($p < 0.01$). This result confirms that dynamical encoding can confer different computational advantages to SNNs. Further analysis of the neural activity patterns reveals the mechanistic underpinnings of this bimodality. As shown in Figures~\ref{fig:neural_dynamics}B and C, the SNN's internal computation organizes into two starkly different modes, each corresponding to one of the optimal dynamical regions:
\begin{itemize}
    \item \textbf{The Dissipative Mode - Energy-First Minimalism:} This mode produces extremely sparse and precise spike patterns, forming a \textit{sparse temporal coding}. Information is carried by a small number of neurons (low firing rate, Figure~\ref{fig:neural_dynamics}D) firing in precise temporal coordination (high temporal synchrony, Figure~\ref{fig:neural_dynamics}E), which explains its superior energy efficiency.
    \item \textbf{The Expansive Mode - Performance-First Redundancy:} This mode exhibits highly active, distributed neuron firing, forming a \textit{distributed rate coding}. Information is represented by significantly higher inter-neuronal correlations (a measure of spatial co-activation) across all layers (Figure~\ref{fig:neural_dynamics}F) and in a higher-dimensional representational space (Figures~\ref{fig:neural_dynamics}H,I). This combination of high-dimensional, distributed representation and the SNN's intrinsic temporal processing capability accounts for its superior classification performance, even surpassing the MLP.
\end{itemize}

The difference between the two modes is also evident in their robustness to neuron silencing (Figure~\ref{fig:neural_dynamics}G). The expansive mode maintains >90\% accuracy even when 40\% of spikes are deleted, whereas the dissipative mode's performance collapses with just a 10\% deletion rate, confirming a fundamental difference in information redundancy between the two strategies.

\subsubsection{Theoretical Explanatory: Timescale Alignment}
\label{subsubsec:theoretical_framework}

To explain the underlying mechanism, we applied a theoretical framework based on the established theory of neurons driven by correlated noise\cite{brunel2000dynamics, fourcaud2002dynamics} (see Appendix \ref{sec:appendix_math_model} for full details and experiment validation). Our analysis reveals that the computational mode is governed by the ratio of the input's autocorrelation time ($\tau_{corr}$) to the neuron's membrane time constant ($\tau_m$), as well as the mean input intensity generated by the dynamical encoding: When $\tau_{corr} \ll \tau_m$, as in the dissipative mode, the input appears as fast, uncorrelated fluctuations to the neuron. The membrane integrates these fluctuations efficiently, enabling a \textbf{fluctuation-driven sparse coding} strategy where brief, precise spikes carry maximum information per energy unit. Conversely, when $\tau_{corr} \approx \tau_m$, as in the expansive mode, the input's temporal correlations persist across the neuronal integration window. The neuron's membrane acts as a low-pass filter that suppresses these slow input fluctuations. Our data show the network benefit from the expansive dynamic of the input signal and compensates for this suppression by increasing its mean drive, steering a robust, \textbf{mean-driven rate coding} strategy.

Furthermore, this framework gives the performance valley a deeper mechanistic explanation: The network's processing fidelity collapses because it cannot form a coherent strategy for this information-poor input. In this transition region, while the input correlations are short enough for fluctuation-driven coding, the degraded signal structure (low AIS) provides insufficient coherent fluctuations to drive effective sparse coding, yet the network fails to support a sufficiently strong mean drive for robust mean-driven coding. The network thus results in a noisy neural code with a high CV of its firing rate, which reflects unreliable spike timing and a poor signal-to-noise ratio. This fundamentally limits the information transmission capacity within the network. Therefore, the performance valley emerges from the system's inability to compress and stabilize information (as in dissipative mode) or amplify and distinguish it (as in expansive mode). This represents a misalignment where neither optimal strategy can be effectively implemented.

Herein lies the essence of dynamical alignment: a single, fixed network can be steered into starkly different computational paradigms—from an energy-first, sparse coding strategy to a performance-first, dense coding one—governed entirely by the dynamics of its input. This strategic divergence, arising from 'software' (input dynamics) rather than 'hardware' (model structure), provides a direct mechanistic explanation for the bimodal performance landscape (Figure \ref{fig:phase_transition}). The ability to dynamically switch between computational modes is not merely an advantage of temporal processing, it confirms that a system's computational behavior is determined by achieving a dynamic, effective alignment between the information it must process and the network itself.

Furthermore, this bimodality phenomenon, together with its observed network mechanisms, shows a striking correspondence with the 'segregation-integration' state transitions observed in neuroscience\cite{li2019transitions}. The phase transition driven by the encoding dynamics in our computational model can be seen as the system's shift from an efficient, low-synchrony, modular 'segregated phase' to a high-synchrony, high-dimensional, and high-capacity 'integrated phase'. From an information-theoretic perspective, these two modes represent different Pareto-optimal solutions in the information-energy space: the dissipative mode achieves high compression, while the expansive mode preserves more relevant information (see Appendix~\ref{app:d4} for additional control experiments, including systematic evaluation of representational dimensionality, linear separability analysis, and information bottleneck trajectory tracking). This capacity for paradigm-switching, which is absent in conventional ANNs, provides a powerful new computational basis and phenomenological support for understanding the origins of the adaptability and efficiency inherent in biological systems.

\section{Universality and Scalability of Dynamical Alignment}
\label{sec:cross_domain_validation}

The preceding experiments have revealed the bimodality of dynamical alignment under controlled conditions. This section aims to demonstrate that these findings are not an isolated phenomenon but reflect a scalable and general principle of neural computation. To do so, we investigate whether the dissipative and expansive modes confer specific advantages in three distinct computational contexts: the high-capacity feature processing of deep learning, the sequential temporal integration of dynamic decision-making, and the robust information binding required for cognitive integration.

All subsequent experiments leverage the parameterized mixed oscillator system introduced in Section \ref{sec:phase_transition} as the dynamical encoder. This unified approach allows us to consistently probe the computational characteristics of both optimal modes, as well as the suboptimal 'transition' region across all validation tasks.

\subsection{Scalability in Deep Learning Architectures}
\label{subsec:scalability}

Our initial experiments demonstrated that a dynamical encoding layer could be effectively inserted into a standard convolutional pipeline to connect with SNNs and boost performance (the 'network-in bridging' approach, Section~\ref{subsec:strategy_results}). To validate the scalability and universality of this principle, we now apply it in a deep learning setting: using the complex TinyImageNet dataset\cite{deng2009imagenet} and employing a ResNet-18 model\cite{he2016deep}, pretrained on ImageNet, as a powerful feature extraction backbone.

We removed the original classification head from the pretrained ResNet-18. The high-dimensional features extracted by its backbone were then projected by a linear layer to various target dimensions (ranging from 32 to 512). Subsequently, the projected vectors were processed by one of two distinct classifier heads, each containing two hidden layers. In our proposed pipeline, a dynamical encoding layer first transformed the feature vectors into spatiotemporal trajectories, which were then processed by the SNN head. For comparison, the baseline pipeline fed the static feature vectors directly into a structurally-equivalent MLP head. Crucially, to ensure a fair comparison where the classifier's capacity scaled with the richness of the input features, the width of both the SNN and MLP classifier heads (i.e., the number of neurons in their hidden layers) was directly tied to the projection dimension (see Appendix~\ref{app:e1} for detailed setup).

\begin{figure}[h!]
    \centering
    \includegraphics[width=\textwidth]{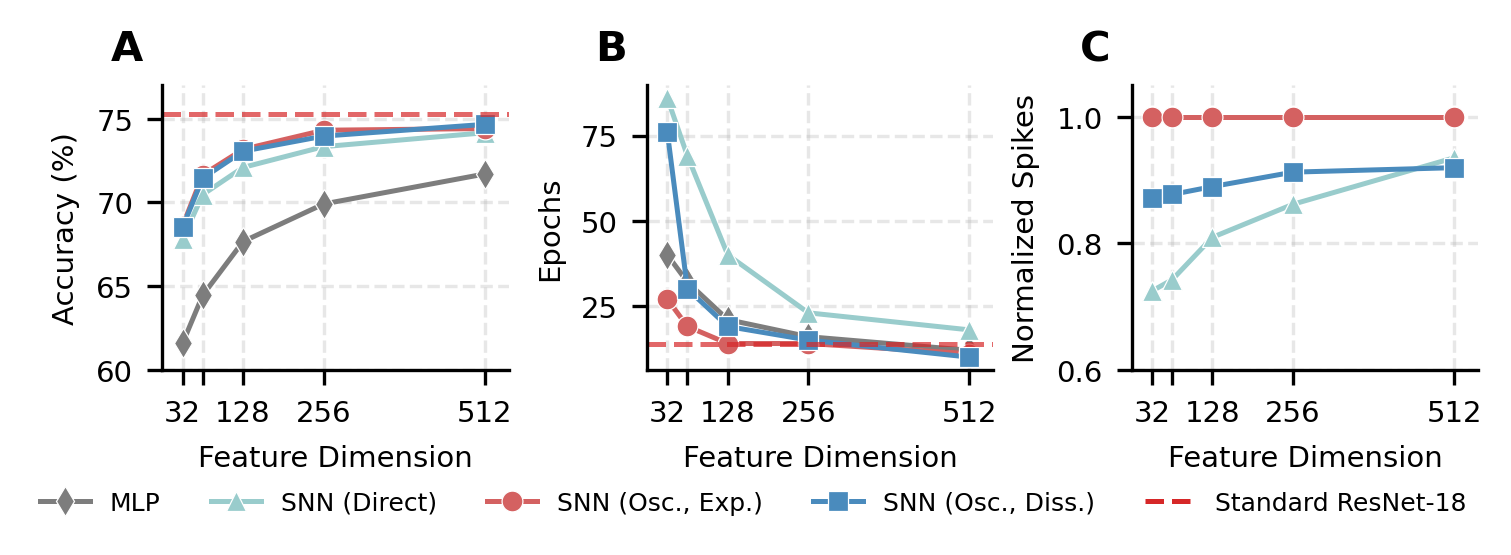}
    \caption{
        \textbf{Dynamical encoding confers a robust performance advantage on a deep residual network.}
        \textbf{a}, Across all tested feature dimensions, both dynamically encoded SNNs achieve higher classification accuracy than the identically structured MLP baseline.
        \textbf{b}, Both dynamical modes accelerate convergence compared to the baselines. The expansive mode proves to be the fastest, showing a particularly strong advantage in lower-dimensional spaces.
        \textbf{c}, The two dynamical encoding modes maintain their distinct energy efficiency profiles. This panel shows their normalized energy consumption (spike count) relative to a direct SNN, highlighting their characteristic differences across dimensions.
    }
    \label{fig:resnet_validation}
\end{figure}
In a stark reversal of their behavior in shallower networks (cf. Figure~\ref{fig:exp2_cifar}), SNNs with dynamical encoding consistently establish a stable performance lead when integrated into a deep network architecture. Specifically, across all tested dimensional configurations, the SNNs surpassed the equivalent ANNs by a non-trivial margin of 2.46\% to 6.21\% (Figure~\ref{fig:resnet_validation}). This advantage manifested differently depending on the dynamical encoded mode: the expansive mode converged up to 2-3 times faster in low-dimensional feature spaces, while the dissipative mode achieved an optimal balance of performance and energy efficiency at higher dimensions. These findings strongly suggest that the computational benefits of temporal information processing scale with the network's expressive capacity.

Furthermore, dynamical encoding demonstrates adaptability to different input dimensionalities. In low-dimensional feature spaces (32 to 128), the expansive mode holds a significant performance advantage. In high-dimensional spaces (512), however, their performances converge, both approaching the standard ResNet-18 baseline with a performance gap of less than 0.6\%. Despite this convergence in accuracy, they retain their distinct energy efficiency profiles. By demonstrating near-complete gap closure, this result opens a viable path toward integrating the energy-efficient computing paradigm of SNNs into existing deep learning workflows.

\subsection{Strategic Advantages in Dynamic Decision-Making Tasks}
\label{subsec:rl_validation}

Having validated the bimodal property in static classification tasks, we next investigated its strategic implications in a dynamic decision-making task requiring temporal information integration. We conducted a reinforcement learning task in the CartPole-v1  environment\cite{towers2024gymnasium}, which requires an agent to make continuous balancing actions based on real-time state information, a classic sequential decision-making problem that necessitates the integration of historical information\cite{barto2012neuronlike} (see Appendix~\ref{app:e2} for detailed setup).

\begin{figure}[h!]
    \centering
    \includegraphics[width=\textwidth]{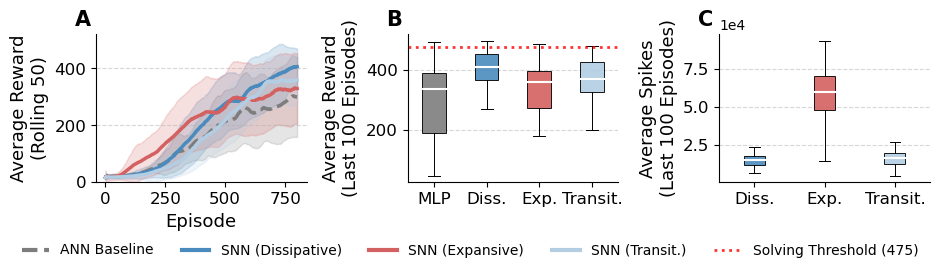}
    \caption{
        \textbf{Dynamical encoding enables superior and strategically diverse policies for SNN agents in reinforcement learning.}
        \textbf{a}, Learning curves comparing the different configurations. All dynamically encoded SNNs outperform the MLP baseline, with the expansive mode learning fastest but exhibiting the greatest volatility.
        \textbf{b}, Box plot of performance distribution for the final 100 episodes. The red dashed line indicates the solving threshold (a score of 475).
        \textbf{c}, Energy efficiency analysis showing the spike activity levels of the different modes. Experiments were based on 30 independent runs; error bars represent standard error.
    }
    \label{fig:rl_validation}
\end{figure}

Our results reveal a clear performance advantage for SNNs with dynamical encoding compared to the baseline MLP (Table~\ref{table:rl_results}): Both optimal modes (dissipative and expansive) achieved success rates ($\sim$20\%) significantly higher than the MLP baseline (6.7\%). Furthermore, the different dynamical configurations gave rise to a clear \textit{strategic divergence} in decision-making. While Figure~\ref{fig:rl_validation} illustrates the resulting differences in performance and learning dynamics, the underlying behaviors themselves were highly distinct:

\begin{itemize}
    \item \textbf{The Robust Mode (Dissipative):} Exhibited the most stable decision-making capability, achieving the highest average reward ($397.8$) with the lowest reward variance ($\pm 72.8$). This robust and reliable strategy was achieved while consuming about a quarter of the energy (spike count) of the expansive mode. Visually, this corresponded to minimal and fine-tuned corrective actions of agent to hold a stable equilibrium with subtle movements. 
    \item \textbf{The Exploratory Mode (Expansive):} Demonstrated the strongest exploratory behavior, attaining the highest success rate (23.3\%) and the fastest solving time (346 episodes), but this came with substantial reward variance ($\pm 125.7$). In rendered visualizations, this strategy manifested as larger and more frequent cart movements, reflecting an agent actively testing the boundaries of its state space.
    \item \textbf{The Transition Region:} Was unable to form an effective policy (3.3\% success rate), performing even worse than the MLP baseline and confirming the existence of the 'performance valley' in dynamic tasks.
\end{itemize}

\begin{table}[htbp]
    \centering
    \caption{Performance comparison of different network configurations in the reinforcement learning task.}
    \begin{tabular}{llccc}
        \toprule
        \textbf{Configuration} & \textbf{Success} & \textbf{Best Solving} & \textbf{Avg. Reward} & \textbf{Performance} \\
        \textbf{Type} & \textbf{Rate} & \textbf{Episode} & \textbf{(last 100 epi.)} & \textbf{Characteristics} \\
        \midrule
        SNN (Dissipative) & 20.0\% (6/30) & 664 & $397.8 \pm 72.8$ & Robust policy, high peak potential \\
        SNN (Expansive) & 23.3\% (7/30) & 346 & $328.5 \pm 125.7$ & Exploratory policy, high-risk/high-reward \\
        SNN (Transition) & 3.3\% (1/30) & 797 & $358.2 \pm 81.2$ & Ineffective policy in transition region \\
        MLP Baseline & 6.7\% (2/30) & 745 & $291.0 \pm 130.4$ & Fails to solve consistently \\
        \bottomrule
    \end{tabular}
    \label{table:rl_results}
\end{table}

The significance of this strategic divergence extends beyond mere performance enhancement, but also introduces an inherent, controllable degree of freedom for adjusting the agent's behavior. While classic RL algorithms engineer the exploration-exploitation balance with extrinsic, explicit rules (e.g., $\epsilon$-greedy\cite{sutton1998reinforcement} or the uncertainty bonus in upper confidence bound\cite{auer2002finite}), dynamical alignment offers a more fundamental approach, with an emergent mechanism: the trade-off is not a pre-programmed parameter but an inherent property, arising from the way the agent's internal state trace is sculpted by the input's temporal dynamics.

This internal 'state trace' can be viewed as an intrinsic form of memory, formed by the temporal continuity of the SNN neuron's membrane potential. It serves as the physical basis for this emergent control. Dynamical encoding guides the agent's strategy precisely by shaping the properties of this trace, and the characteristics of the two modes (Figure~\ref{fig:neural_dynamics}) provide a reasonable mechanistic explanation for the observed strategic behaviors:

\begin{itemize}
    \item The contractive dynamics of the \textbf{dissipative mode} tend to compress information and filter noise, forming stable, long-lasting state traces of critical historical states, which naturally leads to reliable and exploitative decisions.
    \item Conversely, the divergent dynamics of the \textbf{expansive mode} amplify subtle changes in recent states, creating state traces that are highly sensitive to new information, thereby driving efficient exploration.
\end{itemize}

More profoundly, the conservative/aggressive characteristics exhibited by these two encoding strategies have a deep correspondence with the exploration-exploitation trade-off, a central theme in the cognitive neuroscience of adaptive behavior and decision-making\cite{cohen2007should}. Our framework provides a more parsimonious, emergent mechanism for this trade-off. The 'state trace', flexibly shaped by input dynamics with different time scales and sensitivities, is precisely what allows the SNN to more effectively link delayed rewards at the end of an episode to the critical early actions that led to them, thereby demonstrating a performance advantage over the static MLP in this complex dynamic balancing challenge.

Next, we further explore whether this capacity for temporal information integration is also evident in a classic, biologically relevant cognitive computation: the binding of information across space and time.

\subsection{Mechanistic Advantages in a Cognitive Integration Task}
\label{subsec:cognitive_validation}

The preceding experiments have not only demonstrated a scalable temporal processing advantage but have also revealed that this advantage can manifest as the emergence of distinct, task-dependent strategies. The very nature of this advantage, which lies in its capacity to shape computation through the temporal dimension, bears a profound similarity to the dynamic computational phenomena that are ubiquitous in biological neural systems. To probe this connection further, we turn to a classic problem in cognitive neuroscience: \textbf{feature binding}. This refers to the ability to integrate separate sensory features into a unified perceptual object\cite{treisman1996binding}, which represents a core function of the brain's capacity for spatiotemporal information integration\cite{shadlen1999synchrony}. Notably, one of the central hypotheses regarding this problem in neuroscience posits that it relies on the synchronous activity of neurons in the temporal dimension.

To test the network's ability to integrate distributed information, we designed a synthetic feature binding task inspired by the cognitive challenge of avoiding 'illusory conjunctions'\cite{greff2020binding}. The task is explicitly designed to be more than a simple logical AND operation; it requires robust statistical pattern recognition in a noisy, high-dimensional space. Specifically, we embedded two distinct target feature patterns within a 1000-dimensional vector, which was then corrupted by substantial Gaussian noise ($\sigma$=0.25). This task is analogous to binding the features of an object, such as its color and shape, to form a unified percept. This setup poses a significant challenge for any connectionist system: to detect the co-occurrence of features that are high-dimensional, noisy, and intertwined. To manage complexity while preserving the core difficulty, these input vectors were first compressed to 64 dimensions using UMAP. We then applied our established methodology, comparing the performance of the SNN across different dynamical modes against an MLP baseline (see Appendix~\ref{app:e3} for details).

\begin{figure}[h!]
    \centering
    \includegraphics[width=\textwidth]{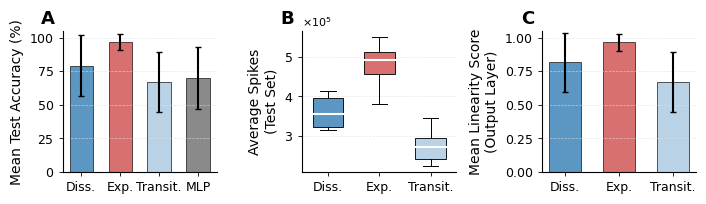}
    \caption{
        \textbf{Dynamical encoding provides a stark advantage in a cognitive binding task.}
        \textbf{a}, The expansive mode achieves a significant performance advantage in the feature-binding task. The performance of the MLP baseline is comparable to the transition region, highlighting the limitations of static networks in cognitive integration.
        \textbf{b}, Energy efficiency characteristics: the expansive mode consumes more spike resources, consistent with previous observations.
        \textbf{c}, Representation quality, as measured by a linear probe: the final hidden layer of the expansive mode produces a representation with near-perfect linear separability (0.98), indicating that its features can be easily classified. Experiments were based on 30 independent runs; error bars represent standard error.
    }
    \label{fig:cognitive_task}
\end{figure}

The results reveal a key phenomenon: in a cognitive task requiring spatiotemporal information integration, the dynamically encoded SNN shows a significant performance advantage over traditional neural networks (Figure~\ref{fig:cognitive_task}). The expansive mode achieved near-perfect linear separability (0.98) in its output layer, with an accuracy of 96.8\%, a 28.1\% improvement over the MLP baseline---a margin far exceeding the performance differences observed in static classification tasks.

This finding provides strong computational evidence and a potential mechanistic support for the classic 'binding-by-synchrony' hypothesis in cognitive neuroscience\cite{singer1995visual, singer1999neuronal}. It has long been speculated that the brain achieves information integration through the synchronization of neural activity, the physical substrate of which is often considered to be gamma oscillations\cite{buzsaki2012mechanisms}. This aligns closely with our observations: the high degree of neuronal co-activation (Figure~\ref{fig:neural_dynamics}F) and the high-dimensional representational space (Figure~\ref{fig:neural_dynamics}I) induced by the expansive mode provide the SNN with the computational resources to effectively 'bind' separate features into a unified object representation, thus explaining its superior performance.

Furthermore, the result presented in this experiment supports a unique value of spiking computation: it leverages the temporal dimension itself as a core computational resource to solve integration problems that static networks like MLPs can only struggle with in the spatial domain. When processing complex cognitive tasks, the intrinsic temporal processing capabilities and dynamical flexibility of spiking neural mechanisms enable an integration effectiveness that is difficult for traditional static computational paradigms to achieve. This suggests a broader evolutionary advantage for spiking mechanisms: beyond merely processing time-varying signals, they can transform static problems into the temporal domain, leveraging time itself as a computational resource to effectively solve tasks that prove difficult for static architectures.

Taken together, the results from these three cross-domain validations converge on a single conclusion: the optimal dynamical state for a network is not fixed, but is instead determined by the specific information processing demands of the task at hand. This work establishes that by treating the temporal dimension as a computational resource, a network gains the capacity for dynamic mode-switching. This allows the system to strategically prioritize the energy efficiency and stability of the dissipative mode for some tasks, while leveraging the representational power and sensitivity of the expansive mode for others. In the Discussion that follows, we explore how this principle of task-dependent, dynamic computation offers a unifying framework for understanding intelligence, both biological and artificial.

\section{Discussion}
\label{sec:discussion}

\subsection{The Unifying Explanatory Power of the Bimodal Property}
\label{subsec:unifying_framework}
Our research has revealed a bimodal optimization landscape in neural computation, governed by the principle of 'Dynamical Alignment'. We propose that this phenomenon is not merely a technical curiosity discovered in specific cases but a fundamental organizing principle with broad explanatory. This framework appears to offer a unified, computable explanation for a wide range of long-debated dualities across multiple scales, from neural coding and cognitive function to global brain organization. Table~\ref{table:unifying_framework} summarizes these cross-scale correspondences.

\begin{table}[h!]
\centering
\caption{The cross-scale unifying explanatory framework of the dynamical bimodal theory.}
\label{table:bimodal_dualities}
\begin{tabularx}{\textwidth}{@{} 
    >{\raggedright\arraybackslash}p{2cm}  
    >{\raggedright\arraybackslash}p{2cm}   
    >{\raggedright\arraybackslash}X          
    >{\raggedright\arraybackslash}X          
    >{\centering\arraybackslash}p{1cm}                                      
    >{\centering\arraybackslash}p{1cm} @{}}                                  
\toprule
\textbf{Level of Analysis} & \textbf{Core Duality} & \textbf{Dissipative Mode ($\Sigma \lambda_i < 0$)}: compressive, stable, energy-efficient & \textbf{Expansive Mode ($\Sigma \lambda_i > 0$)}: generative, sensitive, high-performance & \textbf{Evidence} & \textbf{Refs.} \\ 
 
\midrule
\multirow{2}{2cm}{\textbf{(1) Neural}\newline\textbf{Coding}} & Rate vs. Temporal / \newline Distributed vs. Sparse & \textbf{Temporal Code:} Generates sparse, temporally precise spike patterns, low neural redundancy. & \textbf{Rate Code:} Generates distributed, high-rate spike patterns, high neural redundancy enables a rich representational capacity. & Fig.~\ref{fig:neural_dynamics}B-D,G &\cite{panzeri2010sensory, tovee1993information} \\
\addlinespace

\multirow{3}{2cm}{\textbf{(2) Dynamics \& Cognition}} & Stability vs. Plasticity & \textbf{Stability:} Forms robust memory representations for information preservation. & \textbf{Flexibility:} Highly sensitive to initial conditions, allowing rapid state changes to encode new information. & Fig.~\ref{fig:phase_transition}D, \ref{fig:neural_dynamics}H &\cite{rabinovich2008transient, grossberg2012studies, takesian2013balancing} \\
\addlinespace
& Exploitation vs. Exploration & \textbf{Exploitation:} Compresses state space to reinforce and execute known optimal policies; low decision risk. & \textbf{Exploration:} Amplifies state differences, highly sensitive to new information and potential rewards, driving trials of new policies. & Table~\ref{table:rl_results}, Fig.~\ref{fig:rl_validation} &\cite{cohen2007should} \\
\addlinespace

\multirow{3}{2cm}{\textbf{(3) Global Brain}\newline\textbf{Organization}} & Segregation vs. Integration & \textbf{Segregated Phase:} Characterized by local, modular processing within functional subnetworks. Corresponds to low-synchrony, energy-efficient 'resting' or default states. & \textbf{Integrated Phase:} Characterized by long-range, global functional connectivity supporting high-dimensional representations. Corresponds to high-synchrony, high-energy 'task-positive' states. & Fig.~\ref{fig:phase_transition}, \ref{fig:neural_dynamics}E,F,I &\cite{tononi1994measure, li2019transitions} \\
\addlinespace
& Local vs. Global Access & \textbf{Sub-threshold / Preprocessing:} Information is processed within encapsulated, modular loops. The resulting representations are stable but remain below the threshold for global broadcast. & \textbf{Global Access / 'Ignition':} A critical transition where information 'ignites,' entering a high-dimensional, globally accessible broadcast state, available for flexible cognitive operations. & Fig.~\ref{fig:phase_transition}B,E, \ref{fig:neural_dynamics}E,F,I, \ref{fig:cognitive_task} &\cite{dehaene2001towards} \\
\bottomrule
\end{tabularx}
\label{table:unifying_framework}
\end{table}

The correspondences outlined in Table~\ref{table:unifying_framework} are not a mere list of analogies; collectively, they point to a core mechanism: a fixed neural network (space) provides the computational 'hardware', while the input signal's dynamics (time) act as the 'software' that instructs this hardware on whether to execute a 'stable' or a 'flexible' computational program. This suggests that the dualistic behaviors emerging at different levels and among various domains can be viewed as different facets of the same underlying phenomenon within our framework---an optimal adaptation to different computational demands. When a system requires stability (e.g., for memory consolidation or fine-motor control), dissipative dynamics guide the network into a 'conservative' state that contracts, denoises, and reinforces existing patterns. Conversely, when a system requires flexibility (e.g., for feature binding or environmental exploration), expansive dynamics guide the same network into an 'open' state that diverges, sensitizes, and creates new patterns.

This 'dynamic software on fixed hardware' paradigm offers more than just an explanatory perspective; it suggests a new approach to designing intelligent systems. By learning to sculpt computation mode on demand via temporal dimension, we can envision systems that—much like the brain—flexibly shift between stability and plasticity, or between rapid, efficient processing and deep, powerful computation, depending on the task at hand. 

\subsection{A New Perspective on the 'Edge of Chaos'}
\label{subsec:edge_of_chaos}

Furthermore, our findings offer an important supplement to the 'edge of chaos' theory by reframing computation not as a monolithic state, but as a complete processing pipeline that encompasses an information encoder (upstream), a computational substrate (the network), and a task-specific objective (downstream). The classic theory posits that an ideal state for computation exists at the 'edge of chaos', which is a property defined solely at the level of the substrate. Our results challenge this substrate-centric view. We find that overall computational efficiency is not achieved by simply optimizing the substrate's internal dynamics alone. Instead, it is the alignment between the upstream and the core stages that is paramount: the effectiveness of the encoder depends on its ability to generate spatiotemporal patterns that are optimally matched to the processing capabilities of the downstream network, in service of the task objective. The computational behavior can, to a large extent, be dynamically 'shaped' by the intrinsic dynamical properties of the input signal (e.g., $\Sigma \lambda_i$).

From this perspective, the computational substrate (the SNN 'hardware') may indeed possess its greatest potential for complex computation near a critical, 'edge of chaos' state. However, it is the dynamics of the upstream encoder (the 'software') that 'instructs' the network on how to utilize this potential. The dynamical optimization of neural encoding does not point to a single optimal solution, but to a richer, multi-modal landscape where energy efficiency is introduced as a key optimization dimension. The computational efficiency of biological neural systems lies precisely in their ability to flexibly switch between these two optimization targets according to task demands: when energy efficiency is paramount (e.g., maintaining basic functions), the system favors the dissipative mode; when computational power/scale is required (e.g., for complex cognitive tasks), the system switches to the expansive mode.

Our findings suggest, therefore, that the future of intelligent systems may not lie in designing ever-more-complex static architectures, but in creating \textit{dynamic intelligent systems}. Such systems would be engineered to understand and harness the 'dynamic software' of information itself. This paradigm shift not only offers a path toward high-performance, energy-efficient neuromorphic computing but also provides a unifying framework for explaining the diverse dynamics observed in biological neural systems\cite{legenstein2007edge, toyoizumi2011beyond}. By learning to sculpt computation on fixed hardware through dynamics, we may take a crucial step away from today's ANNs and toward an AI that captures the true essence of its biological inspiration: adaptability.

\section{Conclusion}
\label{sec:conclusion}

This research introduces and validates 'Dynamical Alignment' as a powerful framework for neural computation. We have shown that by shaping the temporal dynamics of an input signal, a neural network's computational state can be steered across a bimodal landscape, revealing two distinct and optimal processing regimes. The discovery of this landscape and its governing factors: primarily the rate of phase space change ($\Sigma \lambda_i$) rather than chaotic sensitivity ($\lambda_{\max}$), offers a novel perspective on the long-standing performance paradox of SNNs. This framework also complements the classic 'edge of chaos' theory by introducing energy efficiency as a critical, parallel optimization axis. It demonstrates that optimal computation arises from the dynamic alignment to meet specific task demands.

More profoundly, this principle demonstrates that computational modes can be dynamically 'sculpted' on fixed network 'hardware' by using input dynamics as a form of 'software'. This perspective offers a new, parsimonious framework for understanding how biological systems might achieve their profound adaptability. For the future of AI, it points toward a potential paradigm shift: away from the search for optimal static architectures, and toward the creation of dynamic intelligent systems that master the art of shaping computation through the dimension of time.

\section*{Acknowledgments}
The author would like to express his sincere gratitude to Prof. Philipp Geyer for his invaluable support during the author's doctoral research at Leibniz University Hannover and Technical University Berlin, which laid the foundation for this work, and for the meaningful discussions with Junyi Jiang. The author thanks the Georg Nemetschek Institute and the Munich Data Science Institute at the Technical University of Munich for the support provided during the final stages of this work.

The source code and experimental data for this study are publicly available on GitHub at \url{https://github.com/chenxiachan/Dynamical_Alignment}.

\printbibliography[title={References}]

\appendix
\section*{Appendix}
\section{Experimental Setup for Network-Out Preprocessing}
\label{app:A}
\begin{refsection} 
\setappendixprefix{A} 

This experiment is designed to test the core hypothesis: the efficiency of neural computation is significantly enhanced when the dynamical properties of the input signal are effectively aligned with the network's intrinsic dynamics. The specific experiment implementation and technical details are described below.

\subsection{Data and Preprocessing}
\label{app:a1}
The MNIST handwritten digit dataset was used as the primary evaluation benchmark\cite{lecun1998mnist}. The dataset consists of 60,000 training images and 10,000 testing images, each being a 28*28 pixel grayscale image. All images were normalized to the [0,1] range before encoding.

Dynamical encoding expands input features into a higher-dimensional spatiotemporal representation. To ensure a fair comparison and demonstrate that the observed performance gains are due to the quality of the temporal code rather than the confounding effect of dimensionality expansion, we first projected the raw image data into a common low-dimensional space using UMAP\cite{mcinnes2018umap}.

\begin{enumerate}
    \item The original 784-dimensional MNIST images were reduced to a 7-dimensional space.
    \item The UMAP parameters were set as default: \texttt{n\_neighbors=15, min\_dist=0.1, metric='euclidean'}.
\end{enumerate}

\subsection{Network Architecture}
\label{app:a2}
To ensure fairness, we used a consistent network architecture across all comparative experiments:
\begin{itemize}
    \item \textbf{Input Layer:} Varies by encoding method (784 neurons for standard input; 7 neurons after UMAP reduction).
    \item \textbf{Hidden Layers:} Three fully-connected hidden layers, each with 500 neurons.
    \item \textbf{Output Layer:} 10 neurons, corresponding to the 10 digit classes.
\end{itemize}
For the SNN, we used Leaky Integrate-and-Fire (LIF) neurons with learnable parameters, implemented via the \texttt{snn.Leaky} function from the SNNTorch library\cite{eshraghian2023training_snntorch}. Key parameters include:
\begin{itemize}
    \item Membrane potential decay factor ($\beta = 0.95$, learnable).
    \item Learnable firing threshold.
    \item Soft reset mechanism.
\end{itemize}
For the ANN baseline model, an MLP with an identical layer structure was implemented using PyTorch's \texttt{nn.Linear} module\cite{paszke2019pytorch}, but with ReLU activation functions replacing the LIF neurons. To validate the gains from dynamical alignment, we benchmarked our proposed encoding method against two sets of baselines. For the SNN, we compared it against six mainstream encoding methods. For the MLP, we established three distinct benchmarks: one with UMAP reduction followed by Lorenz encoding, one with UMAP reduction only, and one using the raw flattened input.

\subsection{Encoding Implementation}
\label{app:a3}
The Lorenz-SNN encoding process consists of the following steps:
\begin{enumerate}
    \item \textbf{Initialization:} Each data point from the UMAP-reduced feature vector $\mathbf{x}$ was used as an initial condition for the Lorenz system. The system's initial state $(x_0, y_0, z_0)$ was set as:
    \begin{align*}
        x_0 &= \mathbf{x} \\
        y_0 &= 0.2 \cdot \mathbf{x} \\
        z_0 &= -\mathbf{x}
    \end{align*}
    This setup ensures that different features generate distinct trajectories on the Lorenz attractor, thus preserving their discriminability.

    \item \textbf{Integration:} The Lorenz differential equations were solved:
    \begin{align*}
        \frac{dx}{dt} &= \sigma(y-x) \\
        \frac{dy}{dt} &= x(\rho-z) - y \\
        \frac{dz}{dt} &= xy - \beta z
    \end{align*}
    where $\sigma=10$, $\rho=28$, $\beta=2.667$, and the time step $dt=0.01$. To ensure numerical stability, we used the fourth-order Runge-Kutta (RK4) method with a resolution of $h=0.01$:
    \begin{align*}
        \mathbf{k}_1 &= h \cdot f(t_n, \mathbf{y}_n) \\
        \mathbf{k}_2 &= h \cdot f(t_n + h/2, \mathbf{y}_n + \mathbf{k}_1/2) \\
        \mathbf{k}_3 &= h \cdot f(t_n + h/2, \mathbf{y}_n + \mathbf{k}_2/2) \\
        \mathbf{k}_4 &= h \cdot f(t_n + h, \mathbf{y}_n + \mathbf{k}_3) \\
        \mathbf{y}_{n+1} &= \mathbf{y}_n + \frac{1}{6}(\mathbf{k}_1 + 2\mathbf{k}_2 + 2\mathbf{k}_3 + \mathbf{k}_4)
    \end{align*}
    where $f(t, \mathbf{y})$ is the right-hand side of the Lorenz system and $y(t)=[x(t),y(t),z(t)]^T$ is the state vector. RK4's fourth-order precision ensures the accurate generation of chaotic trajectories.
    
    \item \textbf{Sampling:} Trajectories were sampled at equal intervals $N=5$ within the time frame $t_{\max}=2$.
    
    \item \textbf{Conversion:} The three spatial coordinates of the trajectory were concatenated and directly fed as the input current to the first SNN layer.
\end{enumerate}

For systematic comparison, six mainstream SNN encoding methods were also implemented:
\begin{itemize}
    \item \textbf{Default Encoding:} Raw pixel intensities are directly used as input currents to the neurons without temporal conversion. The 28*28 MNIST image is flattened into a 784-dimensional vector and fed sequentially to the first layer.
    \item \textbf{Rate Coding}\cite{eshraghian2023training, diehl2015unsupervised}: Input intensities are linearly mapped to spiking probabilities. At each time step, spike generation follows a Bernoulli distribution where the probability equals the normalized input intensity. Implemented via \texttt{spikegen.rate} in SNNTorch\cite{eshraghian2023training_snntorch}.
    \item \textbf{Latency Coding}\cite{eshraghian2023training}: Higher input values lead to earlier first-spike times. The maximum input value corresponds to zero delay, with spike times increasing linearly as input values decrease. Implemented via \texttt{spikegen.latency} in SNNTorch\cite{eshraghian2023training_snntorch}.
    \item \textbf{Phase Coding}\cite{montemurro2008phase}: The input value determines the relative phase of a spike within a time window. The phase is calculated as $\phi = \text{data} \times 2\pi$, and a spike is generated when $\sin(\phi - t) \geq 0$. This method can encode multiple values within a single cycle.
    \item \textbf{Time-to-First-Spike (TTFS) Coding}\cite{rueckauer2018conversion}: An accumulator for each pixel, initialized to the input intensity, fires a spike when its value exceeds a threshold (e.g., 0.1), after which the threshold value is subtracted. This process repeats until the accumulator value is below the threshold.
    \item \textbf{Delta Coding}\cite{eshraghian2023training}: Encodes the change in input signals. It computes the difference between adjacent pixels, and a spike is generated if the difference exceeds a set threshold. Implemented via \texttt{spikegen.delta} in SNNTorch\cite{eshraghian2023training_snntorch}.
    \item \textbf{Burst Coding}\cite{park2019fast}: Simulates bursting behavior using an adaptive threshold. The threshold is modulated by a \texttt{burs\_function} that decays after a spike (with factor $\beta=0.95$) and resets if no spike is fired, mimicking neuronal adaptation.
\end{itemize}

\subsection{Training Details}
\label{app:a4}
All networks were trained under a unified protocol for fair comparison. We configured the training with the Adam optimizer at a learning rate of $5 \times 10^{-5}$ and a batch size of 32. An early stopping strategy prevented overfitting by terminating training if validation accuracy failed to improve for 5 consecutive epochs (with a 500-epoch maximum). For the loss function, we used cross-entropy, which was calculated over the membrane potentials at all time steps for SNNs:
$$
\mathcal{L} = \frac{1}{T} \sum_{t=1}^{T} \text{CrossEntropy}(U_{\text{output}}^{(t)}, y)
$$
where $U_{\text{output}}^{(t)}$ is the membrane potential of the output layer at time step $t$, and $y$ is the ground-truth label.

\subsection{Evaluation Metrics and Results}
\label{app:a5}
The primary evaluation metrics included classification accuracy, convergence speed, and energy efficiency. For SNNs, the predicted class was determined by the accumulated spike counts over all time steps:
$$
\hat{y} = \underset{j}{\mathrm{argmax}} \sum_{t=1}^{T} S_{j}^{(t)}
$$
where $S_{j}^{(t)}$ is the spike output of the $j$-th output neuron at time step $t$. Convergence speed was defined as the number of epochs required to reach 90\% of the maximum achieved accuracy. Energy efficiency for SNNs was evaluated by the total spike count at convergence, summed across all layers and time steps:
$$
E = \sum_{l=1}^{L} \sum_{t=1}^{T} \sum_{n=1}^{N_l} S_{l,n}^{(t)}
$$
The convergence point, $\text{epoch}_{\text{conv}}$, was defined as the first epoch where this 90\% threshold was met: $\text{epoch}_{\text{conv}} = \min \{ e : \text{Acc}_e \geq 0.9 \times \max(\text{Acc}) \}$. All experiments were repeated for 10 independent runs with different random seeds. Results are reported as mean and standard deviation, presented in Table~\ref{table:appendix_a_results}.

\begin{table}[h!]
\centering
\caption{Performance comparison of dynamical spatiotemporal encoding and traditional methods on the MNIST dataset, showing how Lorenz-SNN achieves both high accuracy and low energy consumption.}
\label{table:appendix_a_results}
\begin{tabular}{@{}lccc@{}}
\toprule
\textbf{Encoding Method} & \textbf{Convergence (Epoch)} & \textbf{Spikes at Convergence ($10^8$)} & \textbf{Accuracy (\%)} \\ \midrule
Default-SNN & $4.5 \pm 5.0$ & $1.77 \pm 0.01$ & $92.32 \pm 1.23$ \\
Rate-SNN & $2.4 \pm 1.1$ & $2.22 \pm 0.01$ & $91.62 \pm 1.23$ \\
Phase-SNN & $8.1 \pm 3.4$ & $1.87 \pm 0.02$ & $90.13 \pm 1.40$ \\
Latency-SNN & $7.8 \pm 2.2$ & $2.03 \pm 0.02$ & $93.86 \pm 0.88$ \\
TTFS-SNN & $12.7 \pm 3.4$ & $2.80 \pm 0.02$ & $95.08 \pm 0.89$ \\
Delta-SNN & $4.2 \pm 1.8$ & $2.02 \pm 0.03$ & $85.15 \pm 2.38$ \\
Burst-SNN & $2.1 \pm 1.0$ & $1.76 \pm 0.02$ & $91.54 \pm 1.48$ \\
\textbf{Lorenz-SNN}\textsuperscript{a} & $\mathbf{11.6 \pm 5.7}$ & $\mathbf{1.67 \pm 0.06}$ & $\mathbf{96.98 \pm 0.06}$ \\
\midrule
UMAP-MLP\textsuperscript{b} & $17.4 \pm 12.8$ & --- & $96.99 \pm 0.05$ \\
Lorenz-MLP\textsuperscript{c} & $12.6 \pm 6.6$ & --- & $96.36 \pm 0.16$ \\
Default-MLP\textsuperscript{d} & $30.3 \pm 11.7$ & --- & $98.16 \pm 0.08$ \\ \bottomrule
\end{tabular}
\flushleft
\textsuperscript{a} UMAP applied to reduce data dimension from 784 to 7 before Lorenz transformation. \\
\textsuperscript{b} Only UMAP applied to the data without Lorenz transformation. \\
\textsuperscript{c} Same as in \textsuperscript{a}, but the time dimension is flattened for MLP input. \\
\textsuperscript{d} No specific encoding methods applied.
\end{table}

\subsection{UMAP Preprocessing on Other SNN Encodings}
\label{app:a6}
To rigorously test whether the performance gains observed with our Lorenz-SNN were solely attributable to UMAP dimensionality reduction, a supplementary control experiment was conducted. This experiment applied the identical UMAP preprocessing pipeline (reducing data from 784 to 7 dimensions) to a set of six conventional SNN encoding methods, as detailed in Section \ref{app:a3}. The objective was to isolate the effect of feature pre-conditioning from the dynamical encoding mechanism itself.

The SNN architecture and training protocol remained consistent with the main Experiment 1 for a fair comparison. The UMAP-reduced data was first normalized to the [0,1] range to be compatible with the input requirements of these traditional encoding methods.

The results, as summarized in Table~\ref{table:appendix_a6_results}, showed no performance gains for these UMAP-preprocessed SNNs. In fact, their classification accuracy all turn lower and did not approach the performance of the Lorenz-SNN or the MLP baselines.

\begin{table}[h!]
\centering
\caption{Performance comparison of traditional SNN encoding methods with UMAP preprocessing.}
\label{table:appendix_a6_results}
\begin{tabular}{@{}lccc@{}}
\toprule
\textbf{Encoding Method} & \textbf{Convergence (Epoch)} & \textbf{Spikes at Convergence ($10^8$)} & \textbf{Accuracy (\%)} \\ \midrule
UMAP-Rate-SNN & $1.0 \pm 0.0$ & $1.39 \pm 0.08$ & $48.00 \pm 2.47$ \\
UMAP-Phase-SNN & $1.2 \pm 0.4$ & $1.09 \pm 0.11$ & $88.83 \pm 1.76$ \\
UMAP-Latency-SNN & $9.8 \pm 3.2$ & $0.62 \pm 0.07$ & $35.67 \pm 5.57$ \\
UMAP-TTFS-SNN & $0.6 \pm 0.5$ & $0.83 \pm 0.14$ & $20.54 \pm 8.83$ \\
UMAP-Burst-SNN & $4.4 \pm 2.3$ & $2.20 \pm 0.05$ & $84.19 \pm 8.89$ \\
\bottomrule
\end{tabular}
\end{table}

\printbibliography[heading=subbibliography, title={References for Appendix A}, env=appendixbib]
\end{refsection} 

\begin{refsection} 
\setappendixprefix{B} 

\section{Experimental Setup for Network-In Bridging}
\label{app:B}

This experiment, the second phase of our validation, explores the feasibility of 'network-in' dynamical bridging. Unlike the 'network-out' preprocessing in Appendix ~\ref{app:A}, this approach integrates a Lorenz transformation layer directly within the network. The goal is to validate a core hypothesis: that this dynamical bridge can significantly enhance SNN performance without requiring costly redesigns of existing deep learning architectures.

\subsection{Data and Preprocessing}
\label{app:b1}
This experiment uses the CIFAR-10 dataset\cite{krizhevsky2009learning} as the evaluation benchmark. The dataset contains 50,000 training images and 10,000 testing images, each being a $32 \times 32$ pixel RGB color image across 10 classes. Data preprocessing follows standard deep learning best practices: random cropping (with a padding of 4), random horizontal flipping, and normalization using the statistical values of CIFAR-10 (mean: (0.4914, 0.4822, 0.4465), standard deviation: (0.2470, 0.2435, 0.2616)). The batch size is set to 128 to balance memory efficiency and gradient estimation stability.

\subsection{Network Architecture}
\label{app:b2}

\subsubsection{CNN-Lorenz-SNN (CLSNN) Architecture}
This architecture, representing the dynamical bridging solution proposed in this study, consists of three main components:
\begin{itemize}
    \item \textbf{CNN Feature Extraction Part:} A two-layer convolutional structure. \texttt{conv1} and \texttt{conv2} use the specified channel and kernel size configurations, each followed by a ReLU activation and a $2 \times 2$ max-pooling layer. The output from the convolutional layers is globally flattened and then projected down to a \texttt{chaos\_dim=32} dimension via a projection layer \texttt{proj}. A \texttt{tanh} activation function is applied for feature normalization.
    \item \textbf{Lorenz Transformation Part:} The \texttt{chaos\_dim}-dimensional feature vector is input to a \texttt{LorenzTransformRK4} layer, which outputs a spatiotemporal trajectory sequence of shape \texttt{[batch\_size, num\_steps, chaos\_dim * 3]}. This sequence contains the evolutionary information of the original features within the chaotic dynamical system. The Lorenz transformation layer implements the same RK4 numerical integration method as in Appendix~\ref{app:A}, using each projected feature from the convolutional layer as an initial condition for the Lorenz system. The key role of this chaotic bridge is to re-encode convolutional features into a spatiotemporal representation suitable for SNN processing, with \texttt{num\_steps=5}.
    \item \textbf{SNN Processing Part:} Two layers of Leaky Integrate-and-Fire (LIF) neurons (\texttt{lif1}, \texttt{lif2}) process the Lorenz trajectories timestep by timestep. The membrane potential decay factor is $\beta=0.95$, and a \texttt{fast\_sigmoid} surrogate gradient function is used to support backpropagation. An output layer \texttt{fc\_out} maps the final membrane potential to the 10 classes.
\end{itemize}

\subsubsection{Baseline Model Architectures}
\begin{itemize}
    \item \textbf{CNN-SNN (BasicCSNN):} This model maintains the same CNN and SNN architectures as the CLSNN but removes the Lorenz transformation layer. The same CNN features are used as input to the SNN at every timestep, simulating a traditional static-feature-to-spike conversion.
    \item \textbf{CNN-ANN (BaseCNN):} This model uses the same CNN architecture but replaces the SNN component with a two-layer fully connected network (\texttt{fc1}, \texttt{fc2}), with each layer followed by a ReLU activation. This architecture serves as the upper-bound performance benchmark, representing the optimal performance of traditional deep learning with the same feature extraction capability.
\end{itemize}
The CNN was implemented using the \texttt{nn.Conv} module from the PyTorch library\cite{paszke2019pytorch}. The SNN implementation is identical to that described in Appendix~\ref{app:A}.

\subsection{Parameter Design of the Configuration Space}
\label{app:b3}
The experiment systematically explores the impact of CNN architecture parameters on the effectiveness of dynamical encoding by designing 9 configuration combinations that cover key architectural dimensions:
\begin{itemize}
    \item \textbf{Channel Configuration:} Three levels of feature extraction capability.
    \begin{itemize}
        \item C8-32: Lightweight configuration (input 3 $\rightarrow$ layer-1 8 $\rightarrow$ layer-2 32 channels)
        \item C16-64: Medium configuration (input 3 $\rightarrow$ layer-1 16 $\rightarrow$ layer-2 64 channels)
        \item C32-128: Heavy configuration (input 3 $\rightarrow$ layer-1 32 $\rightarrow$ layer-2 128 channels)
    \end{itemize}
    \item \textbf{Kernel Size:} Three different spatial receptive fields.
    \begin{itemize}
        \item $3 \times 3$: Standard small kernel for fine-grained feature extraction.
        \item $5 \times 5$: Medium kernel, balancing receptive field and computational complexity.
        \item $7 \times 7$: Large kernel for aggregating features over a larger area.
    \end{itemize}
\end{itemize}
All convolutional layers use the same padding strategy (padding=2) to ensure consistent feature map dimensions, and a max-pooling layer (kernel\_size=2) is used for down-sampling.

\subsection{Training Details}
\label{app:b4}
All models were trained under a unified protocol to ensure fair comparison. The Adam optimizer was used with a learning rate of $lr=5 \times 10^{-4}$ and a weight decay of $1 \times 10^{-4}$ to control for overfitting. A cosine annealing strategy (\texttt{CosineAnnealingLR}) was employed for the learning rate schedule to ensure fine-tuning in the later stages of training. The training was set for 50 epochs, a duration that balances sufficient convergence with computational resources. 

\subsection{Evaluation Metrics and Results}
\label{app:b5}
To assess the method's robustness across different model complexities, we evaluated the cross-architecture consistency of the dynamical encoding effect by calculating the mean and standard deviation of the gap closure for all 9 configurations. The results are shown in Table~\ref{table:b2}, which correspond to the data points within the same region in Figure~\ref{fig:exp2_cifar}.

\begin{table}[h!]
\centering
\caption{Average performance comparison of different models across various configurations.}
\label{table:b2}
\begin{tabular}{@{}lcccc@{}}
\toprule
\textbf{Configuration} & \textbf{CNN-SNN Acc. (\%)} & \textbf{CNN-Lorenz-SNN Acc. (\%)} & \textbf{CNN-ANN Acc. (\%)} & \textbf{Gap Closure (\%)} \\ \midrule
C8-32 & $69.3 \pm 0.4$ & $71.3 \pm 0.3$ & $72.0 \pm 0.7$ & +74.0 \\
C16-64 & $75.1 \pm 1.1$ & $76.8 \pm 0.4$ & $77.2 \pm 0.9$ & +79.2 \\
C32-128 & $79.2 \pm 0.8$ & $80.2 \pm 0.6$ & $80.9 \pm 0.1$ & +58.2 \\ \bottomrule
\end{tabular}
\end{table}

The Gap Closure is defined as:
\begin{align*}
\text{Gap Closure} = \frac{\text{Acc}_{\text{CNN-Lorenz-SNN}} - \text{Acc}_{\text{CNN-SNN}}}{\text{Acc}_{\text{CNN-ANN}} - \text{Acc}_{\text{CNN-SNN}}} \times 100\%
\end{align*}

This metric reflects the extent to which the CNN-Lorenz-SNN model narrows the performance gap between the CNN-SNN and the CNN-ANN. A value of 100\% indicates the gap is fully closed, 0\% indicates no improvement, and a negative value indicates a performance decrease. 

\printbibliography[title={References for Appendix B}, heading=subbibliography, env=appendixbib]
\end{refsection} 


\section{Comparative Study of Chaotic Attractors}
\label{app:C}
\begin{refsection}
\setappendixprefix{C}

\subsection{Selection of Chaotic Systems}
\label{app:c1}
To gain a deeper understanding of the dynamical encoding and to optimize its parameterization, we evaluated six classic chaotic attractor systems as neural encoding mechanisms:

\begin{enumerate}
    \item \textbf{Lorenz System\cite{lorenz2017deterministic}:} The most common chaotic system, forming a butterfly-shaped attractor, characterized by strong chaos, strong dissipation, and rapid phase space contraction.
    \begin{align*}
        \frac{dx}{dt} &= \sigma(y-x) \\
        \frac{dy}{dt} &= x(\rho-z)-y \\
        \frac{dz}{dt} &= xy-\beta z
    \end{align*}
    Parameters: $\sigma=10, \rho=28, \beta=2.667$.

    \item \textbf{Rössler System\cite{rossler1976equation}:} A single-scroll attractor with simpler dynamics than Lorenz, characterized by weak chaos, moderate dissipation, and a larger phase space volume.
    \begin{align*}
        \frac{dx}{dt} &= -y-z \\
        \frac{dy}{dt} &= x+ay \\
        \frac{dz}{dt} &= b+z(x-c)
    \end{align*}
    Parameters: $a=0.2, b=0.2, c=5.7$.

    \item \textbf{Aizawa System\cite{aizawa1982global, langford1984numerical}:} A complex toroidal attractor with a multi-vortex structure, characterized by moderate chaos and weak dissipation.
    \begin{align*}
        \frac{dx}{dt} &= (z-\beta)x - \delta y \\
        \frac{dy}{dt} &= \delta x + (z-\beta)y \\
        \frac{dz}{dt} &= \gamma + \alpha z - \frac{z^3}{3} - (x^2+y^2)(1+\epsilon z) + \zeta z x^3
    \end{align*}
    Parameters: $\alpha=0.95, \beta=0.7, \gamma=0.6, \delta=3.5, \epsilon=0.25, \zeta=0.1$.

    \item \textbf{Nosé-Hoover System\cite{hoover1985canonical}:} Represents thermostated chaotic oscillations, characterized by weak chaos, near-conservative behavior, and slow phase space volume evolution.
    \begin{align*}
        \frac{dx}{dt} &= y \\
        \frac{dy}{dt} &= -x - yz \\
        \frac{dz}{dt} &= y^2 - \alpha
    \end{align*}
    Parameters: $\alpha=1.0$.

    \item \textbf{Sprott Case C System\cite{sprott1994some}:} A simple algebraic chaotic system characterized by weak chaos and weak dissipation.
    \begin{align*}
        \frac{dx}{dt} &= yz \\
        \frac{dy}{dt} &= x-y \\
        \frac{dz}{dt} &= 1-ax^2
    \end{align*}
    Parameters: $a=3.0$.

    \item \textbf{Chua's Circuit System\cite{chua2003double}:} The first experimentally verified chaotic circuit system, featuring a double-scroll structure with strong chaos and moderate dissipation.
    \begin{align*}
        \frac{dx}{dt} &= \alpha(y-x-h(x)) \\
        \frac{dy}{dt} &= x-y+z \\
        \frac{dz}{dt} &= -\beta y - \gamma z
    \end{align*}
    where $h(x)$ is the nonlinear function of the Chua diode:
    \[
    h(x) = m_1 x + \frac{1}{2}(m_0-m_1)(|x+1|-|x-1|)
    \]
    Parameters: $\alpha=15.6, \beta=28.58, \gamma=0.0, m_0=-1.143, m_1=-0.714$.
\end{enumerate}
All these systems have one Lyapunov exponent equal to zero, corresponding to perturbations along the direction of the flow, a universal feature of all autonomous systems. The selected systems provide a representative testbed for systematically evaluating the effects of chaotic encoding, as they offer a good contrast in dynamical complexity, phase space structure, and dissipative properties.

\subsection{Method for Calculating Lyapunov Exponents}
\label{app:c2}
To quantify the dynamical properties of the different chaotic systems, we calculated two key metrics:
\begin{itemize}
    \item \textbf{Maximum Lyapunov Exponent ($\lambda_{\max}$):} Characterizes the system's sensitivity to small perturbations in the initial conditions.
    \[
    \lambda_{\max} = \lim_{t\to\infty} \lim_{|\delta \mathbf{x}_0|\to 0} \frac{1}{t} \ln \frac{|\delta \mathbf{x}_t|}{|\delta \mathbf{x}_0|}
    \]
    Here, $\delta \mathbf{x}_0$ is an infinitesimal initial perturbation, and $\delta \mathbf{x}_t$ is its magnitude after time $t$. A positive $\lambda_{\max}$ is a necessary condition for chaos, indicating sensitive dependence on initial conditions.
    \begin{itemize}
        \item $\lambda_{\max} > 0$: The system exhibits chaotic behavior, with exponentially diverging trajectories.
        \item $\lambda_{\max} = 0$: The system is at a critical state or has a periodic trajectory.
        \item $\lambda_{\max} < 0$: The system exhibits stable behavior, with converging trajectories.
    \end{itemize}

    \item \textbf{Lyapunov sum ($\Sigma\lambda_i$):} Describes the rate of change of a volume element in phase space over time.
    \[
    \frac{d}{dt}\ln(V(t)) = \sum_{i=1}^{n} \lambda_i
    \]
    Here, $V(t)$ is the volume element at time $t$.
    \begin{itemize}
        \item $\Sigma\lambda_i < 0$: The phase space volume contracts, indicating the presence of an attractor.
        \item $\Sigma\lambda_i = 0$: The system is conservative (Hamiltonian), and phase space volume is preserved.
        \item $\Sigma\lambda_i > 0$: The phase space volume expands, indicating system divergence.
    \end{itemize}
\end{itemize}
We employed a modified QR decomposition method to calculate the Lyapunov exponents, with the following steps:
\begin{enumerate}
    \item \textbf{Jacobian Matrix Construction:} For each chaotic system, we first analytically derived the Jacobian matrix $\mathbf{J}(\mathbf{x})$, which describes the local linear rate of change of the state vector at each point.
    \item \textbf{Trajectory Integration and Orthonormal Basis Evolution:} We initialized an orthonormal basis matrix $\mathbf{Q}$ as the identity matrix. At each time step, we calculated the Jacobian matrix $\mathbf{J}$ at the current state, integrated the system's trajectory using the RK4 method, and then updated the orthonormal basis via the matrix exponential:
    \[
    \mathbf{Q}_{\text{new}} = e^{\mathbf{J}\Delta t}\mathbf{Q}
    \]
    At fixed intervals (we used every 5 steps), we performed a QR decomposition:
    \[
    \mathbf{Q}_{\text{new}} = \mathbf{Q'}\mathbf{R}
    \]
    Then, we accumulated the logarithms of the diagonal elements of $\mathbf{R}$:
    \[
    s_i += \ln(\mathbf{R}_{ii})
    \]
    \item \textbf{Exponent Calculation:} After completing the trajectory integration, the Lyapunov exponents were calculated by normalizing the accumulated values:
    \[
    \lambda_i = \frac{s_i}{t_{\text{total}}}
    \]
    where $t_{\text{total}}$ is the total integration time (number of QR decompositions $\times$ time step).
\end{enumerate}
For each system, calculations were based on the chaos evolution time $t_{\max}$ and a time step of $h=0.01$, using the same initial conditions for all systems to ensure a fair comparison. To validate the robustness of our results, we performed the following stability checks:
\begin{itemize}
    \item Performed multiple calculations (10 times) for each system using different initial conditions and computed the variance.
    \item Progressively increased the total integration time until the exponent values stabilized (variation < 1\%).
    \item Compared our results with known values from the literature.
\end{itemize}

\subsection{Experimental Parameter Space}
\label{app:c3}
We employed a grid search strategy to systematically evaluate two key temporal parameters of attractor evolution:
\begin{enumerate}
    \item \textbf{Chaos evolution time:} $t_{\max} \in \{2, 4, 8, 16, 32\}$
    \item \textbf{Sampling steps:} $N \in \{1, 3, 5, 10, 20\}$
\end{enumerate}
To more clearly observe differences in encoding efficiency, we conducted experiments on a 10\% random subsample of the MNIST dataset (6,000 training and 1,000 testing samples), continuing within the SNN/ANN comparison framework from Appendix \ref{app:A}. This subsampling was intentionally designed to accentuate performance differences arising from encoding efficiency, as smaller datasets make such disparities more significant and measurable. The preprocessing method was identical to that described in Appendix \ref{app:A}.

\subsection{Theoretical Basis for Time-Embedding Parameters}
\label{app:c4}
To understand the theoretical basis for selecting the number of evolution steps ($N$) in dynamical encoding, we draw inspiration from Takens' Embedding Theorem\cite{takens2006detecting}, which addresses state space reconstruction from a time series. The theorem states that for a $d$-dimensional dynamical system, a full reconstruction of its state space requires an embedding dimension of at least $2d+1$. For a 3-dimensional system (such as the Lorenz attractor mentioned in \ref{app:c1}), the theoretical minimum embedding dimension is 7. This corresponds to at least $\lceil 7/3 \rceil = 3$ time steps, meaning $N=3$ satisfies the minimum requirement.

Our systematic parameter exploration (Figure \ref{fig:param_search}A in the main text) experimentally supports this theoretical lower bound. In the under-sampled region (1-3 steps), performance is significantly below the stable level, indicating that an overly short evolution time is insufficient to capture the essential dynamics of the chaotic system. When the number of steps increases to what we term the 'adequately-sampled region' (5-10 steps), performance reaches a stable, high level. Although the corresponding embedding space dimension (e.g., $5 \times 3=15$ to $10 \times 3=30$) far exceeds the theoretical minimum, this fuller evolution is crucial for an effective \textit{encoding} task. The reason is that the goal of encoding is to distinguish trajectories generated from different input feature values (as initial conditions). An evolution of only a few steps (e.g., $N=3$) might only capture the system's near-linear behavior, failing to fully express the sensitive dependence on initial conditions that is characteristic of chaos. Trajectories from different initial conditions need sufficient time to accumulate differences and separate effectively in phase space.

However, as the number of evolution steps increases further into the 'over-sampled region' (e.g., $N > 10$, as shown in Figure 3A), the performance improvement shows diminishing returns (Figure \ref{fig:param_search}C) and may even be slightly negatively affected by unnecessary computational complexity and potential trajectory self-folding, which can hinder discriminability. This suggests that the additional temporal information provides no significant benefit for the current discrimination task and instead increases computational cost.

Considering performance, energy efficiency, and cross-task universality, we fixed $N=5$ for all main experiments in this paper. The theoretical foundation for this decision lies in our core encoding strategy: each feature dimension of the input data is independently encoded through a 3-dimensional chaotic system run for $N$ steps. Therefore, an evolution of 5-10 steps provides sufficient time to express its sensitive dependence on the initial condition (i.e., a single input feature value) and generate spatiotemporal trajectories with enough discriminability to exhibit robust efficiency when processing tasks with varying input dimensions and complexities. Ultimately, our choice of $N=5$ was made to validate the universality and effectiveness of dynamical encoding in a scenario that prioritizes lower energy consumption and computational requirements. This choice implies that there is still room for further performance optimization in neural computation. This strategy of selecting $N$ based on the dynamics of the encoder itself, rather than the dimensionality of the input data, reveals a core advantage of our method: it provides a unified and efficient theoretical foundation for cross-task applications.

\subsection{Theoretical Basis and Calculation Method for Active Information Storage (AIS)}
\label{app:c5}
Active Information Storage (AIS) is a core concept in information dynamics\cite{lizier2012local}, used to quantify the mutual information between a system's current state and its past states. For a time series $X = \{x_1, x_2, \dots, x_n\}$, the general definition of AIS is $AIS(X) = I(X_t; X_t^{-})$, where $I(\cdot;\cdot)$ denotes Shannon mutual information\cite{shannon1948mathematical}, $X_t$ is the current state, and $X_t^{-}$ is the past state vector. Since we are analyzing discrete time series generated by numerical integration, a lag of $\tau=1$ corresponds to the fundamental time scale of the system. Therefore, in our implementation, we define AIS as the mutual information between the current state and the immediately preceding state:
\[
AIS = I(X_t; X_{t-1})
\]
To ensure the numerical stability of the AIS calculation, we preprocessed the chaotic trajectories. Specifically, we applied Z-score normalization to each coordinate dimension: $x'_{i}(t) = (x_i(t) - \mu_i) / \sigma_i$, where $\mu_i$ and $\sigma_i$ are the mean and standard deviation of that dimension, respectively. The normalized data was then clipped to the range $[-5, 5]$ to prevent extreme values from affecting probability estimates, and any trajectory segments containing NaN or infinite values were excluded.

Probability density was estimated using the equal-width histogram method\cite{scott2015multivariate}. We divided the data space into $K$ equal-width bins and added a Laplace smoothing factor of $\epsilon = 10^{-10}$ to avoid zero-probability issues. The joint probability distribution was estimated via a 2D histogram:
\[
P(X_t=x_i, X_{t-1}=x_j) = \frac{n_{ij} + \epsilon}{N + K^2\epsilon}
\]
where $n_{ij}$ is the number of samples falling into the $(i,j)$-th joint bin, and $N$ is the total number of samples. The number of bins was chosen based on Scott's rule\cite{scott1979optimal}, which suggests an optimal bin count of $K \approx 2N^{1/3}$ for a univariate case. Given our trajectory lengths (typically 800-1600 data points), 8 bins provided a reasonable resolution.

The marginal probability distribution was obtained by projection:
\[
P(X_t=x_i) = \frac{n_i + \epsilon}{N + K\epsilon}
\]
Based on the Kullback-Leibler divergence definition, AIS was calculated as\cite{kullback1951information}:
\[
AIS = \sum_{i,j} P(x_i, x_j) \log_2 \frac{P(x_i, x_j)}{P(x_i)P(x_j)}
\]
For the 3-dimensional chaotic systems, we adopted a per-dimension analysis strategy, calculating the AIS for each spatial coordinate separately: $AIS_x = I(x_t; x_{t-1})$, $AIS_y = I(y_t; y_{t-1})$, and $AIS_z = I(z_t; z_{t-1})$. This approach reveals the independent memory properties of each axis and identifies the information processing characteristics of different systems in each dimension, while avoiding the sparse sampling problems associated with high-dimensional joint probability estimation. Finally, we obtained a comprehensive measure of the system's overall information storage capacity by calculating the average: $AIS_{\text{avg}} = (AIS_x + AIS_y + AIS_z) / 3$. The quantitative relationship between AIS and other dynamical metrics shown in Figure 4 provides a theoretical basis for optimizing the choice of chaotic systems in neural dynamic encoding.

\subsection{Evaluation Metrics and Results}
\label{app:c6}
To evaluate the effectiveness of different chaotic systems and parameter combinations, we measured the following metrics:
\begin{enumerate}
    \item \textbf{Accuracy and Convergence Speed:} same metrics as detailed in Appendix~\ref{app:a5}.
    \item \textbf{Energy Efficiency:} Total spike count, used to assess energy consumption (in units of $10^6$ for comparability).
\end{enumerate}
To ensure the robustness of our results, we performed the following statistical analyses:
\begin{enumerate}
    \item Each parameter combination was run 10 times with different random initializations.
    \item Paired t-tests were used to compare performance differences between chaotic systems.
    \item Significance analysis was conducted to assess the reliability of the estimates.
\end{enumerate}

Table~\ref{table:param_search_N} shows the performance metrics for each chaotic system from the grid search (see \ref{app:c3}), corresponding to the results in Figures \ref{fig:param_search}a, c, and d of the main text.

\begin{table}[h!]
\centering
\caption{Parameter search statistics for different chaotic attractors (varying $N$).}
\label{table:param_search_N}
\resizebox{\textwidth}{!}{%
\begin{tabular}{@{}llcccccc@{}}
\toprule
 & \textbf{Sampling Steps (N)} & \textbf{Aizawa} & \textbf{Chua} & \textbf{Lorenz} & \textbf{Nosé-Hoover} & \textbf{Rössler} & \textbf{Sprott} \\ \midrule
\multirow{5}{*}{\textbf{Accuracy (\%)}} & 1 & 91.8$\pm$0.3 & 92.1$\pm$0.3 & 91.8$\pm$0.6 & 92.2$\pm$0.2 & 92.0$\pm$0.4 & 92.0$\pm$0.5 \\
 & 3 & 93.9$\pm$0.3 & 92.9$\pm$0.2 & 93.5$\pm$0.5 & 92.5$\pm$0.3 & 93.4$\pm$0.6 & 93.9$\pm$0.3 \\
 & 5 & 94.2$\pm$0.2 & 94.1$\pm$0.2 & 93.1$\pm$0.3 & 93.8$\pm$0.1 & 93.9$\pm$0.2 & 94.3$\pm$0.1 \\
 & 10 & 94.2$\pm$0.2 & 94.0$\pm$0.3 & 93.5$\pm$0.8 & 93.7$\pm$0.1 & 94.0$\pm$0.2 & 94.4$\pm$0.2 \\
 & 20 & 94.2$\pm$0.2 & 94.1$\pm$0.2 & 93.3$\pm$1.4 & 93.6$\pm$0.1 & 94.0$\pm$0.0 & 94.1$\pm$0.2 \\ \midrule
\multirow{5}{*}{\textbf{Convergence (epochs)}} & 1 & 39.4$\pm$5.2 & 37.6$\pm$2.1 & 39.3$\pm$3.9 & 37.6$\pm$3.4 & 36.1$\pm$4.9 & 35.9$\pm$2.6 \\
 & 3 & 24.9$\pm$2.8 & 25.6$\pm$3.5 & 23.8$\pm$8.5 & 17.1$\pm$1.0 & 25.7$\pm$6.1 & 19.8$\pm$2.5 \\
 & 5 & 21.6$\pm$2.8 & 29.9$\pm$1.9 & 30.2$\pm$2.7 & 22.1$\pm$1.8 & 20.6$\pm$2.7 & 22.6$\pm$1.2 \\
 & 10 & 21.3$\pm$3.5 & 24.4$\pm$3.4 & 30.6$\pm$6.6 & 24.3$\pm$2.9 & 23.3$\pm$4.8 & 27.8$\pm$3.0 \\
 & 20 & 22.2$\pm$4.5 & 30.1$\pm$2.9 & 28.9$\pm$7.8 & 26.4$\pm$2.6 & 25.6$\pm$1.9 & 24.3$\pm$2.2 \\ \midrule
\multirow{5}{*}{\textbf{Energy ($10^6$ spikes)}} & 1 & 1.5$\pm$0.2 & 1.5$\pm$0.2 & 1.5$\pm$0.2 & 1.4$\pm$0.2 & 1.4$\pm$0.3 & 1.5$\pm$0.3 \\
 & 3 & 4.3$\pm$0.4 & 6.3$\pm$0.3 & 9.1$\pm$0.6 & 5.1$\pm$0.3 & 3.3$\pm$0.4 & 5.1$\pm$0.3 \\
 & 5 & 7.5$\pm$0.8 & 14.6$\pm$0.7 & 18.1$\pm$1.5 & 8.5$\pm$0.6 & 6.6$\pm$0.8 & 8.7$\pm$0.6 \\
 & 10 & 12.3$\pm$2.8 & 25.9$\pm$1.6 & 34.2$\pm$3.7 & 18.2$\pm$1.0 & 13.4$\pm$0.7 & 14.0$\pm$2.6 \\
 & 20 & 26.1$\pm$5.0 & 41.6$\pm$3.6 & 58.7$\pm$3.5 & 37.2$\pm$1.9 & 29.5$\pm$2.1 & 27.3$\pm$5.5 \\ \bottomrule
\end{tabular}
}
\end{table}

Table~\ref{table:param_search_tmax} presents the results corresponding to Figure 3b in the main text. For this analysis, only parameter combinations with a sufficient number of time steps ($N \ge 5$) were included (i.e., combinations for $N$=5, 10, 20, each run 10 times for a total of 30 statistical combinations) to ensure adequate time embedding.

\begin{table}[h!]
\centering
\caption{Parameter search statistics for different chaotic attractors (varying Tmax).}
\label{table:param_search_tmax}
\begin{tabular}{@{}lccccc@{}}
\toprule
\textbf{Attractor System} & \textbf{Lyapunov Sum} & \textbf{\makecell{Acc. on \\ $T_{max}$=4 (\%)} } & \textbf{\makecell{Acc. on \\ $T_{max}$=8 (\%)} } & \textbf{\makecell{Acc. on \\ $T_{max}$=16 (\%)} } & \textbf{\makecell{Acc. on \\ $T_{max}$=32 (\%)} } \\ \midrule
Aizawa & 0.067 & 94.3$\pm$0.1 & 94.2$\pm$0.2 & 94.2$\pm$0.2 & 94.2$\pm$0.2 \\
Chua & -4.893 & 94.0$\pm$0.1 & 94.3$\pm$0.1 & 93.9$\pm$0.2 & 94.1$\pm$0.3 \\
Lorenz & -13.667 & 94.0$\pm$0.8 & 93.0$\pm$0.7 & 92.6$\pm$1.2 & 93.5$\pm$0.1 \\
Nosé-Hoover & 0.966 & 93.8$\pm$0.1 & 93.8$\pm$0.1 & 93.7$\pm$0.1 & 93.7$\pm$0.1 \\
Rössler & -5.588 & 94.0$\pm$0.1 & 94.1$\pm$0.1 & 93.8$\pm$0.2 & 94.0$\pm$0.1 \\
Sprott & -1.000 & 94.2$\pm$0.2 & 94.3$\pm$0.1 & 94.3$\pm$0.1 & 94.4$\pm$0.3 \\ \bottomrule
\end{tabular}
\end{table}

\textbf{Key finding:} When $t_{\max}$ is between 8 and 16, the Lyapunov sum and classification accuracy show the strongest correlation ($r>0.7, p<0.01$), establishing an optimal time window for chaotic evolution. As $t_{\max}$ increases to 32, this correlation remains significant but weakens slightly, supporting the hypothesis of an optimal evolution time window.

Based on these conclusions, after determining the optimal time parameters ($t_{\max}=8, N=5$), we conducted a systematic analysis of the dynamical properties of all chaotic attractor systems. Tables~\ref{table:perf_metrics} and \ref{table:dyn_metrics} show the performance and dynamical property metrics for each system (corresponding to Figure \ref{fig:dynamics_correlation} in the main text).

\begin{table}[h!]
\centering
\caption{Performance metrics of different chaotic attractors.}
\label{table:perf_metrics}
\begin{tabular}{@{}lccc@{}}
\toprule
\textbf{Attractor System} & \textbf{Accuracy (\%)} & \textbf{Convergence (epochs)} & \textbf{Spikes ($\times 10^6$)} \\ \midrule
Lorenz & 93.0$\pm$1.1 & 34.3$\pm$14.9 & 18.22$\pm$0.43 \\
Rössler & 93.6$\pm$0.7 & 22.2$\pm$11.3 & 6.17$\pm$0.87 \\
Aizawa & 94.0$\pm$0.6 & 22.4$\pm$6.5 & 6.52$\pm$0.47 \\
Nosé-Hoover & 93.7$\pm$1.1 & 15.8$\pm$10.2 & 6.40$\pm$0.50 \\
Sprott & 93.6$\pm$1.4 & 24.4$\pm$12.5 & 9.23$\pm$0.43 \\
Chua & 94.1$\pm$0.9 & 23.5$\pm$8.0 & 5.72$\pm$0.58 \\ \bottomrule
\end{tabular}
\end{table}

\begin{table}[h!]
\centering
\caption{Dynamical property metrics of different chaotic attractors.}
\label{table:dyn_metrics}
\resizebox{\textwidth}{!}{%
\begin{tabular}{@{}lcccccc@{}}
\toprule
\textbf{Attractor System} & \textbf{Max. Lyapunov Exp. ($\lambda_{\max}$)} & \textbf{Lyapunov Sum ($\Sigma\lambda_i$)} & \textbf{AIS (x-axis)} & \textbf{AIS (y-axis)} & \textbf{AIS (z-axis)} & \textbf{Mean AIS} \\ \midrule
Lorenz & 0.370 & -13.667 & 2.43 & 2.46 & 3.04 & 2.64 \\
Rössler & 0.099 & -5.497 & 2.76 & 2.82 & 2.84 & 2.81 \\
Aizawa & 0.176 & -0.834 & 1.46 & 1.38 & 1.32 & 1.39 \\
Nosé-Hoover & 0.203 & 0.000 & 2.55 & 2.89 & 3.11 & 2.85 \\
Sprott & 0.352 & -1.000 & 3.05 & 3.10 & 0.58 & 2.24 \\
Chua & 0.403 & -4.416 & 2.98 & 3.08 & 3.16 & 3.07 \\ \bottomrule
\end{tabular}
}
\end{table}

These results support our core hypothesis: the phase space contraction property (quantified by $\Sigma\lambda_i$) has a more significant impact on neural encoding efficiency than the degree of chaos alone (quantified by $\lambda_{\max}$). In particular, systems where $\Sigma\lambda_i$ is close to zero but still negative exhibit the best performance-energy trade-off. Concurrently, AIS is a good predictor of post-encoding performance.

\printbibliography[title={References for Appendix C}, heading=subbibliography, env=appendixbib]
\end{refsection}

\section{Mixed Oscillator System Experiments and Critical Phenomenon Analysis}
\label{app:d}
\begin{refsection}
\setappendixprefix{D}

This appendix details the experimental design for the systematic exploration of how dynamical properties affect neural computation. Its core focus is the bimodal performance and energy phase transition phenomena presented in Figure~\ref{fig:phase_transition} of the main text.

\subsection{Mixed Oscillator System Model Design}
\label{app:d1}
To investigate the relationship between Lyapunov exponents and neural computation performance, we designed a parameterized mixed oscillator system based on the classic Duffing oscillator\cite{kovacic2011duffing}. This system retains the core nonlinear dynamical features of the Duffing oscillator (linear restoring force, cubic nonlinearity, tunable damping) while adding a third dimension and coupling terms. It inherits the mature theoretical foundation and known dynamical characteristics of the Duffing oscillator, and by precisely controlling its properties, we can achieve continuous sampling of the state space from strongly expansive to strongly dissipative. The system is described by the following set of differential equations:

\begin{align*}
    \frac{dx}{dt} &= y \\
    \frac{dy}{dt} &= -\alpha x - \beta x^3 - \delta y + \gamma z \\
    \frac{dz}{dt} &= -\omega x - \delta z + \gamma xy
\end{align*}

The physical significance and control effect of each parameter are as follows:
\begin{itemize}
    \item \textbf{$\alpha$}: The linear restoring force coefficient, which controls the basic resonant properties of the system.
    \item \textbf{$\beta$}: The nonlinear term coefficient, controlling the strength of the nonlinearity. The system is purely linear when $\beta=0$.
    \item \textbf{$\delta$}: The core control parameter for energy dissipation/injection. $\delta < 0$ corresponds to energy injection, leading to expansive dynamics (energy increases, phase space volume expands). $\delta > 0$ corresponds to energy dissipation (energy decreases, phase space volume contracts). $\delta=0$ represents the critical point of a conservative system.
    \item \textbf{$\gamma$}: The coupling coefficient between dimensions, used to break system symmetry and generate complex dynamics.
    \item \textbf{$\omega$}: The natural frequency of the system.
\end{itemize}
By varying the value of $\delta$, we can precisely control the Lyapunov sum ($\Sigma\lambda_i$) and thus systematically study the impact of the phase space volume's rate of contraction/expansion on the efficiency of neural encoding and computation.

\subsection{Experimental Design and Procedure}
\label{app:d2}
To systematically map the dynamical spectrum, we fixed the base parameters of the oscillator system ($\alpha=2.0, \beta=0.1, \gamma=0.1, \omega=1.0$) and performed a fine-grained sampling of the core dissipation parameter $\delta$ over the interval $[-1.5, 10.0]$ (see Table~\ref{table:d1_oscillator_data}). This ensures comprehensive coverage from a strongly expansive regime ($\Sigma\lambda_i \approx +3$) to a strongly dissipative regime ($\Sigma\lambda_i \approx -20$), with a particular focus on the critical transition region around $\Sigma\lambda_i \approx 0$.

The SNN network configuration was consistent with that described in Appendix~\ref{app:A}, including learning rate, hidden layer sizes, and training strategy. We also used the optimal parameters ($t_{\max}=8, N=5$) identified in Appendix \ref{app:C} with a random 10\% sample of the MNIST dataset (Appendix \ref{app:c3}). Each parameter configuration was run for 10 independent trials with different random seeds to ensure statistical reliability. During training, the following metrics were recorded:
\begin{itemize}
    \item \textbf{Performance Metrics}: The primary metrics were the highest classification accuracy achieved for each configuration, the number of epochs to convergence, and the total spike count at convergence (as a proxy for energy consumption).
    \item \textbf{Dynamical Metrics}: For each value of $\delta$, we calculated the maximum Lyapunov exponent ($\lambda_{\max}$) and the Lyapunov sum ($\Sigma\lambda_i$) for the corresponding oscillator system.
    \item \textbf{Statistical Robustness}: To ensure the reliability of our conclusions, each experiment was repeated 10 times. The final reported results are the mean and standard deviation of these 10 runs.
\end{itemize}
The specific details of metric calculation are the same as described in Appendix \ref{app:c2}.

\begin{table}[h!]
\centering
\caption{\textbf{Encoding performance characteristics of the mixed oscillator system across different dynamical states.} The parameter $\delta$ controls the system's dynamics. $\lambda_{\max}$ denotes the maximum Lyapunov exponent, $\Sigma\lambda_i$ denotes the Lyapunov sum, and AIS denotes the average Active Information Storage. \textbf{Bold} values indicate high-performance regions (accuracy $\ge 93.50\%$), while \textit{italicized} values indicate computationally inefficient regions (AIS $< 1.70$). The critical point ($\delta=0$) lies at the boundary between the expansive and dissipative regions.}
\label{table:d1_oscillator_data}
\resizebox{0.8\textwidth}{!}{%
\begin{tabular}{@{}lcccccc@{}}
\toprule
\textbf{$\delta$} & \textbf{Accuracy (\%)} & \textbf{Convergence (epochs)} & \textbf{Spikes (millions)} & \textbf{$\lambda_{\max}$} & \textbf{$\Sigma\lambda_i$} & \textbf{AIS} \\ \midrule
-1.5 & \textbf{93.94 $\pm$ 0.96} & 9.1 $\pm$ 4.0 & 19.16 $\pm$ 0.33 & 1.17 & 3 & 2.99 \\
-1.0 & \textbf{93.87 $\pm$ 0.45} & 9.2 $\pm$ 5.0 & 17.85 $\pm$ 0.28 & 0.79 & 2 & 2.95 \\
-0.6 & \textbf{93.82 $\pm$ 0.64} & 8.7 $\pm$ 4.0 & 15.93 $\pm$ 0.93 & 0.44 & 1.2 & 3.05 \\
-0.3 & 93.00 $\pm$ 0.75 & 12.5 $\pm$ 4.7 & 13.68 $\pm$ 0.89 & 0.24 & 0.6 & 2.80 \\
-0.15& \textbf{93.68 $\pm$ 1.08} & 14.3 $\pm$ 6.0 & 11.98 $\pm$ 1.28 & 0.16 & 0.3 & 3.02 \\
0    & 92.82 $\pm$ 0.69 & 15.6 $\pm$ 5.7 & 7.40 $\pm$ 0.93 & 0.08 & 0 & 2.97 \\
0.15 & 93.19 $\pm$ 0.74 & 12.2 $\pm$ 3.7 & 4.80 $\pm$ 0.68 & 0 & -0.3 & 2.99 \\
0.3  & 92.54 $\pm$ 0.71 & 11.7 $\pm$ 3.3 & 4.23 $\pm$ 0.52 & -0.07 & -0.6 & 2.99 \\
0.6  & 92.16 $\pm$ 0.85 & 12.9 $\pm$ 4.6 & 4.34 $\pm$ 0.58 & -0.22 & -1.2 & 2.66 \\
1.0  & 92.67 $\pm$ 0.83 & 19.9 $\pm$ 5.3 & 4.47 $\pm$ 0.26 & -0.41 & -2 & 2.15 \\
1.5  & 92.04 $\pm$ 1.46 & 15.3 $\pm$ 5.4 & 4.22 $\pm$ 0.50 & -0.70 & -3 & \textit{1.62} \\
2.0  & 92.91 $\pm$ 0.89 & 15.6 $\pm$ 5.7 & 4.06 $\pm$ 0.42 & -0.88 & -4 & \textit{1.06} \\
2.5  & 92.10 $\pm$ 1.01 & 13.9 $\pm$ 6.0 & 4.00 $\pm$ 0.60 & -1.15 & -5 & \textit{1.09} \\
4.0  & 92.27 $\pm$ 0.97 & 14.9 $\pm$ 5.1 & 3.92 $\pm$ 0.38 & -0.55 & -8 & 2.07 \\
5.0  & 93.07 $\pm$ 0.99 & 13.8 $\pm$ 5.4 & 4.04 $\pm$ 0.21 & -0.42 & -10 & 2.44 \\
7.0  & 92.93 $\pm$ 1.24 & 15.5 $\pm$ 3.1 & 4.44 $\pm$ 0.48 & -0.29 & -14 & 2.84 \\
10.0 & \textbf{93.29 $\pm$ 0.78} & 14.5 $\pm$ 5.1 & 4.92 $\pm$ 0.38 & -0.20 & -20 & 3.08 \\
\bottomrule
\end{tabular}
}
\end{table}

\subsection{Quantitative Analysis of Critical Phenomena and Symmetry Breaking}
\label{app:d3}
To substantiate the critical phase transition revealed in Figure~\ref{fig:phase_transition}, we adopted core methods from the study of phase transitions in statistical physics\cite{stanley1971phase} to quantitatively analyze the scaling behavior of system performance metrics near the critical point ($\lambda_c = 0$). This method aims to verify whether the macroscopic behavior of the system (e.g., accuracy, energy consumption) follows a specific power-law relationship as it crosses the critical point, and to calculate the critical exponent, $\beta$, that describes this relationship. Our analysis procedure followed these steps:
\begin{enumerate}
    \item \textbf{Locating the phase transition point}: With $\lambda_c = 0$ as the critical point, we calculated the distance of the system parameters ($\Sigma\lambda_i$ and $\lambda_{\max}$) from this point: $|\lambda - \lambda_c|$.
    \item \textbf{Calculating relative change}: For each performance metric (accuracy, spike count, convergence epochs), we calculated its change, $\Delta\text{Metric}$, relative to a baseline value.
    \item \textbf{Power-law fitting}: We hypothesized a power-law relationship near the critical point: $\Delta\text{Metric} \propto |\lambda - \lambda_c|^\beta$.
    \item \textbf{Linear regression}: By performing a linear regression on a log-log plot ($\log(\Delta\text{Metric})$ vs. $\log|\lambda - \lambda_c|$), we can fit the slope, which corresponds to the critical exponent $\beta$. The quality of the fit is assessed by the coefficient of determination, $R^2$, and the p-value.
\end{enumerate}

\begin{figure}[h!]
    \centering
    \includegraphics[width=\textwidth]{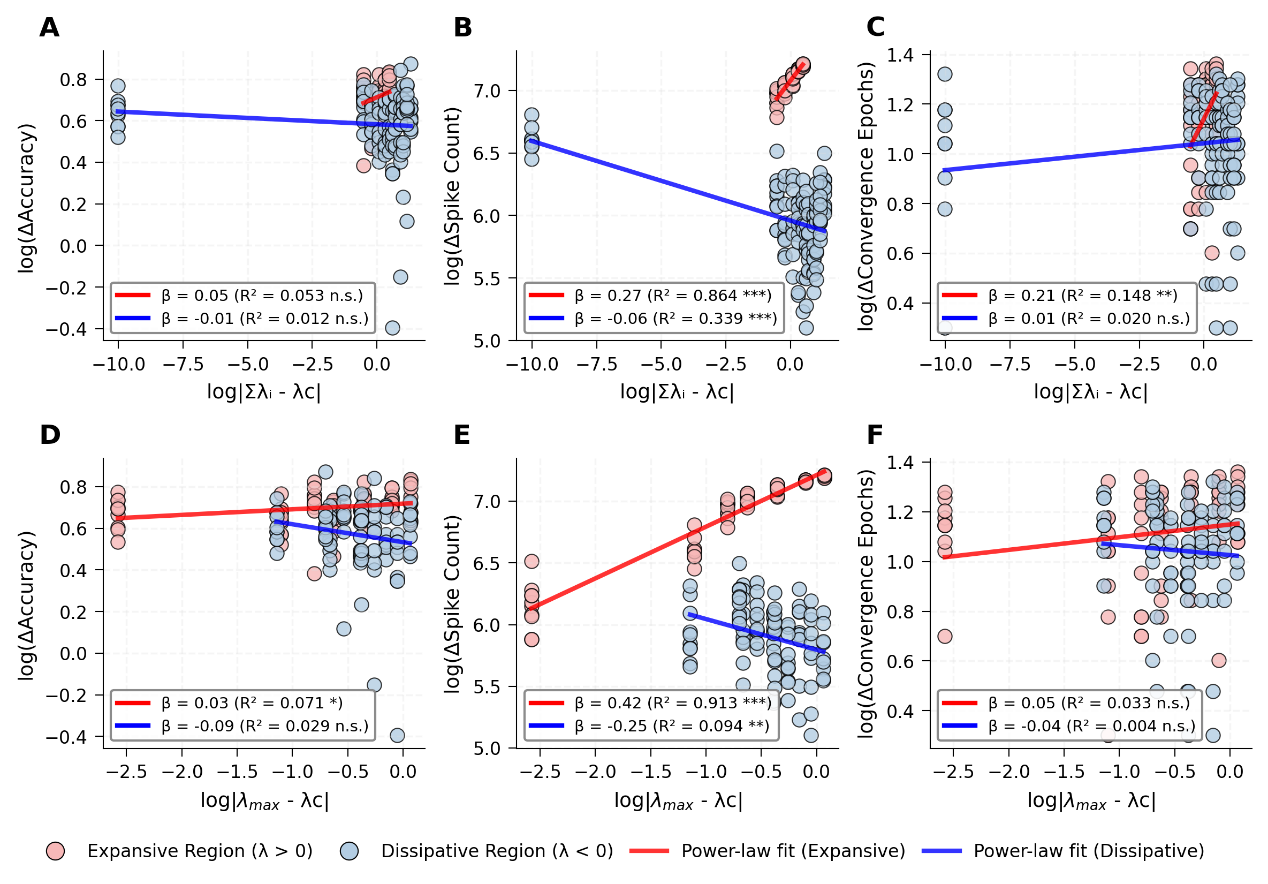} 
    \caption{\textbf{Power-law scaling and symmetry breaking in the critical region under different Lyapunov indicators.} The log-log plots show the critical scaling properties of three performance metrics against two Lyapunov indicators. Top row (\textbf{a-c}) corresponds to the Lyapunov sum ($\Sigma\lambda_i$), and the bottom row (\textbf{d-f}) to the maximum Lyapunov exponent ($\lambda_{\max}$). From left to right, the columns show accuracy, spike count, and convergence epochs. The horizontal axis is the log of the distance to the critical point, $\log|\lambda - \lambda_c|$, and the vertical axis is the log of the relative change in the performance metric, $\log(\Delta\text{Metric})$. Red data points represent the expansive region ($\lambda > 0$), and blue data points represent the dissipative region ($\lambda < 0$). Solid lines are power-law fits, with legends showing the critical exponent $\beta$, coefficient of determination $R^2$, and statistical significance. Spike count (\textbf{b, e}) exhibits the strongest power-law relationship ($R^2 > 0.86$ in the expansive region), confirming the strong coupling between neural activity and dynamical parameters. while convergence epochs (\textbf{c, f}) show moderately strong critical behavior and accuracy (\textbf{a, d}) displays the weakest properties ($R^2 < 0.1$). Notably, $\lambda_{\max}$ is a very strong predictor in the expansive region but its predictive power drops sharply in the dissipative region, whereas $\Sigma\lambda_i$ maintains predictive power in both regions, revealing a significant symmetry breaking.}
    \label{fig:d1_critical_scaling}
\end{figure}

The analysis in Figure~\ref{fig:d1_critical_scaling} reveals the scaling properties for the three performance metrics:
\begin{itemize}
    \item \textbf{Strong Power-Law Property (Spike Activity)}: Spike count shows the strongest critical behavior. Under the $\Sigma\lambda_i$ indicator, the critical exponent $\beta$ reaches 0.27 in the expansive region, with an explained variance of 86.4\%. The behavior is even more pronounced for the $\lambda_{\max}$ indicator, where $\beta = 0.42$ and $R^2 = 0.913$ in the expansive region. This strong power-law relationship directly confirms the fundamental link between the intensity of neural activity and the dynamical parameters, providing powerful evidence for the existence of a critical phase transition. Notably, both indicators show a significant sign reversal for $\beta$ from positive in the expansive region to negative in the dissipative region, suggesting a fundamental shift in the mechanism of neural activity regulation across the critical point.
    
    \item \textbf{Moderate Criticality (Convergence Speed)}: Convergence epochs exhibit a moderately strong critical behavior. For the $\Sigma\lambda_i$ indicator, $\beta = 0.21$ in the expansive region is statistically significant ($p = 0.006$). However, the criticality is weaker and not significant for the $\lambda_{\max}$ indicator. This difference highlights the complementary roles of the two Lyapunov indicators in predicting convergence performance: the change in phase space volume ($\Sigma\lambda_i$) has a more direct impact on the learning process, while the role of trajectory sensitivity ($\lambda_{\max}$) is more indirect.
    
    \item \textbf{Weak Criticality (Learning Accuracy)}: In contrast, accuracy shows relatively low explained variance ($R^2 < 0.1$) for both Lyapunov indicators, and in most cases, the fits are not statistically significant.
\end{itemize}

In this experiment, we observed a significant \textbf{symmetry breaking}. $\lambda_{\max}$ demonstrates extremely strong predictive power in the expansive region (e.g., $R^2 = 0.913$ for spike count) but this power drops sharply in the dissipative region (to $R^2 = 0.094$), whereas $\Sigma\lambda_i$ maintains a degree of predictive power in both regions, showing better symmetry. Furthermore, for most metrics, not only the magnitude but also the sign of the critical exponent changes across the critical point, indicating a fundamental shift in the system's dynamical mechanisms.

Additionally, the analysis reveals the complementary nature of the two Lyapunov indicators:
\begin{itemize}
    \item \textbf{Advantage of $\Sigma\lambda_i$}: It maintains strong predictive power for spike activity in the dissipative region ($R^2 = 0.339$), reflecting the control exerted by phase space volume contraction on neural activity.
    \item \textbf{Advantage of $\lambda_{\max}$}: It has extremely strong predictive power for spike activity and convergence in the expansive region ($R^2 > 0.9$), reflecting the dominant influence of chaotic sensitivity on neural dynamics.
\end{itemize}
This complementarity suggests that a complete description of the critical behavior of neural computation requires consideration of both the change in phase space volume ($\Sigma\lambda_i$) and trajectory sensitivity ($\lambda_{\max}$).

\subsection{Mathematical Model of Dynamical Alignment}
\label{sec:appendix_math_model}

This section details the mathematical framework used to provide a unified, mechanistic explanation for the observed bimodal phenomena. We apply the established theory of neurons driven by correlated noise to model our system, focusing on the essential time-scale dynamics. This approach allows us to use the input's autocorrelation time ($\tau_{corr}$) as a key parameter to explain the full spectrum of behaviors within a single, self-consistent theory.

To start with, we select the Leaky Integrate-and-Fire (LIF) model as the fundamental computational unit of our network. It is the standard model in computational neuroscience, capturing the essential dynamics of neuronal integration and firing\cite{dayan2005theoretical}. The evolution of its sub-threshold membrane potential, $V(t)$, is governed by:
\begin{equation}
\tau_m \frac{dV(t)}{dt} = -(V(t) - V_{\text{rest}}) + I(t)
\label{eq:lif_model}
\end{equation}
where:
\begin{itemize}
    \item $V(t)$: The neuron's membrane potential at time $t$.
    \item $\tau_m$: The membrane time constant, which defines the timescale of neuronal integration.
    \item $V_{\text{rest}}$: The resting potential, set to 0 for simplicity.
    \item $I(t)$: The total synaptic input current received by the neuron.
\end{itemize}
When $V(t)$ reaches a threshold $V_{\text{th}}$, the neuron fires a spike and its potential is reset to $V_{\text{reset}}$.

In our multi-layered network architecture, the source of the input current $I(t)$ depends on the neuron's position. For the first hidden layer ($l=1$), the input current $I^{(1)}(t)$ is directly generated by our external dynamical encoder. For subsequent layers ($l>1$), the input current to a neuron $i$ in layer $l$, denoted $I_i^{(l)}(t)$, is the sum of synaptic currents elicited by the spiking activity of neurons in the preceding layer ($l-1$):
\begin{equation}
    I_i^{(l)}(t) = \sum_{j} w_{ij} \sum_{k} \alpha(t - t_j^{(l-1),k})
\end{equation}
where $w_{ij}$ is the synaptic weight from neuron $j$ (in layer $l-1$) to neuron $i$ (in layer $l$), $t_j^{(l-1),k}$ is the time of the $k$-th spike from neuron $j$, and $\alpha(t)$ represents the postsynaptic current waveform.

Therefore, whether originating from the external encoder or generated by internal network activity, the current $I(t)$ is not simple white noise but rather a complex signal with temporal memory. For a tractable theoretical analysis that applies across all layers, we approximate this current as a well-defined stochastic process with a mean-field approach. Specifically, we use the Ornstein-Uhlenbeck (OU) process, a well-established model for exponentially correlated Gaussian noise (``colored noise'')\cite{brunel2000dynamics, fourcaud2002dynamics}. This allows us to capture the three essential statistical properties that govern the neuron's response:
\begin{itemize}
    \item Mean ($\mu_I$): The average intensity of the current.
    \item Variance ($\sigma_I^2$): The amplitude of the current's fluctuations.
    \item Autocorrelation Time ($\tau_{corr}$): This is the most critical parameter, quantifying the duration of the input's temporal memory.
\end{itemize}
The dynamics of the OU process are described by the following stochastic differential equation:
\begin{equation}
\tau_{corr} \frac{dI(t)}{dt} = -(I(t) - \mu_I) + \sqrt{2\tau_{corr}\sigma_I^2} \eta(t)
\label{eq:ou_process_model}
\end{equation}
where $\eta(t)$ is a Gaussian white noise process with zero mean and unit variance, i.e., $\langle \eta(t) \rangle = 0$ and $\langle \eta(t)\eta(t') \rangle = \delta(t-t')$. 

Before proceeding with the analysis, we should establish the intrinsic timescale of our 'hardware' first. In our computational experiments using the `snn.Leaky` neuron, the membrane time constant $\tau_m$ is determined implicitly by the discrete-time decay parameter $\beta$ and the simulation time step $\Delta t$. Their relationship is:
\begin{equation}
\beta = e^{-\Delta t / \tau_m}
\label{eq:beta_tau_relation}
\end{equation}
We can solve for $\tau_m$ by rearranging the equation:
\begin{equation}
\tau_m = -\frac{\Delta t}{\ln(\beta)}
\label{eq:timescale_separation}
\end{equation}
In our experimental setup, the total evolution time is set as $t_{max} = 8$ time units, divided into $N=5$ steps. This sets the duration of a single simulation step to $\Delta t = t_{max} / N = 8 / 5 = 1.6$ time units. With the fixed decay factor $\beta = 0.95$, we calculate the precise value of $\tau_m$ for our network:
\begin{equation}
\tau_m = -\frac{1.6}{\ln(0.95)} \approx -\frac{1.6}{-0.05129} \approx \textbf{31.19 time units}
\label{eq:tau_m_calculation_final}
\end{equation}

To understand how the input's temporal memory affects computation, we analyze the firing rate $\nu$ of a neuron driven by correlated noise. The rate is determined by the first-passage time of the membrane potential $V(t)$ to the threshold $V_{th}$. For a correlated input, this can be approximated using an effective 1D Fokker-Planck approach, yielding a solution analogous to the classic Siegert formula. A crucial insight from this theory is that the neuron's membrane acts as a low-pass filter with a characteristic timescale $\tau_m$. It is most responsive to input fluctuations that are faster than this timescale. When the input noise becomes slow and correlated, its power is concentrated at frequencies that the membrane filter attenuates. This reduces the variance of the resulting membrane potential, an effect captured by a now-classic correction factor to the variance term\cite{brunel2000dynamics}.

The firing rate $\nu$ can thus be approximated by\cite{brunel2000dynamics, fourcaud2002dynamics}:
\begin{equation}
\nu \approx \left( \tau_m \sqrt{\pi} \int_{ \frac{V_{reset}-\mu_V}{\sigma'_V} }^{ \frac{V_{th}-\mu_V}{\sigma'_V} } dy \, e^{y^2} (1 + \text{erf}(y)) \right)^{-1}
\label{eq:colored_noise_rate_integral_final}
\end{equation}

\textbf{Note}: This approximation assumes that the effective variance approach captures the essential effect of input correlations.

Here, $\mu_V $ is the mean membrane potential (equal to $\mu_I$ as $V_{rest}=0$), and the crucial component is $\sigma'_V$, the effective standard deviation of the membrane potential. This term quantifies the magnitude of voltage fluctuations that are effective at driving the neuron to threshold.

To accurately calculate the firing rate, we must use the effective variance, $(\sigma'_V)^2$, which correctly weighs the contribution of different input frequencies. This quantity is derived from the first-passage time analysis of the Fokker-Planck equation for correlated noise\cite{brunel2000dynamics, fourcaud2002dynamics}. A full derivation relies on perturbation expansions in the small parameter $\sqrt{\tau_s/\tau_m}$ (where $\tau_s$ is the synaptic/correlation time), but the well-established result for the effective variance is\cite{fourcaud2002dynamics, brunel2000dynamics}:
\begin{equation}
(\sigma'_V)^2 \propto \underbrace{\frac{\tau_m \sigma_I^2}{2}}_{\text{White Noise Limit}} \cdot \underbrace{\left( 1 + \frac{\tau_{corr}}{\tau_m} \right)^{-1}}_{\text{Correction for Memory}}
\label{eq:effective_sigma_final}
\end{equation}
The first term, $\tau_m \sigma_I^2 / 2$, is the correct variance of the membrane potential in the white-noise limit ($\tau_{corr} \to 0$), as derived from the Fokker-Planck formalism. The second term is the crucial correction factor that accounts for the input's temporal memory.

Equation \eqref{eq:effective_sigma_final} is the cornerstone of our theoretical framework. It describes that for a fixed input variance $\sigma_I^2$, increasing the input's temporal memory ($\tau_{corr}$) \textbf{reduces} the effective voltage fluctuations $(\sigma'_V)^2$ that are available to drive firing. This single principle allows us to explain the entire bimodal landscape: it implies that the ratio $\tau_{corr}/\tau_m$ is the critical parameter controlling the computational regime. To validate this, we experimentally measured $\tau_{corr}$ for our dynamical encoders' modes by computing the Autocorrelation Function (ACF) from high-resolution trajectories and identifying the time required for it to decay to $1/e$. The ACF is computed as:
\begin{equation}
    \text{ACF}(\tau) = \frac{\langle x(t)x(t+\tau) \rangle - \langle x(t) \rangle^2}{\langle x(t)^2 \rangle - \langle x(t) \rangle^2}
\end{equation}

where $x(t)$ represents the dynamical trajectory of the encoding system, $\tau$ is the time lag, $\langle \cdot \rangle$ denotes time averaging, and $\tau_{corr}$ is defined as the time when $\text{ACF}(\tau_{corr}) = 1/e$.

Our analysis yielded the following timescale ratios:
\begin{itemize}
    \item \textbf{Transition Mode ($\delta = 2.0$):} $\tau_{corr} \approx 0.78$ time units $\implies \boldsymbol{\tau_{corr}/\tau_m \approx 0.025}$
    \item \textbf{Dissipative Mode ($\delta = 10.0$):} $\tau_{corr} \approx 2.59$ time units $\implies \boldsymbol{\tau_{corr}/\tau_m \approx 0.083}$
    \item \textbf{Expansive Mode ($\delta = -1.5$):} $\tau_{corr} > 25.0$ time units $\implies \boldsymbol{\tau_{corr}/\tau_m \gtrsim 0.801}$
\end{itemize}
This empirical data confirms a bifurcation in computational mechanisms. We validated this theoretical framework by combing the internal activity statistics ($\mu_I$, $\sigma_I^2$) from our pre-trained networks (see Table \ref{tab:sup_data_summary} for full data), allowing us to identify the network's behavior in distinct regimes:

\paragraph{1. The Fluctuation-Driven Regime ($\boldsymbol{\tau_{corr} \ll \tau_m}$)}
This regime, which includes the \textbf{Dissipative Mode and Transition Region}, is defined by a small timescale ratio. As predicted by Eq. \eqref{eq:effective_sigma_final}, this ensures the effective variance $(\sigma'_V)^2$ is nearly maximal, meaning the input acts as an efficient, fluctuation-rich signal. As evidence in Table~\ref{tab:sup_data_summary}, the network learns to operate with a low mean input current in its core computational layers (e.g., $\boldsymbol{\mu_I \approx 0.15-0.2}$ in Layer 2-3 for the Dissipative mode), confirming its reliance on fluctuations to drive sparse, informative spikes. The resulting sparse activity is also visually apparent in the raster plots (Figure \ref{fig:neural_dynamics}B).

\paragraph{2. The Mean-Driven Regime ($\boldsymbol{\tau_{corr} \approx \tau_m}$)}
This regime corresponds to the \textbf{Expansive Mode}, where the input's long autocorrelation time ($\tau_{corr} \gtrsim \tau_m$) results in slow input fluctuations. Crucially, the neuron's membrane acts as a low-pass filter, inherently attenuating such slow fluctuations and rendering a fluctuation-driven coding strategy ineffective. As shown in Table~\ref{tab:sup_data_summary}, the network's mean drive current (e.g., \textbf{$\mu_I \approx 0.66$} in Layer 3) approximately 3-4 times higher than in the fluctuation-driven regime. This strong mean drive continuously pushes the membrane potential towards the firing threshold, making spike generation highly sensitive to the input's mean level. This steers the network into a robust mean-driven rate coding state at the cost of higher energy consumption. Computation is therefore dominated by this strong mean input, leading to the robust, high-fidelity rate code observed in our experiments.

\paragraph{3. The Performance Valley}
Our framework also provides insights for the performance degradation observed in the \textbf{Transition Region}: The primary cause is not the inefficiency of neural processing, but rather the degraded quality of the input signal itself. As shown in our initial analysis (Figure \ref{fig:phase_transition}A, D), this dynamical regime corresponds to a significant drop in AIS, indicating that the encoder generates trajectories with less structure and lower information content. The OU theory describes how a neuron responds to a statistically defined input, but if the input lacks meaningful, learnable patterns (i.e., low AIS), the network cannot establish a stable mapping. A direct evidence is that the input current statistics in the transition region are not inherently weak, in fact, both the mean and variance are slightly higher than in the high-performing dissipative mode (Table~\ref{tab:sup_data_summary}). 

\begin{figure}[ht!]
    \centering
    \includegraphics[width=\textwidth]{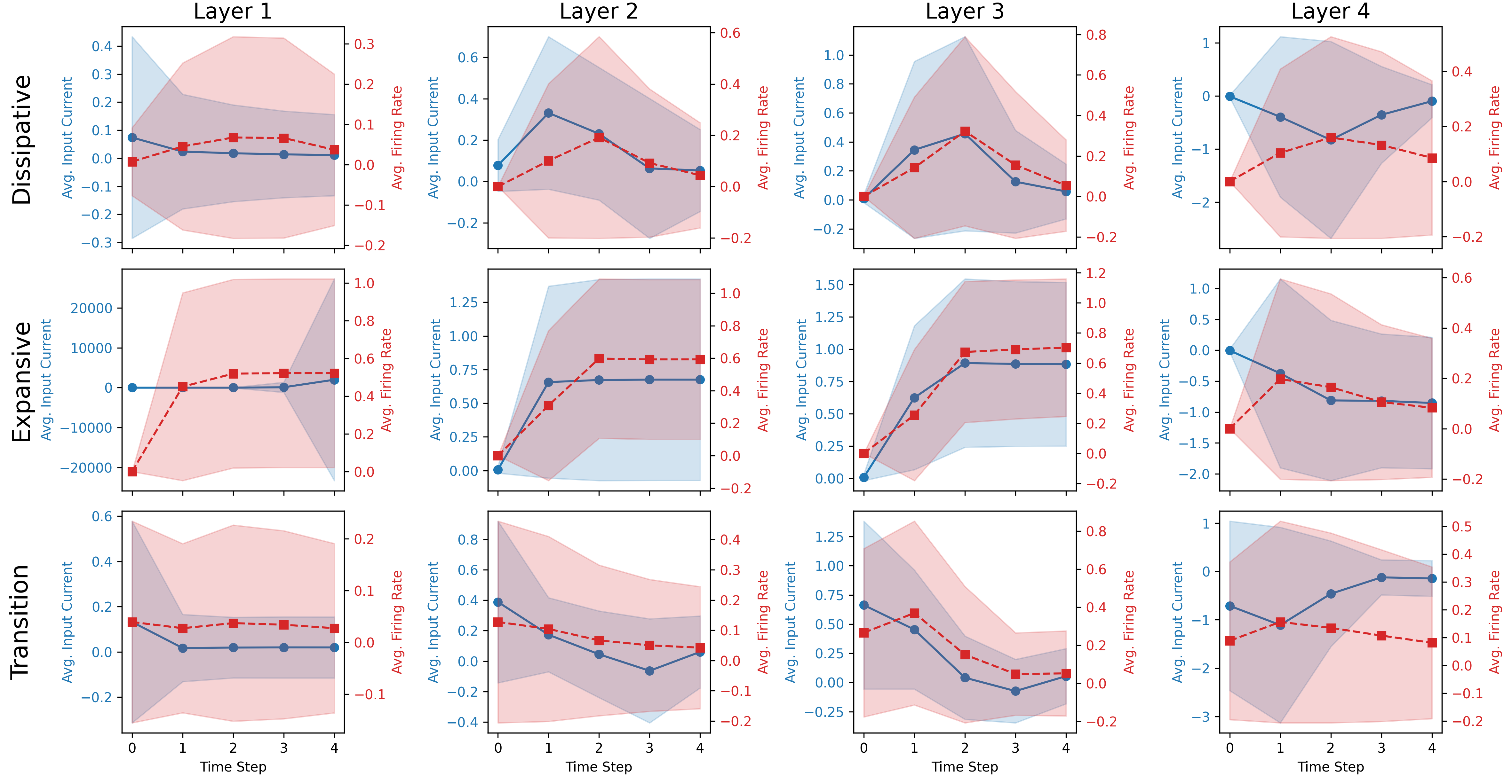}
    \caption{
        \textbf{Time-resolved internal dynamics across the three computational modes.} 
        Each row corresponds to a dynamical mode, and each column to a network layer. 
        Blue lines (left y-axis) show the mean input current ($\mu_I(t)$) over 5 time steps, with the shaded area representing $\pm 1$ standard deviation ($\sigma_I(t)$). 
        Red lines (right y-axis) show the mean firing rate ($\mu_\nu(t)$) with its standard deviation ($\sigma_\nu(t)$). 
        The plots visually confirm the stable, growing, and erratic dynamics of the dissipative, expansive, and transition region, respectively.
    }
    \label{fig:sup_dynamics}
\end{figure}

A further evidence is visible in the network's temporal evolution (Figure \ref{fig:sup_dynamics}): Unlike the dissipative and expansive modes, where firing rates build up in deeper layers over the first few timesteps (indicating successful signal integration), the transition mode exhibits an initial burst of activity that quickly decays as the signal propagates through the network. This suggests that while the initial input has sufficient energy to trigger firing, its lack of structure prevents the formation of self-sustaining, coordinated activity, causing the neural computation to dissipate. Ultimately, the network is caught in a strategy crisis: it receives a signal that is too unstructured for efficient fluctuation-driven coding. Table~\ref{tab:sup_data_summary} shows that the mean input current in the transition region's core layers (e.g., $\boldsymbol{\mu_I \approx 0.23}$ in Layer 3) is similar to the dissipative mode, and is insufficient to enable a robust mean-driven strategy. This manifests as a collapse in \textbf{information fidelity}, which we can quantify using the CV of the firing rate time series, $\nu(t)$:
\begin{equation}
    \text{CV}_{\nu} = \frac{\text{std}[\nu(t)]}{|\text{mean}[\nu(t)]| + \epsilon}
    \label{eq:cv}
\end{equation}
where:
\begin{itemize}
    \item $\nu(t)$: The firing rate of a neuron or a neural population over time.
    \item $\text{mean}[\nu(t)]$: The time-averaged mean of the firing rate, denoted as $\langle \nu \rangle$.
    \item $\text{std}[\nu(t)]$: The standard deviation of the firing rate over time.
    \item $\epsilon$: A small constant to prevent division by zero.
\end{itemize}
The CV serves as an inverse proxy for the signal-to-noise ratio (SNR) of the neural code, which can be understood from an information-theoretic perspective. The mutual information, $I(X;Y)$, between an input signal $X$ and the neural response $Y$ (represented by the firing rate $\nu$) is bounded by the channel capacity. For a channel with Gaussian noise, this is given by:
\begin{equation}
    I(X;Y) = H(Y) - H(Y|X) \leq \frac{1}{2}\log_2\left(1 + \frac{\text{var}(\text{signal})}{\text{var}(\text{noise})}\right)
\end{equation}
where $H(Y)$ is the entropy of the output and $H(Y|X)$ is the conditional entropy. For Poisson-like spike trains, where the signal is encoded in the mean firing rate $\langle \nu \rangle$ and the noise is captured by its variance $\text{Var}[\nu]$, this bound can be expressed in terms of the CV:
\begin{equation}
    I(X;Y) \leq \frac{1}{2}\log_2\left(1 + \frac{\langle \nu \rangle^2}{\text{Var}[\nu]}\right) = \frac{1}{2}\log_2\left(1 + \frac{1}{\text{CV}_\nu^2}\right)
\end{equation}
This relationship, often used in neuroscience, establishes $\text{CV}_\nu^{-1}$ as a measure of the code's reliability: a high $\text{CV}_{\nu}$ implies a noisy, unreliable code with low information content, as it corresponds to a lower upper-bound on the information that can be transmitted.

Our experimental data provides evidence for this fidelity collapse. As shown in Table~\ref{tab:sup_data_summary}, the transition region exhibits the highest mean Firing Rate CV ($\boldsymbol{\approx 3.36}$), significantly greater than both the expansive ($\boldsymbol{\approx 1.29}$) and dissipative ($\boldsymbol{\approx 2.99}$) modes. This high relative volatility is also visually apparent in Figure~\ref{fig:sup_dynamics}, which shows a different, dropping evolution of both mean currents and firing rates in this regime.

\begin{table}[ht!]
\centering
\caption{
    \textbf{Layer-wise time-averaged statistics of internal network dynamics across the three computational modes.} 
    Data was collected under the same experiment setup based on Section \ref{sec:phase_transition} from trained SNNs with fixed neuron parameters ($\beta=0.95, V_{\text{th}}=1.0$). 
    $\mu_I$ and $\sigma_I$ represent the mean and standard deviation of the input current to each layer, respectively. 
    $\mu_\nu$ is the corresponding statistics for the output firing rate. 
    The Firing Rate CV ($\sigma_\nu / |\mu_\nu|$) serves as a proxy for the inverse of the signal-to-noise ratio, quantifying the information fidelity of the neural code. 
    All values are averaged over 5 time steps and the entire test set.
}
\label{tab:sup_data_summary}
\resizebox{\textwidth}{!}{%
\begin{tabular}{@{}lcccc|cccc|cccc@{}}
\toprule
\multicolumn{1}{c}{} & \multicolumn{4}{c}{\textbf{Dissipative Mode}} & \multicolumn{4}{c}{\textbf{Expansive Mode}} & \multicolumn{4}{c}{\textbf{Transition Region}} \\
\cmidrule(lr){2-5} \cmidrule(lr){6-9} \cmidrule(lr){10-13}
\textbf{Layer} & $\mu_I$ & $\sigma_I$ & $\mu_\nu$ & \textbf{Rate CV} & $\mu_I$ & $\sigma_I$ & $\mu_\nu$ & \textbf{Rate CV} & $\mu_I$ & $\sigma_I$ & $\mu_\nu$ & \textbf{Rate CV} \\
\midrule
Layer 1 & 0.028 & 0.207 & 0.044 & \textbf{4.408} & 420.783 & 5327.060 & 0.402 & \textbf{0.992} & 0.041 & 0.200 & 0.033 & \textbf{5.384} \\
Layer 2 & 0.151 & 0.270 & 0.086 & \textbf{2.777} & 0.537 & 0.596 & 0.418 & \textbf{0.925} & 0.121 & 0.327 & 0.078 & \textbf{3.347} \\
Layer 3 & 0.199 & 0.370 & 0.135 & \textbf{2.086} & 0.659 & 0.501 & 0.465 & \textbf{0.785} & 0.227 & 0.419 & 0.178 & \textbf{1.939} \\
Layer 4 & -0.338 & 0.921 & 0.096 & \textbf{2.679} & -0.573 & 1.000 & 0.110 & \textbf{2.456} & -0.509 & 1.120 & 0.113 & \textbf{2.772} \\
\bottomrule
\end{tabular}%
}
\end{table}

In conclusion, the performance valley is not caused by a mere lack of activity, but by the poor quality and low information content of that activity. It is a direct consequence of the network's inability to settle into an efficient coding strategy that matches the input's intermediate temporal statistics, leading to a fundamental corruption of information during neural processing. While our mean-field theory provides valuable insights into the time-scale matching principle, future work should incorporate the full nonlinear dynamics and network interactions for a complete understanding.

\subsection{Analysis of Different Dynamical Regimes}
\label{app:d4}

Based on the bimodal discovery presented in Figure~\ref{fig:phase_transition}, we selected four key dynamical regimes from Table~\ref{table:d1_oscillator_data} for a mechanistic analysis: the dissipative regime ($\delta=10.0$), the transition regime ($\delta=2.0$), the critical point ($\delta=0.0$), and the expansive regime ($\delta=-1.5$). The SNN and MLP (ANN) models used the same network architecture as described in Appendix~\ref{app:A} to ensure a fair comparison. All models were trained under the same dataset, hyperparameters, and training protocol detailed in Appendix \ref{app:d2}. Each model was trained for 10 independent runs to ensure statistical reliability. The results are shown in Table~\ref{table:d2_snn_vs_mlp}.

\begin{table}[h!]
\centering
\caption{Performance comparison between SNN and MLP across different dynamical regions (n=10)}
\label{table:d2_snn_vs_mlp}

\begin{tabular}{@{}lccccc@{}}
\toprule
\textbf{Region} & \textbf{$\delta$} & \textbf{MLP (\%)} & \textbf{SNN (\%)} & \textbf{Difference} & \textbf{p-value} \\ \midrule
Dissipative & 10.0 & 93.06 $\pm$ 0.94 & 92.92 $\pm$ 1.12 & -0.14 & 0.470 \\
Transition  & 2.0  & 92.71 $\pm$ 1.08 & 91.25 $\pm$ 1.12 & -1.46** & 0.007 \\
Critical    & 0.0  & 93.12 $\pm$ 0.88 & 92.82 $\pm$ 0.72 & -0.30 & 0.426 \\
Expansive   & -1.5 & 91.79 $\pm$ 1.77 & 93.94 $\pm$ 1.02 & +2.15** & 0.006 \\ \bottomrule
\end{tabular}
\par
\small{** $p < 0.01$}
\end{table}

We systematically evaluated the impact of different dynamical encodings on SNN information processing across five dimensions: (1) comparative classification performance (including against MLP), (2) spatiotemporal patterns of neuronal activity, (3) quantification of collective dynamics, (4) network robustness analysis, and (5) assessment of information representation capabilities. The results are summarized in Figure~\ref{fig:neural_dynamics}. Given that neural network performance data may not follow a normal distribution, we used the non-parametric Mann-Whitney U test to assess statistical significance between different conditions.

\subsubsection{Neuronal Activity Pattern Analysis (Figure~\ref{fig:neural_dynamics}B-C)}
During the model's forward pass, we recorded the information propagation trajectory of a single representative sample through the network. We fully recorded the spike sequences of all four LIF neuron layers over 5 time steps. The state of a neuron can be represented as $S_{n,t}^l \in \{0, 1\}$, where $n$ is the neuron index, $l$ is the network layer, and $t \in \{1, ..., T\}$ is the time step ($N=5$). A raster plot visualizes this three-dimensional activity matrix, where a black pixel represents a spike event $S_{n,t}^l=1$. By tracking the propagation of spike signals between network layers, we observed that the dissipative regime exhibits sparse, precise spike patterns, whereas the expansive regime shows dense, highly active patterns.

\subsubsection{Quantification of Collective Dynamics}
To quantify the differences in collective dynamics between the modes, we calculated three core metrics: neuron firing rate, coefficient of variation for synchrony, and neuronal correlation.

\paragraph{Neuron Firing Rate (Figure~\ref{fig:neural_dynamics}D)} is obtained by calculating the average spike density across all time steps and batch samples:
\begin{align*}
    r_l = \frac{1}{T \cdot B \cdot N_l} \sum_{i=1}^{B} \sum_{j=1}^{N_l} \sum_{t=1}^{T} S_{i,j,t}^l
\end{align*}
where $N_l$ is the number of neurons in layer $l$, $B$ is the batch size, and $T$ is the total number of time steps. The results (Figure~\ref{fig:neural_dynamics}D) show that the dissipative regime ($\delta=10.0$) exhibits highly sparse spike patterns with a spike density of approximately 0.05-0.15. In contrast, the expansive regime ($\delta=-1.5$) displays dense activity with a significantly higher spike density of 0.38-0.46. These distinct firing patterns are mechanistically explained by the different input statistics learned by the network in each mode, as detailed in Table~\ref{tab:sup_data_summary}, where the high-gain inputs ($\mu_I, \sigma_I$) of the expansive mode drive its dense activity. The statistical robustness of these firing rates over multiple runs is confirmed in Table~\ref{table:d3_firing_rate}.

\begin{table}[h!]
\centering
\caption{Neuron firing rate statistics in different dynamical regimes (n=10)}
\label{table:d3_firing_rate}
\begin{tabular}{@{}lccccc@{}}
\toprule
\textbf{Dynamical Regime} & \textbf{$\delta$} & \textbf{Layer 1} & \textbf{Layer 2} & \textbf{Layer 3} & \textbf{Layer 4} \\ \midrule
Expansive & -1.5 & 0.383 $\pm$ 0.009 & 0.429 $\pm$ 0.006 & 0.463 $\pm$ 0.010 & 0.093 $\pm$ 0.006 \\
Critical  & 0.0  & 0.081 $\pm$ 0.006 & 0.189 $\pm$ 0.029 & 0.249 $\pm$ 0.045 & 0.107 $\pm$ 0.009 \\
Transition& 2.0  & 0.043 $\pm$ 0.003 & 0.097 $\pm$ 0.006 & 0.162 $\pm$ 0.023 & 0.125 $\pm$ 0.010 \\
Dissipative& 10.0 & 0.050 $\pm$ 0.004 & 0.102 $\pm$ 0.006 & 0.152 $\pm$ 0.005 & 0.105 $\pm$ 0.005 \\ \bottomrule
\end{tabular}
\par
\small{Note: Values represent mean $\pm$ standard deviation. Firing rate is defined as spike density.}
\end{table}

\paragraph{Temporal Synchrony (CV, Figure~\ref{fig:neural_dynamics}E)}
To quantify the degree to which population activity is synchronized in time, we calculated the CV of the population firing rate across discrete time steps. This metric is distinct from the Rate CV used to measure information fidelity (Eq. \ref{eq:cv}) . It is calculated as:
\begin{align*}
    a_t^l = \frac{1}{N_l} \sum_{j=1}^{N_l} S_{j,t}^l \quad \text{and} \quad CV_l = \frac{\sigma_{a_t^l}}{\mu_{a_t^l}}
\end{align*}
where $a_t^l$ is the average activation rate across all neurons in layer $l$ at a single time step $t$. The mean $\mu_{a_t^l}$ and standard deviation $\sigma_{a_t^l}$ are then computed over the temporal dimension ($t=1, ..., T$). A higher CV value in this context indicates a greater modulation of population activity over time; that is, neural activity is concentrated within specific temporal windows, reflecting a higher degree of synchrony, rather than being uniformly distributed.

\paragraph{Neuronal Correlation Analysis (Figure~\ref{fig:neural_dynamics}F)} 
To quantify the spatial correlation of spike activity between neurons, we use the Pearson correlation coefficient, $\rho_{jk}^l$, calculated for each pair of neurons ($j, k$) within a layer $l$:
\begin{align*}
    \rho_{jk}^l = \frac{\text{Cov}(S_j^l, S_k^l)}{\sigma_{S_j^l} \sigma_{S_k^l}}
\end{align*}
where $S_j^l$ is the spike train (a time series of 0s and 1s) of neuron $j$ in layer $l$, $\text{Cov}(\cdot, \cdot)$ is the covariance, and $\sigma_{S_j^l}$ is the standard deviation of that spike train over time.

The layer-wise average correlation, $\bar{\rho}^l$, is then defined as the mean absolute correlation value over all pairs of active neurons in that layer:
\begin{align*}
    \bar{\rho}^l = \frac{1}{N(N-1)} \sum_{j \neq k} |\rho_{jk}^l|
\end{align*}
where $N$ is the number of active neurons considered. To avoid the influence of silent neurons, only active neurons with a spike train standard deviation $\sigma_{S_j} > \epsilon$, where $\epsilon = 10^{-8}$, were included in this calculation. The results for both Temporal Synchrony and Neuronal Correlation are presented in Table~\ref{table:d4_collective_dynamics}.

\begin{table}[h!]
\centering
\caption{Summary of collective dynamics metrics (n=10)}
\label{table:d4_collective_dynamics}
\begin{tabular}{@{}llcc@{}}
\toprule
\textbf{Dynamical Regime} & \textbf{Layer} & \textbf{CV for Synchrony} & \textbf{Neuronal Correlation} \\ \midrule
\multirow{4}{*}{Expansive} & Layer 1 & 0.56 $\pm$ 0.00 & 0.42 $\pm$ 0.01 \\
 & Layer 2 & 0.63 $\pm$ 0.00 & 0.36 $\pm$ 0.01 \\
 & Layer 3 & 0.67 $\pm$ 0.01 & 0.37 $\pm$ 0.01 \\
 & Layer 4 & 0.65 $\pm$ 0.02 & 0.11 $\pm$ 0.01 \\ \midrule
\multirow{4}{*}{Dissipative} & Layer 1 & 0.52 $\pm$ 0.02 & 0.18 $\pm$ 0.01 \\
 & Layer 2 & 0.76 $\pm$ 0.05 & 0.16 $\pm$ 0.01 \\
 & Layer 3 & 0.92 $\pm$ 0.05 & 0.20 $\pm$ 0.01 \\
 & Layer 4 & 0.67 $\pm$ 0.03 & 0.13 $\pm$ 0.01 \\ \bottomrule
\end{tabular}
\end{table}

\subsubsection{Network Robustness to Perturbations (Figure~\ref{fig:neural_dynamics}G)}
We designed a progressive spike deletion experiment to assess the network's robustness to perturbations. In real biological neural networks, neurons face two primary types of perturbations: signal loss and spurious activation. We chose spike deletion as the perturbation model because it is more biologically plausible. Under conditions of metabolic constraint, hypoxia, fatigue, or pathology, neurons are more likely to exhibit functional deficits or synaptic transmission failures rather than abnormal over-activation\cite{attwell2001energy}.

During the forward pass, we randomly set spikes that should have fired in the hidden layers (L1-L3) to zero with a probability $p \in [0, 0.8]$. The performance at each perturbation level was evaluated 10 times to obtain a stable estimate. Network robustness was assessed by calculating the drop in accuracy at different perturbation levels. The two dynamical states exhibited starkly different robustness decay curves (Table~\ref{table:d5_robustness}, Figure~\ref{fig:neural_dynamics}G): the expansive regime showed a \textit{trapezoidal decay}, maintaining stable performance up to a certain threshold and retaining relatively high performance even at a 40\% spike deletion rate. In contrast, the dissipative regime showed an \textit{exponential decay}, with performance dropping rapidly with perturbation intensity and showing significant decline even at low levels (20\%). This difference reveals the fundamental distinction in fault tolerance between dense and sparse coding schemes.

\begin{table}[h!]
\centering
\caption{Quantitative results of spike deletion robustness in different dynamical modes (n=10).}
\label{table:d5_robustness}
\resizebox{0.8\textwidth}{!}{%
\begin{tabular}{@{}ccccc@{}}
\toprule
\textbf{\makecell{Spike Deletion \\ Rate (p)}} &
\textbf{\makecell{Expansive Mode \\ Mean Acc. (\%)} } &
\textbf{\makecell{Expansive Mode \\ Acc. Drop (\%)} } &
\textbf{\makecell{Dissipative Mode \\ Mean Acc. (\%)} } &
\textbf{\makecell{Dissipative Mode \\ Acc. Drop (\%)} } \\ \midrule
0.00 & 93.35 $\pm$ 0.00 & 0.00 & 91.03 $\pm$ 0.00 & 0.00 \\
0.01 & 93.26 $\pm$ 0.24 & 0.09 & 90.60 $\pm$ 0.39 & 0.43 \\
0.02 & 93.31 $\pm$ 0.17 & 0.04 & 89.95 $\pm$ 0.19 & 1.08 \\
0.05 & 92.96 $\pm$ 0.18 & 0.39 & 86.59 $\pm$ 0.82 & 4.44 \\
0.10 & 92.88 $\pm$ 0.19 & 0.47 & 79.24 $\pm$ 1.23 & 11.79 \\
0.15 & 92.37 $\pm$ 0.29 & 0.98 & 70.54 $\pm$ 0.86 & 20.49 \\
0.20 & 92.17 $\pm$ 0.49 & 1.18 & 59.86 $\pm$ 0.49 & 31.17 \\
0.30 & 91.17 $\pm$ 0.38 & 2.18 & 33.47 $\pm$ 1.26 & 57.56 \\
0.40 & 88.20 $\pm$ 0.34 & 5.15 & 13.13 $\pm$ 0.56 & 77.90 \\
0.50 & 80.77 $\pm$ 0.87 & 12.58 & 10.06 $\pm$ 0.08 & 80.97 \\
0.60 & 22.34 $\pm$ 0.86 & 71.01 & 9.98 $\pm$ 0.00 & 81.05 \\
0.70 & 9.88 $\pm$ 0.00 & 83.47 & 9.98 $\pm$ 0.00 & 81.05 \\
0.80 & 9.88 $\pm$ 0.00 & 83.47 & 9.98 $\pm$ 0.00 & 81.05 \\ \bottomrule
\end{tabular}
}
\end{table}

\subsubsection{Analysis of Information Representation}
To assess the quality and characteristics of the neural representations in each layer, we quantified the strength of the association between each layer's representation and the classification labels.

\paragraph{Mutual Information with Labels (Figure~\ref{fig:neural_dynamics}H)} 

The mutual information between layer $l$ and the labels $Y$ is defined as:
\begin{align*}
    I(T^l; Y) = \sum_{t^l, y} p(t^l, y) \log \frac{p(t^l, y)}{p(t^l)p(y)}
\end{align*}
where $T^l = \sum_{t=1}^T S_t^l$ is the aggregated spike count representation of layer $l$, and $p(t^l, y)$ is the joint probability distribution. To compute the mutual information, continuous neuron activity values were discretized using a binning method (KBinsDiscretizer with $K=20$ bins). For high-dimensional representations, PCA was used for pre-dimensionality reduction to 10 dimensions to balance computational efficiency and information preservation\cite{bishop2006pattern}. The expansive regime ($\delta=-1.5$) exhibits higher label-related information starting from the input layer (L1), reflecting that dense chaotic encoding can establish stronger input-label associations in the shallower layers. This difference is amplified at the output layer (L4), indicating higher overall information transmission efficiency in the expansive regime. In contrast, while the dissipative regime ($\delta=10.0$) shows more stable information transmission, the absolute amount of label-related information is lower, reflecting the conservative nature of sparse coding in information acquisition.

\paragraph{Intrinsic Dimensionality Estimation (Figure~\ref{fig:neural_dynamics}I)} We used Principal Component Analysis (PCA) to estimate the effective dimensionality of each layer's temporal spike representation. The complete temporal spike sequence was flattened into a vector $s^l = (S_{: ,1}^l, S_{:, 2}^l, ..., S_{:, T}^l)^T \in \mathbb{R}^{N \times T}$, where $N$ is the number of neurons. The 95\% variance dimensionality metric is defined as:
\begin{align*}
    ID_{95}(l) = \min \left( k \mid \frac{\sum_{i=1}^k \lambda_i^l}{\sum_{i=1}^{NT} \lambda_i^l} \ge 0.95 \right)
\end{align*}
where $\lambda_i^l$ is the $i$-th eigenvalue of the covariance matrix of the representation for layer $l$. By analyzing the cumulative explained variance, we determine the number of principal components needed to explain 95\% of the variance as the intrinsic dimensionality metric. The expansive regime shows higher representational dimensionality in deeper layers, reflecting its rich feature learning capacity, while the dissipative regime maintains a relatively low-dimensional, compact representation. The results are shown in Table~\ref{table:d6_pca_dim}.

\begin{table}[h!]
\centering
\caption{Intrinsic dimensionality of neural representations (PCA 95\% Var, n=10)}
\label{table:d6_pca_dim}
\begin{tabular}{@{}llc@{}}
\toprule
\textbf{Dynamical Regime} & \textbf{Layer} & \textbf{Intrinsic Dimension} \\ \midrule
\multirow{4}{*}{Expansive} & Layer 1 & 58.7 $\pm$ 3.5 \\
 & Layer 2 & 327.9 $\pm$ 9.6 \\
 & Layer 3 & 385.6 $\pm$ 9.4 \\
 & Layer 4 & 12.1 $\pm$ 1.7 \\ \midrule
\multirow{4}{*}{Dissipative} & Layer 1 & 66.5 $\pm$ 4.5 \\
 & Layer 2 & 190.0 $\pm$ 13.0 \\
 & Layer 3 & 283.0 $\pm$ 16.1 \\
 & Layer 4 & 14.6 $\pm$ 1.4 \\ \bottomrule
\end{tabular}
\end{table}

\subsubsection{Additional Supplementary Analysis}
In addition to the visualizations presented in the main text, we conducted further supplementary analyses on the experimental data, including linear separability of neural representations and information bottleneck trajectory tracking. We focused on three key dynamical regimes: expansive, dissipative, and transition, to explore the neural computation characteristics under different dynamical encoding conditions.

\paragraph{Linear Separability Analysis} This analysis aims to answer two questions: \textit{1. Does dynamical encoding disrupt the separable structure of neural representations? 2. Can it explain the observed differences in classification performance?} These questions are crucial for understanding the effectiveness of chaotic encoding and for guiding its practical application. 

We used a \textbf{Linear Probe} method to assess the linear separability of the representation space, quantifying the feature learning complexity of each layer's representation. Specifically, we trained a simple logistic regression classifier on the spike count representation $T^l$ of each layer. We used \texttt{StandardScaler} for normalization and stratified splits to ensure reliable evaluation. The classification accuracy directly reflects the degree of linear separability. Higher separability indicates that the representation can be effectively classified by a simple linear transformation, while lower separability suggests that more complex nonlinear transformations are required.

The results, shown in Table~\ref{table:d7_linearity}, indicate that all dynamical regimes maintain good linear separability (89-95\%), and this separability is highly correlated with the corresponding neural computation performance. This finding suggests that dynamical encoding primarily modulates information processing at the spatiotemporal statistical level, rather than destroying the intrinsic separability of features. Detailed layer-wise analysis reveals the feature learning patterns of the different regimes:
\begin{itemize}
    \item \textbf{Expansive Regime ($\delta=-1.5$)}: Linear separability remains at a high level of 94-95\% across all layers, with a slight upward trend from L1 to L4 (0.94$\rightarrow$0.95), indicating that the representations generated by dense spike encoding have an excellent linear structure.
    \item \textbf{Dissipative Regime ($\delta=10.0$)}: Linear separability is stable at 93-94\% with little variation between layers, reflecting the consistency and stability of sparse coding representations.
    \item \textbf{Transition Regime ($\delta=2.0$)}: Linear separability is relatively lower (89-91\%) and fluctuates in layers L2-L3, reflecting the instability of feature learning in the critical dynamical state.
\end{itemize}

\begin{table}[h!]
\centering
\caption{Linear separability assessment across different dynamical regimes}
\label{table:d7_linearity}
\begin{tabular}{@{}llc@{}}
\toprule
\textbf{Condition} & \textbf{Layer} & \textbf{Linear Separability} \\ \midrule
\multirow{4}{*}{Expansive Region} & layer1 & 0.94 \\
 & layer2 & 0.94 \\
 & layer3 & 0.94 \\
 & layer4 & 0.95 \\ \midrule
\multirow{4}{*}{Dissipative Region} & layer1 & 0.93 \\
 & layer2 & 0.94 \\
 & layer3 & 0.93 \\
 & layer4 & 0.94 \\ \midrule
\multirow{4}{*}{Transition Region} & layer1 & 0.89 \\
 & layer2 & 0.90 \\
 & layer3 & 0.91 \\
 & layer4 & 0.90 \\ \bottomrule
\end{tabular}
\end{table}

\paragraph{Information Bottleneck Analysis} To gain a deeper, information-theoretic understanding of how different dynamical encodings affect the network's learning process, we introduced the Information Bottleneck (IB) theory as our analytical framework. The IB theory posits that the learning process of an excellent deep learning model involves finding an optimal balance on the Information Plane: on one hand, maximally \textit{compressing} the input signal (i.e., discarding task-irrelevant information), and on the other hand, maximally \textit{preserving} information relevant to the classification labels for prediction.

Based on Tishby's IB principle\cite{tishby2000information}, we tracked the network's information processing trajectory during learning in a two-dimensional information plane. This plane is defined by a prediction axis, $I(T^l; Y)$ (calculation method described before in 'Mutual Information with Labels'), and a compression axis, $I(T^l; T^0)$, where $T^0$ represents the input representation after dynamical encoding. The mutual information between layer representations and the input/labels is calculated as:
\begin{align*}
    I(T^l; T^0) = \sum_{t^l, t^0} p(t^l, t^0) \log \frac{p(t^l, t^0)}{p(t^l)p(t^0)}
\end{align*}
To ensure the stability and accuracy of the MI estimation, we used a binning method for data discretization and pre-emptively reduced the dimensionality of high-dimensional representations using PCA (consistent with the method described in the 'Information Representation Analysis' section).

\begin{table}[h!]
\centering
\caption{Position of layer representations on the information plane at the end of training for different dynamical regimes}
\label{table:d8_infobottleneck}
\begin{tabular}{@{}lccc@{}}
\toprule
\textbf{Condition} & \textbf{Expansive Region} & \textbf{Dissipative Region} & \textbf{Transition Region} \\ \midrule
layer1\_Input\_MI & 6.65 & 5.92 & 4.94 \\
layer1\_Label\_MI & 2.29 & 2.23 & 2.09 \\
layer2\_Input\_MI & 6.78 & 6.73 & 6.32 \\
layer2\_Label\_MI & 2.30 & 2.29 & 2.27 \\
layer3\_Input\_MI & 6.82 & 6.72 & 6.52 \\
layer3\_Label\_MI & 2.29 & 2.29 & 2.28 \\
layer4\_Input\_MI & 3.08 & 3.34 & 3.61 \\
layer4\_Label\_MI & 2.12 & 2.11 & 2.04 \\ \bottomrule
\end{tabular}
\end{table}

We observed that different dynamical encoding modes exhibit distinct learning trajectories and final convergence states on the information plane (Table~\ref{table:d8_infobottleneck}).

\textbf{Efficient Learning Strategies (Expansive and Dissipative Modes)}:
\begin{itemize}
    \item The final output layers (layer4) of both the expansive and dissipative modes achieve similarly high predictive power ($I(T^4;Y)$ of 2.12 and 2.11, respectively), indicating that they both successfully learned the key information required to complete the task.
    \item Notably, in the shallower layers (layer1), the expansive mode retains more information about the input ($I(T^1;X)=6.65$) compared to the dissipative mode (5.92). This aligns with our earlier observation that expansive encoding generates a 'richer' initial representation, while dissipative encoding is more 'compact'.
    \item From L3 to L4, both efficient modes exhibit significant \textit{compression}: $I(T;X)$ drops sharply (e.g., from 6.82 to 3.08 in the expansive mode) while maintaining a high level of $I(T;Y)$. This is precisely the ideal learning objective described by IB theory—discarding redundant information while retaining the task-relevant essence.
\end{itemize}

\textbf{Inefficient Learning Strategy (Transition Region)}:
\begin{itemize}
    \item The transition region performs poorly at all levels. Its output layer's predictive information ($I(T^4;Y)=2.04$) is markedly lower than the other two modes, which explains its lower classification accuracy from an information-theoretic perspective.
    \item Its compression process is also less efficient. The final output layer retains the most input information ($I(T^4;X)=3.61$), much of which is likely task-irrelevant, indicating the network failed to effectively distinguish and discard redundant information.
\end{itemize}
The conclusion from the information bottleneck analysis is that both the expansive and dissipative modes are capable of finding an excellent 'bottleneck' position on the information plane, achieving efficient compression of input information and effective preservation of task-relevant information. In contrast, the transition region represents an inefficient learning dynamic, where the network's representations are neither sufficiently compressed nor sufficiently predictive, leading to poor computational performance. This information-theoretic failure is the direct counterpart to the mechanistic collapse detailed in our timescale alignment analysis (Appendix \ref{sec:appendix_math_model}), where the network's inability to form a coherent coding strategy results in a high-CV, low-fidelity neural code.

Ultimately, integrating all analytical results shows that the expansive and dissipative modes represent two different, yet highly efficient, computational strategies based on their learning dynamics. The dense chaotic encoding of the expansive regime establishes stronger label correlations from the very first layers and maintains a higher representational dimension throughout the network depth. This pattern indicates that while dense spike activity consumes more computational resources, it provides richer representational capacity. The dissipative regime, on the other hand, achieves a balance between computational efficiency and information preservation through low-dimensional, sparse representations.

\printbibliography[title={References for Appendix D}, heading=subbibliography, env=appendixbib]
\end{refsection}


\section{Detailed Experimental Setup for Universality Validation}
\label{app:e}
\begin{refsection}
\setappendixprefix{E}

This appendix details the setup for the three cross-domain experiments used to validate the universality of the 'dynamical alignment' principle, as described in Section~\ref{sec:cross_domain_validation}. In all experiments, we compared three key dynamical regimes identified in Appendix~\ref{app:d4}: the expansive regime ($\delta=-1.5, \Sigma\lambda_i \approx +2$), the transition regime ($\delta=2.0, \Sigma\lambda_i \approx -4$), and the dissipative regime ($\delta=10.0, \Sigma\lambda_i \approx -15$).

\subsection{Scalability on ResNet-18 with TinyImageNet}
\label{app:e1}

\paragraph{Experimental Design Rationale} This experiment was designed to assess the scalability and effectiveness of the 'dynamical alignment' principle on modern deep learning architectures. It aimed to answer a key scientific question: \textit{Do the bimodal properties of dynamical encoding remain effective in modern deep architectures, independent of network complexity?} The core hypothesis is that if dynamical alignment reveals a general principle of neural computation, its advantages should be more fully realized in modern architectures with sufficient expressive capacity. This hypothesis is based on the following considerations: (1) Simple architectures may not be able to fully exploit the rich spatiotemporal information provided by dynamical encoding, whereas modern deep architectures have enough parametric capacity to learn complex temporal patterns. (2) Using a powerful visual feature extraction backbone allows us to focus on evaluating the contribution of dynamical encoding to neural computation. (3) By keeping the backbone network unchanged and only inserting the dynamical transformation after the feature projection layer, we can precisely isolate the effect of dynamical encoding and verify its independence from specific architectural designs. Unlike the previous validations on simple fully-connected architectures (Appendices~\ref{app:A}, \ref{app:B}), we tested performance on a more complex and larger-scale dataset by integrating the dynamical encoding module with a ResNet-18 model\cite{he2016deep} pretrained on ImageNet\cite{deng2009imagenet}.

\paragraph{Dataset and Preprocessing} This experiment used the TinyImageNet dataset as the evaluation benchmark. TinyImageNet is a reduced version of ImageNet, containing 200 classes, with 500 training images and 50 validation images per class, each of size 64$\times$64 pixels. Compared to MNIST and CIFAR-10, TinyImageNet's complexity in terms of class number and visual features is closer to real-world scenarios, providing a suitable challenge for evaluating the effectiveness of dynamical encoding in complex visual tasks. Data preprocessing followed standard ImageNet best practices:
\begin{itemize}
    \item \textbf{Training set augmentation}:
        \begin{itemize}
            \item Images resized to 224$\times$224 pixels (to match ResNet-18 pretraining dimensions).
            \item Random resized crop (scale 0.8-1.0).
            \item Random horizontal flip.
            \item ImageNet normalization (mean: [0.485, 0.456, 0.406], std: [0.229, 0.224, 0.225]).
        \end{itemize}
    \item \textbf{Validation set processing}:
        \begin{itemize}
            \item Images resized to 224$\times$224 pixels.
            \item Center crop.
            \item ImageNet normalization.
        \end{itemize}
\end{itemize}
A batch size of 64 was used to balance memory efficiency and gradient estimation stability.

\paragraph{Network Architecture Design} All models shared a common feature extraction backbone to ensure a fair comparison: a ResNet-18 model pretrained on ImageNet-1K was used as the feature extractor. We removed the final fully-connected classification layer of the original ResNet-18 and replaced it with an `nn.Identity()` layer, resulting in a backbone that outputs a 512-dimensional feature vector.

Similar to the bridging experiment (Experiment 2, Appendix~\ref{app:B}), we used a linear layer to project the 512-dimensional features from ResNet-18 to a specified dimension. The dimension values were chosen from [32, 64, 128, 256, 512] to investigate encoding properties and computational performance across different dimensionalities.

For dynamical encoding, we used the mixed oscillator system described in Appendix \ref{app:d1}. The projected data was normalized using a \texttt{tanh} activation function and served as input to the dynamical encoder. The time evolution steps ($N$) for all dynamical encodings were fixed at 5. The resulting spatiotemporal sequences were processed by a spiking classifier head composed of \textbf{two hidden layers of \texttt{snn.Leaky} neurons}, followed by a final \textbf{linear readout layer}. This readout layer integrates the spikes from the last hidden layer over time to produce the class logits. The membrane potential decay factor for the leaky neurons was set to $\beta=0.95$, and their operating mechanism (e.g., soft reset) was consistent with the LIF neurons described in Appendix~\ref{app:a2}. The model architecture was: ResNet-18 + Projection Layer + Oscillator Transform + Spiking Classifier Head.

We also prepared three baseline models for comparison:
\begin{itemize}
    \item \textbf{Direct-SNN}: ResNet-18 + Projection Layer + the same Spiking Classifier Head described above. The dynamical transformation was skipped, and the projected features were repeatedly fed into the spiking layers over 5 time steps.
    \item \textbf{Isomorphic ANN}: ResNet-18 + Projection Layer + an ANN classifier head. To ensure structural equivalence, this head consisted of two hidden fully-connected layers (\texttt{nn.Linear} + \texttt{ReLU}), followed by a final linear classifier layer.
    \item \textbf{Standard ResNet-18}: The original pretrained model.
\end{itemize}

\paragraph{Training Protocol and Evaluation Metrics} Given that the pretrained backbone network already contained rich visual feature knowledge, we employed a differential learning rate strategy: the learning rate for the backbone was $1 \times 10^{-5}$, while the learning rate for the projection, fully-connected, and SNN layers was $1 \times 10^{-4}$. This strategy preserves the effectiveness of the pretrained features while allowing the newly added layers to learn sufficiently. 

For training configuration, we used cross-entropy loss (\texttt{nn.CrossEntropyLoss}), the Adam optimizer (weight decay $5 \times 10^{-4}$), and a cosine annealing learning rate schedule (\texttt{CosineAnnealingLR}, $T_{\max}=200$). Training was terminated with an early stopping policy if there was no improvement in validation accuracy for 15 consecutive epochs. 

For SNN models, the output was determined by accumulating spikes over all time steps: $y = \text{argmax}_j \sum_{t=1}^T S_{j,t}$.

The model's performance was evaluated on three dimensions: accuracy, energy efficiency (spike count), and training efficiency (calculation methods are described in Appendix \ref{app:a5}). We also recorded the training time required for each model under the same environment. All reported metrics were from the best model checkpoint on the validation set during training. 

The experimental results, as shown in Table~\ref{table:e1_resnet_results}, revealed that under all tested feature dimensions, our dynamically encoded SNNs (ResNet-18 + SNN Head (Dynamic)) significantly outperformed their structurally equivalent ANNs (ResNet-18 + MLP Head) in accuracy, with performance improvements ranging from 2.46\% to 6.21\%. This contrasts with the typical underperformance of SNNs in shallow architectures and demonstrates that the advantages of temporal information processing are fully realized when the network has sufficient expressive capacity. 

Furthermore, we observed that in low-dimensional feature spaces (32-128), the expansive mode was significantly superior to the dissipative mode. In high-dimensional spaces (512-1024), the performance of the two modes converged, but they retained their respective energy efficiency characteristics. The expansive mode achieved the fastest convergence across all dimensions (2-3 times faster than the standard SNN and isomorphic MLP), while the dissipative mode achieved the best balance of performance and energy efficiency in high-dimensional feature spaces. The optimal SNN model (at 512 feature dimensions) achieved an accuracy of 74.66\%, narrowing the performance gap with the ResNet-18 baseline (75.26\%) to less than 0.6 percentage points. This substantial gap closure provides convincing evidence for the effectiveness of the dynamical encoding method.

\begin{table}[h!]
\centering
\caption{Cross-scale performance validation of dynamical encoding in a deep architecture (ResNet-18/TinyImageNet)}
\label{table:e1_resnet_results}
\resizebox{\textwidth}{!}{%
\begin{tabular}{@{}llcccc@{}}
\toprule
\textbf{\makecell{Backbone \\ Feature Dimension}} & 
\textbf{Model Head} & 
\textbf{\makecell{Accuracy \\ (\%)} } &
\textbf{\makecell{Relative Spikes \\ (vs Direct SNN)} } &
\textbf{\makecell{Relative Training \\ Time (vs ANN)} } &
\textbf{\makecell{Epochs to \\ Convergence}} \\ \midrule
\multirow{4}{*}{32} & MLP (ANN) & 61.61 & - & 1.00x & 40 \\
 & SNN (Direct) & 67.82 & 1.00x & 1.84x & 86 \\
 & SNN (Osc., Exp.) & 68.61 & 1.38x & 0.77x & 27 \\
 & SNN (Osc., Diss.) & 68.54 & 1.20x & 1.67x & 76 \\ \midrule
\multirow{4}{*}{64} & MLP (ANN) & 64.87 & - & 1.00x & 32 \\
 & SNN (Direct) & 70.47 & 1.00x & 2.04x & 69 \\
 & SNN (Osc., Exp.) & 71.65 & 1.35x & 0.83x & 19 \\
 & SNN (Osc., Diss.) & 71.45 & 1.18x & 1.10x & 30 \\ \midrule
\multirow{4}{*}{128} & MLP (ANN) & 67.67 & - & 1.00x & 21 \\
 & SNN (Direct) & 72.09 & 1.00x & 1.53x & 40 \\
 & SNN (Osc., Exp.) & 73.17 & 1.24x & 0.80x & 14 \\
 & SNN (Osc., Diss.) & 73.06 & 1.10x & 0.95x & 19 \\ \midrule
\multirow{4}{*}{256} & MLP (ANN) & 69.91 & - & 1.00x & 16 \\
 & SNN (Direct) & 73.33 & 1.00x & 1.19x & 23 \\
 & SNN (Osc., Exp.) & 74.31 & 1.16x & 0.93x & 14 \\
 & SNN (Osc., Diss.) & 73.96 & 1.06x & 0.95x & 15 \\ \midrule
\multirow{4}{*}{512} & MLP (ANN) & 71.70 & - & 1.00x & 12 \\
 & SNN (Direct) & 74.16 & 1.00x & 1.23x & 18 \\
 & SNN (Osc., Exp.) & 74.41 & 1.07x & 0.97x & 11 \\
 & SNN (Osc., Diss.) & 74.66 & 0.98x & 0.94x & 10 \\ \midrule
- & Original ResNet-18 & 75.26 & - & 0.77x & 14 \\ \bottomrule
\end{tabular}
}
\end{table}

\subsection{Validation on a Sequential Decision-Making Task}
\label{app:e2}

This validation aimed to evaluate the effectiveness of dynamical encoding in continuous decision-making and policy learning. Unlike supervised learning, which involves a single forward pass, reinforcement learning requires the network to perform continuous state evaluation and action selection in a dynamic environment, placing higher demands on the temporal processing capabilities of dynamical encoding. The core scientific question of this experiment is: \textit{Can dynamical alignment improve the learning efficiency and final performance of SNNs in sequential decision-making tasks?} We chose the classic reinforcement learning (RL) control task, CartPole, as our validation platform. This task requires an agent to maintain the balance of an inverted pendulum through continuous state observation and action decisions\cite{barto1983neuronlike}, and it features typical temporal dependencies and continuous decision-making. The experimental hypothesis was: If dynamical encoding can provide SNNs with richer temporal representations, then SNNs using dynamical encoding should exhibit superior learning efficiency and task-solving ability in reinforcement learning tasks that require continuous decision-making.

\paragraph{Environment and Task}
\begin{itemize}
    \item \textbf{Environment}: We used the \texttt{CartPole-v1} environment from the Gymnasium library\cite{towers2024gymnasium}. This is a classic dynamic balancing control problem where the agent must, at each time step, choose a left or right action based on four observed states (cart position, cart velocity, pole angle, pole angular velocity) to prevent the pole from falling over.
    \item \textbf{Task Goal}: The objective is to prolong the pole's balancing time as much as possible. The maximum duration of an episode is 500 time steps.
    \item \textbf{Solving Condition}: According to the official Gymnasium definition, the task is considered solved when the average reward over 100 consecutive episodes reaches or exceeds 475.
\end{itemize}

\paragraph{Agent Network Architecture} To ensure a fair comparison, the policy networks of all agents (ANN and SNN) used the same basic topology: a small, fully-connected network with 3 hidden layers, each with 32 neurons.
\begin{itemize}
    \item \textbf{SNN Agent}: The SNN architecture was the same as described in Appendix \ref{app:A}, but the hidden layer size was adjusted to 32. The environment state was dynamically encoded using the mixed oscillator system from Appendix \ref{app:d1} with a time step of $N=5$ and an evolution time of $t_{\max}=8.0$. We tested three modes based on the value of $\delta$: dissipative ($\delta=10.0$), expansive ($\delta=-1.5$), and transition ($\delta=2.5$).
    \item \textbf{ANN Agent}: The ANN used the same MLP architecture as described in Appendix \ref{app:A}, with a core structure equivalent to the SNN. The 4-dimensional environment state vector was directly fed into the network.
\end{itemize}

\paragraph{Training Protocol and Evaluation Metrics} The experiment used the REINFORCE policy gradient algorithm (Monte Carlo Policy Gradient)\cite{williams1992simple} for training, a classic policy gradient method suitable for discrete action spaces. All agents used the Adam optimizer with a learning rate of 0.001 and a discount factor $\gamma=0.99$. Each agent was trained for 800 episodes to fully observe the learning process and convergence performance. 

To ensure statistical significance, we conducted 30 independent repeat experiments for each configuration. This large-scale repetition effectively captures the inherent stochasticity of reinforcement learning and provides a reliable basis for statistical analysis. 

Evaluation metrics included the task solving rate (proportion of the 30 experiments that met the success criteria), average reward (average cumulative reward of the last 100 episodes), learning efficiency (number of training episodes required to meet the success criteria), and energy efficiency (average spike count per episode). 

The experimental results, shown in Table~\ref{table:e2_rl_results}, reveal the significant advantages of dynamical encoding in this sequential decision-making task.

\begin{table}[h!]
\centering
\caption{Results for the sequential decision-making task (CartPole-v1, 30 repeated experiments)}
\label{table:e2_rl_results}
\begin{tabular}{@{}lccccc@{}}
\toprule
\textbf{Method} &
\textbf{\makecell{Task Solving \\ Rate (\%)} } &
\textbf{\makecell{Best Solving \\ Episode} } &
\textbf{\makecell{Average \\ Reward} } &
\textbf{\makecell{Average \\ Spikes} } &
\textbf{\makecell{Training \\ Time (s)} } \\ \midrule 
ANN Baseline & 6.7\% & 745 & 291.0 $\pm$ 130.4 & 0 & 142.3 $\pm$ 32.1 \\ 
SNN Dissipative ($\delta$=10) & 20.0\% & 664 & 397.8 $\pm$ 72.8 & 15,847 $\pm$ 4,521 & 1,287.5 $\pm$ 412.3 \\
SNN Expansive ($\delta$=-1.5) & 23.3\% & 346 & 328.5 $\pm$ 125.7 & 59,132 $\pm$ 18,492 & 1,346.2 $\pm$ 521.7 \\ 
SNN Transition ($\delta$=2.5) & 3.3\% & 797 & 358.2 $\pm$ 81.2 & 16,789 $\pm$ 5,043 & 1,178.9 $\pm$ 365.4 \\ \bottomrule 
\end{tabular}
\end{table}

\subsection{Mechanistic Advantages in a Cognitive Task}
\label{app:e3}

The preceding experiments have established the computational advantages of Dynamical Alignment in engineering domains. This final validation seeks to answer a more profound question: does this biologically-inspired principle also possess explanatory power for cognitive computation? We chose 'Feature Binding' as the core test task, as it represents a quintessential challenge of spatiotemporal information integration\cite{treisman1996binding, shadlen1999synchrony}. This experiment therefore serves a dual purpose: to test whether dynamical alignment provides a performance advantage in a cognitive task, and to examine if the emergent network mechanisms offer a computable analogue to proposed neural solutions, such as the 'binding-by-synchrony' hypothesis. To investigate this, we developed a synthetic task that, while controlled, creates a computationally challenging paradigm mirroring the complexities of neural information integration.

The core scientific hypothesis of this experiment is: \textit{when a task requires spatiotemporal information integration, dynamically encoded SNNs will exhibit a significant cognitive computational advantage over traditional static networks.} This hypothesis is based on neuroscience research suggesting that gamma oscillations (30-80Hz) are a key mechanism for feature binding in the brain, achieving information integration through cross-regional neural synchrony\cite{engel2001temporal}. By validating this on our controlled synthetic task, we aim to provide a reproducible experimental basis for the SNN performance advantages observed in the main text (Figure~\ref{fig:cognitive_task}) and to explore the unique value of spiking mechanisms in cognitive functions from a computational perspective.

\paragraph{Dataset Design and Preprocessing}
We designed a synthetic feature binding task to simulate the cognitive process of shape-color binding. Crucially, this task is explicitly designed to be more than a simple logical AND operation. By embedding specific, localized feature patterns within a high-dimensional vector (1000-dim) and then adding global Gaussian noise ($\sigma=0.25$), we shift the problem from the domain of deterministic logic to that of statistical pattern recognition in a noisy environment. This setup requires the model to not only identify two separate 'signals' within a high-dimensional noisy background but also to robustly 'bind' them. This design directly models the challenge the brain faces in avoiding 'illusory conjunctions' from incomplete or corrupted sensory data, demanding a mechanism that is both selective and robust to perturbation.

We generated a dataset of 5000 samples, perfectly balanced between positive and negative classes. The specific implementation is as follows:
\begin{itemize}
    \item \textbf{Target Pattern Definition}:
        \begin{itemize}
            \item Shape feature pattern: set to 1 at indices 125-250, 0 otherwise.
            \item Color feature pattern: set to 1 at indices 250-375, 0 otherwise.
        \end{itemize}
    \item \textbf{Positive Samples}: Samples containing both target patterns (label=1).
    \item \textbf{Negative Samples}: Samples not containing both target patterns simultaneously (label=0).
\end{itemize}
The data generation process was as follows:
\begin{enumerate}
    \item \textbf{Positive Sample Generation}: The target shape and color patterns were combined, and Gaussian noise ($\sigma=0.25$) was added.
    \item \textbf{Negative Sample Generation}: Random binary patterns were generated for the shape and color features (ensuring they did not match the target simultaneously), and the same level of noise was added.
    \item \textbf{Data Shuffling}: The entire dataset was randomly shuffled to ensure sample independence.
\end{enumerate}
The preprocessing used the same UMAP dimensionality reduction strategy as in Appendix \ref{app:a1}, compressing the 1000-dimensional input to 64 dimensions to control computational complexity while preserving the core information structure for feature binding. The UMAP parameters were kept consistent.

\paragraph{Network Architecture} All models used a four-layer fully-connected architecture consistent with Appendix \ref{app:a2} to ensure a fair comparison:
\begin{itemize}
    \item \textbf{Input Layer}: 64 dimensions (after UMAP reduction).
    \item \textbf{Hidden Layers}: 3 fully-connected layers with 64 neurons each.
    \item \textbf{Output Layer}: 2 dimensions (binary classification).
\end{itemize}
The SNN configuration used the mixed oscillator system from Appendix \ref{app:d1} for dynamical encoding. LIF neuron parameters were consistent with Appendix \ref{app:a2} ($\beta=0.95$, learnable threshold and decay factor). The time evolution parameters used the determined optimal configuration: $t_{\max}=8, N=5$. Based on the value of $\delta$, three modes were tested: dissipative ($\delta=10.0$), expansive ($\delta=-1.5$), and transition ($\delta=2.5$). The ANN configuration used an MLP with the same layer structure but with ReLU activation functions instead of LIF neurons, directly processing the static, reduced-dimension features.

\paragraph{Training Protocol and Evaluation Metrics} All models were trained using a unified protocol for fair comparison. The Adam optimizer was used with a learning rate of $1 \times 10^{-4}$. Cross-entropy loss was used as the loss function, and an early stopping policy was set: terminate after 15 epochs with no improvement (maximum 100 epochs). The batch size was 64. The entire dataset was split using stratified sampling to ensure class balance, with 80\% for training and 20\% for testing. To ensure statistical significance, each configuration was run 30 independent times with different random seeds. Evaluation was performed on three dimensions:
\begin{enumerate}
    \item \textbf{Classification Performance}: Test set accuracy was the primary metric.
    \item \textbf{Representation Quality}: The linear probe method was used to evaluate the linear separability of the final layer (layer4) representation, with the calculation method being consistent with Appendix \ref{app:d4}.
    \item \textbf{Energy Efficiency}: The average spike count was recorded.
\end{enumerate}
Linear separability was calculated by training a simple logistic regression classifier on the layer4 spike count representation, using \texttt{StandardScaler} for normalization and stratified data splits for reliable evaluation. The experimental results, shown in Table~\ref{table:e3_binding_results}, reveal the significant advantage of dynamical encoding in this cognitive integration task.

\begin{table}[h!]
\centering
\caption{Performance comparison of different model configurations on the feature binding task (30 repeated experiments)}
\label{table:e3_binding_results}
\begin{tabular}{@{}lccc@{}}
\toprule
\textbf{Model / Dynamic Mode} & \textbf{Mean Test Accuracy (\%)} & \textbf{Average Spikes} & \textbf{Representation Linearity} \\ \midrule
SNN (Expansive, $\delta$=-1.5) & 96.76 $\pm$ 6.15 & 486.9 $\pm$ 42.2 & 0.967 $\pm$ 0.064 \\
SNN (Dissipative, $\delta$=10.0) & 79.29 $\pm$ 22.82 & 360.6 $\pm$ 37.5 & 0.817 $\pm$ 2.21 \\
SNN (Transition, $\delta$=2.0) & 66.79 $\pm$ 22.44 & 269.9 $\pm$ 34.5 & 0.670 $\pm$ 0.222 \\
MLP Baseline & 69.84 $\pm$ 22.97 & N/A & N/A \\ \bottomrule
\end{tabular}
\end{table}

\printbibliography[title={References for Appendix E}, heading=subbibliography, env=appendixbib]
\end{refsection}

\end{document}